\pdfoutput=1

\documentclass[11pt]{article}

\usepackage[preprint]{acl}

\usepackage{times}
\usepackage{latexsym}

\usepackage[T1]{fontenc}

\usepackage[utf8]{inputenc}

\usepackage{microtype}

\usepackage{inconsolata}

\usepackage{graphicx}

\usepackage{mathrsfs}
\usepackage{paralist}
\usepackage{booktabs}
\usepackage{multirow}
\usepackage{pgfplots}
\usepackage{scalefnt}
\usepackage{amsthm,amsmath,amssymb}
\usepackage{mathtools}
\usepackage{bm}
\usepackage[edges]{forest}
\usepackage{soul} 
\usepackage{color, xcolor} 
\usepackage{lipsum}  
\usepackage{colortbl}
\usepackage{booktabs}
\usepackage{array}
\usepackage{stfloats}
\usepackage{tikz-dependency}
\usepackage{tikz}
\usepackage{siunitx}
\usepackage{graphicx}
\usepackage[ruled]{algorithm2e}
\usepackage{lineno}
\usepackage{tabularx}
\usepackage{subcaption}
\usepackage{helvet}
\usepackage{courier}
\usepackage{makecell}
\usepackage{nicematrix}
\usepackage[fixed]{fontawesome5}
\usepackage{CJKutf8}
\pgfplotsset{compat=1.18}
\title{A Systematic Survey of Semantic Role Labeling in the Era of Pretrained Language Models}

\author{
 \textbf{Huiyao Chen\textsuperscript{1,2}, }
 \textbf{Meishan Zhang\textsuperscript{1}, }
 \textbf{Jing Li\textsuperscript{1}, }
 \textbf{Lilja Øvrelid\textsuperscript{3}} \\
 \textbf{Jan Hajič\textsuperscript{4}, }
 \textbf{Hao Fei\textsuperscript{1}}\Thanks{The corresponding author}
 \textbf{, Min Zhang\textsuperscript{1,2}}
  \\
  \textsuperscript{1}Harbin Institute of Technology (Shenzhen)
  \quad
  \textsuperscript{2} Shenzhen Loop Area Institute (SLAI)
  \\
  \textsuperscript{3} University of Oslo
  \quad
  \textsuperscript{4} Charles University
  \\
  \texttt{chenhy1018@gmail.com, mason.zms@gmail.com}
  \\
  \texttt{jingli.phd@hotmail.com, zhangmin2021@hit.edu.cn}
  \\
  \texttt{liljao@ifi.uio.no, hajic@ufal.mff.cuni.cz}
  \\
  \texttt{haofei37@nus.edu.sg}
}

\begin{document}
\maketitle


\begin{abstract}
Semantic role labeling (SRL) is a central natural language processing task for understanding predicate-argument structures within texts and enabling downstream applications.
Despite extensive research, comprehensive surveys that critically synthesize the field from a unified perspective remain lacking.
This survey makes several contributions beyond organizing existing work.
We propose a unified four-dimensional taxonomy that categorizes SRL research along model architectures, syntax feature modeling, application scenarios, and multimodal extensions.
We provide a critical analysis of when and why syntactic features help, identifying conditions under which syntax-aided approaches provide consistent gains over syntax-free counterparts.
We offer the first systematic treatment of SRL in the era of large language models, examining the complementary roles of LLMs and specialized SRL systems and identifying directions for hybrid approaches.
We extend the scope of SRL surveys to cover multimodal settings including visual, video, and speech modalities, and analyze structural differences in evaluation across these modalities.
Literature was collected through systematic searches of the ACL Anthology, IEEE Xplore, the ACM Digital Library, and Google Scholar, covering publications from 2000 to 2025 and applying explicit inclusion and exclusion criteria to yield approximately 200 primary references.
SRL benchmarks, evaluation metrics, and paradigm modeling approaches are discussed alongside practical applications across domains.
Future research directions are analyzed, addressing the evolving role of SRL with large language models and broader NLP impact.
\end{abstract}

\section{Introduction}
\label{sec1}

Within NLP, SRL~\citep{DBLP:conf/acl/GildeaJ00} involves identifying the semantic roles of words or phrases in a sentence.
These roles represent the relationships between various components of a sentence, specifically the who, what, when, where, how, and why of the actions described.
SRL helps determine the underlying meaning of a sentence by labeling the different arguments associated with a verb (the predicate).
As a result, SRL serves as an important step in relevant downstream applications and benefits a multitude of tasks in NLP, including information extraction~\citep{christensen-etal-2010-semantic,DBLP:conf/kcap/ChristensenMSE11,DBLP:conf/ranlp/EvansO19}, machine translation~\citep{DBLP:conf/acl/ShiLRFLZSW16,DBLP:conf/naacl/MarcheggianiBT18}, and question answering~\citep{DBLP:conf/emnlp/ShenL07,DBLP:conf/emnlp/BerantCFL13,DBLP:conf/emnlp/HeLZ15,DBLP:conf/acl/YihRMCS16}.

While previous books and surveys have discussed SRL or broader semantic representation~\citep{DBLP:series/synthesis/2010Palmer,DBLP:conf/acl/AbendR17,jm3}, they predate the widespread adoption of LLMs and do not cover multimodal SRL or provide actionable guidance on syntactic feature modeling.
To address these gaps, this survey makes the following original contributions.
First, we propose a unified four-dimensional taxonomy organizing SRL research along model architectures, syntax feature modeling, application scenarios, and multi-modal extensions, which provides a common framework for comparing methods across the FrameNet~\citep{Baker2017} and PropBank~\citep{DBLP:journals/coling/PalmerKG05} paradigms.
Second, we conduct a critical analysis of the syntax-aided versus syntax-free debate, identifying the specific conditions under which explicit syntactic supervision provides consistent benefits and explaining contradictions in previously reported results.
Third, we provide a systematic analysis of SRL in the LLM era, characterizing the complementary roles of LLMs and specialized systems and identifying concrete hybrid directions that go beyond simply applying prompting to SRL benchmarks.
Fourth, we present the first integrated treatment of multimodal SRL across visual, video, and speech modalities, analyzing structural differences in task formulation and evaluation that make cross-modal comparisons non-trivial.
Fifth, we include a cross-paradigm error analysis and a practical method-selection guide that translates research findings into actionable recommendations for practitioners across both English and non-English contexts~\citep{DBLP:journals/taslp/FeiZLJ20,DBLP:conf/coling/ConiaN20}.

\begin{figure*}[ht]
    \centering
    \includegraphics[width=.95\linewidth]{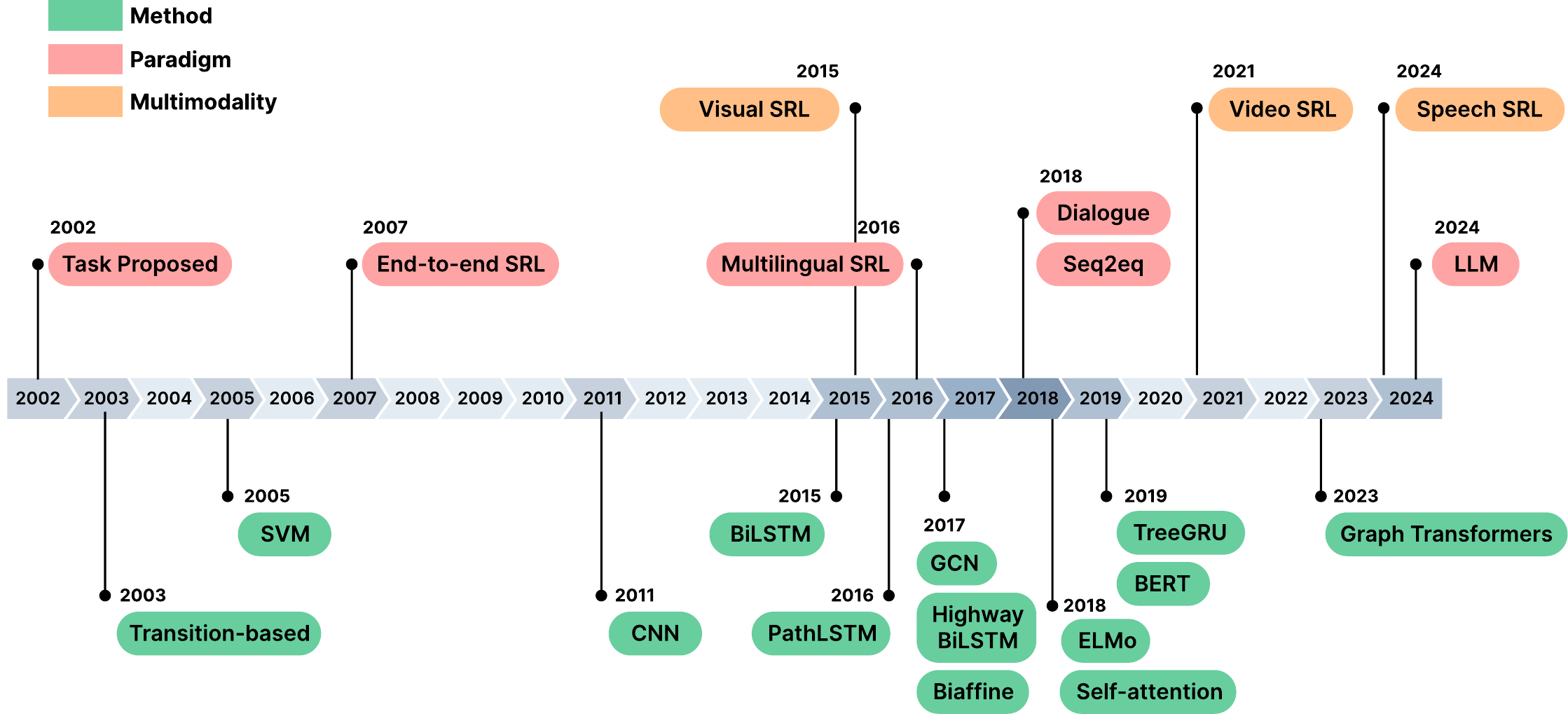}
    \caption{Timeline of key milestones in semantic role labeling research from 2002 to 2024.}
    \label{fig:intro}
\end{figure*}

Figure~\ref{fig:intro} illustrates key milestones in SRL research, starting with the proposal of this task and continuing with the latest progress driven by LLMs.
Concretely, the SRL task was pioneered by~\citet{DBLP:conf/acl/GildeaJ00}, building upon Frame Semantics~\citep{fillmore1976frame,DBLP:conf/acl/BakerFL98} as its theoretical foundation.
Early NLP research treated SRL as fundamental to natural language understanding, with initial algorithms inevitably incorporating syntactic information for feature modeling~\citep{DBLP:conf/emnlp/XueP04,DBLP:conf/conll/HaciogluPWMJ04,DBLP:conf/acl/PradhanWHMJ05,DBLP:conf/acl/SwansonG06,DBLP:conf/emnlp/JohanssonN08}.
A significant breakthrough came when~\citet{DBLP:journals/jmlr/CollobertWBKKK11} developed the first end-to-end SRL system using multilayer neural networks.
Subsequently, most SRL research has shifted toward developing end-to-end methods, making extensive use of deep neural models~\citep{DBLP:conf/acl/ZhouX15,DBLP:conf/emnlp/FitzGeraldTG015,DBLP:conf/acl/HeLLZ17a,DBLP:conf/paclic/OkamuraTITIIAU18} and generative models~\citep{DBLP:conf/rep4nlp/DazaF18,DBLP:conf/ijcai/BlloshmiCTN21}.

In its early stages, SRL was conceptualized as a text-only, single-sentence task.
Although treating SRL as a sentence-level task provided a practical foundation for analysis, it failed to capture the rich contextual information that often spans multiple sentences or even entire conversations.
Semantic roles often extend beyond the boundaries of the sentence, with arguments and predicates establishing relationships in a wider discourse context.
Recent research has therefore expanded SRL to discourse-level and conversational settings~\citep{DBLP:journals/lre/RuppenhoferLSM13,DBLP:journals/coling/RothF15,DBLP:conf/icmlc2/HeWLSC21,DBLP:conf/naacl/WuT0LWS22,DBLP:conf/ijcai/0001WZRJ22}, as well as to multimodal scenarios such as visual SRL in images and videos~\citep{DBLP:journals/corr/GuptaM15,DBLP:conf/cvpr/YatskarZF16,DBLP:conf/eccv/PrattYWFK20,DBLP:conf/cvpr/SadhuGYNK21,DBLP:conf/nips/KhanJT22,DBLP:conf/aaai/YangLZ0JC23,DBLP:conf/mm/Zhao00LZWZC23} and recent advances in speech-based SRL~\citep{DBLP:conf/acl/ChenLZZ24}.
The emergence of LLMs has further transformed the landscape of SRL.
This development has raised important questions in the NLP community regarding traditional tasks like SRL: (1) What is the significance of SRL in this era?
(2) How will SRL research develop in the future?
This survey seeks to systematically review these developments and provide insights into the evolving role of SRL in modern NLP.

To ensure transparency and reproducibility, we describe the procedure used to identify and select the literature covered in this survey.
We conducted systematic searches across four major academic databases: the ACL Anthology, IEEE Xplore, the ACM Digital Library, and Google Scholar.
The primary search terms included ``semantic role labeling'', ``predicate argument structure'', ``frame semantic parsing'', ``situation recognition'', and combinations of these terms with modifiers such as ``neural'', ``multimodal'', ``cross-lingual'', and ``large language model''.
The search covered publications from 2000 to 2025, with 2000 chosen as the starting point because it corresponds to the seminal work of~\citet{DBLP:conf/acl/GildeaJ00} that formally introduced the SRL task.
A paper was included if it met at least one of the following criteria: it proposed a new model or method directly targeting SRL or a closely related structured prediction task; it introduced a benchmark dataset or annotation scheme for SRL evaluation; it conducted an empirical study whose findings bear directly on SRL methodology or evaluation; or it applied SRL to a downstream task in a way that produced insights relevant to SRL system design.
Papers were excluded if they addressed semantic analysis tasks without any direct connection to predicate-argument structure, if they were workshop abstracts without accompanying technical content, or if they were superseded by a substantially extended journal version that was already included.
After applying these criteria, the final corpus covers approximately 200 primary references spanning textual, visual, video, and speech SRL, as well as related tasks such as abstract meaning representation and event extraction where direct comparisons with SRL are drawn.

The remainder of this survey is organized as follows.
We start by laying the definitions of SRL tasks (\S\ref{Task Definition}), covering both traditional text-based and emerging multi-modal scenarios.
Next, we present an overview taxonomy of SRL methodologies (\S\ref{Taxonomy of SRL Methodology}), categorizing the field into four crucial perspectives: model architectures, syntax feature modeling, various application scenarios, and multi-modal extensions.
We then review major SRL benchmarks and evaluation metrics (\S\ref{Benchmarks}), followed by in-depth discussions of different SRL methods (\S\ref{Methods in SRL}) and paradigm modeling approaches (\S\ref{Paradigm Modeling in SRL}).
The survey continues with an analysis of syntax feature modeling (\S\ref{Syntax Feature Modeling in SRL}) and explores SRL applications under various scenarios (\S\ref{SRL under Various Scenarios}), including cross-sentence, multilingual, and multimodal contexts.
We further examine practical applications across different domains (\S\ref{Applications}).
Finally, we discuss future research directions (\S\ref{Feature Research}) before concluding the survey (\S\ref{Conclusion}).
We maintain a public repository and consistently update related resources at: \textbf{\tt \url{https://github.com/DreamH1gh/Awesome-SRL}}

\section{Task Definition}
\label{Task Definition}

In a broad sense, SRL falls under the subtask of semantic analysis. 
Related tasks in semantic analysis also include abstract meaning representation (AMR) and event extraction (EE), etc.

\subsection{General Definition of SRL}

SRL aims to uncover the predicate-argument structure of a sentence, answering essential questions such as who did what to whom, when, where, and how.
In this framework, a \textbf{predicate} (typically a verb or nominalization) denotes an event or state, while its \textbf{arguments} fulfill specific semantic roles in relation to that predicate.
This structure is fundamental to deep semantic understanding and supports a wide range of downstream NLP tasks.

\subsubsection{Major SRL Paradigms}
Over the past decades, three major annotation paradigms have shaped the development of SRL, each with distinct design philosophies and annotation practices rooted in established linguistic theories.
Each paradigm derives its labeling scheme from a specific theoretical ontology that defines what constitutes a semantic role, how roles are organized, and what linguistic evidence licenses a given role assignment.
\begin{compactitem}
    \item \textbf{FrameNet}~\citep{DBLP:conf/acl/BakerFL98} organizes semantic knowledge around conceptual frames grounded in Fillmore's Frame Semantics theory~\citep{fillmore1976frame}. Frame Semantics holds that words evoke structured conceptual scenarios, and that understanding a word requires understanding the entire frame it activates. This theoretical foundation motivates the frame-based annotation scheme, in which each lexical unit is linked to a frame that defines the set of applicable frame elements and their semantic constraints. The validity of this ontology rests on decades of cognitive linguistic research demonstrating that human semantic knowledge is organized around prototypical event schemas rather than isolated word meanings~\citep{fillmore2006frame}. FrameNet employs a frame-based approach to annotate semantic roles through text spans, which may be discontinuous to represent complex argument structures.
    \item \textbf{PropBank}~\citep{DBLP:conf/lrec/KingsburyP02,kingsbury2002adding} provides semantic role annotations for predicates across lexical categories, with its labeling scheme grounded in predicate-specific rolesets derived from verb-class generalizations in the tradition of lexical semantics~\citep{levin1993english}. The numbered argument labels (ARG0, ARG1, etc.) are not universal thematic roles but are defined on a per-predicate basis within each roleset, drawing on the observation that verbs of similar meaning share systematic argument alternation patterns. The validity of this approach is supported by the high inter-annotator agreement achieved for core argument roles and by the broad cross-linguistic applicability demonstrated through the Universal PropBank project~\citep{DBLP:conf/lrec/JindalRULNT0022}. PropBank systematically labels predicate-argument structures in sentences with numbered arguments that correspond to predicate-specific semantic roles defined within each roleset, rather than universal thematic roles shared across predicates.
    \item \textbf{NomBank}~\citep{DBLP:conf/naacl/MeyersRMSZYG04} complements PropBank by focusing on nominal predicates and their argument structures, inheriting PropBank's roleset ontology as its annotation framework while drawing on the NOMLEX nominalization lexicon~\citep{macleod1998nomlex} as a lexical resource to map nominal predicates to their verbal counterparts. The underlying theoretical motivation is that verbal nominalizations preserve the argument structure of their base verbs, a claim supported by extensive work in generative lexicon theory and morphosyntax~\citep{grimshaw1990argument}. This theoretical grounding justifies applying PropBank-style numbered argument labels to nominal predicates, thus extending semantic role annotation beyond verbs for more comprehensive coverage.
\end{compactitem}
These foundational resources collectively address distinct aspects of predicate-argument structures in natural language.
Their respective ontologies reflect different but complementary theoretical commitments: FrameNet prioritizes conceptual richness and frame-level generalization, PropBank prioritizes broad coverage and cross-linguistic consistency, and NomBank prioritizes unified treatment of verbal and nominal predication.
A more detailed comparison of these paradigms and their associated datasets is provided in \S\ref{datasets}.

\subsubsection{Formal Definitions}
\label{sec:formal}
Reflecting annotation practices of these paradigms, SRL is typically formalized in two main ways:
\begin{compactitem}
    \item \textbf{Span-based SRL:} Semantic roles are assigned to contiguous spans of text that correspond to arguments of a predicate. Each argument is represented by two boundary indices, namely a start position and an end position within the sentence, forming a 2:1 mapping from a single argument to a pair of token positions.
    \item \textbf{Dependency-based SRL:} Semantic roles are assigned based on syntactic dependency relations, linking predicates directly to their arguments via the sentence's dependency structure. Each argument is represented by a single head word, forming a 1:1 mapping from a single argument to a single token position.
\end{compactitem}
Formally, given a sentence $S=\{w_1,...,w_n\}$, SRL aims to predict a set of triplets $Y=\{...,<p_k,a_k,r_k>,...\mid p_k\in \mathcal{P}, a_k\in \mathcal{A}, r_k\in \mathcal{R}\}$, where $\mathcal{P}$, $\mathcal{A}$, and $\mathcal{R}$ denote the sets of all possible predicates, arguments, and semantic role labels, respectively.
In span-based SRL, each argument $a_k$ is a text span $a_k = \{w_i,...,w_j\} \subseteq S$, where the pair of indices $(i, j)$ jointly identifies the argument boundary, and a single argument thus corresponds to two positional anchors in the sentence.
In dependency-based SRL, each argument $a_k$ is the syntactic head word of the corresponding argument phrase, i.e., $a_k = w_i \in S$, where a single argument corresponds to exactly one token position.
This structural difference means that span-based SRL requires predicting both the start and end boundaries of each argument, whereas dependency-based SRL requires identifying only the head word, making the two paradigms fundamentally distinct in how argument positions are encoded.
It is worth noting that in PropBank and FrameNet, the semantic meaning of a role label $r_k$ is conditioned on the sense of its predicate $p_k$.
In PropBank, each verb is annotated with a specific sense number and an associated roleset, so that the same argument label may carry different semantic interpretations across different predicate senses.
In FrameNet, the predicate sense corresponds to the evoked frame, which in turn determines the set of applicable frame elements.
Therefore, predicate sense identification is an integral step in fully resolving the semantic content of the predicted triplets, even though the triplet formulation itself remains unchanged.

\newcolumntype{C}[1]{>{\hsize=#1\hsize\centering\arraybackslash}X}

\definecolor{c2}{RGB}{237,110,155} 

\begin{table*}[ht]
\fontsize{9}{10}\selectfont
\setlength{\tabcolsep}{1.3mm}

\begin{center}
\begin{tabularx}{\textwidth}{
l
>{\hsize=0.25\hsize\arraybackslash}X
>{\hsize=0.7\hsize\arraybackslash}X
}
\Xhline{0.08em}
\rowcolor{blue!15} \bf Role & \bf Description & \bf Example\\ \hline
ARG0 & agent & \textcolor{c2}{\textbf{[$_{\text{ARG0}}$ He]}} got a sense of his soul.\\
\rowcolor{gray!15} ARG1 & patient & He got a sense \textcolor{c2}{\textbf{[$_{\text{ARG1}}$ of his soul]}}.\\
ARG2 & instrument, benefactive, attribute & I had no right \textcolor{c2}{\textbf{[$_{\text{ARG2}}$ to print that]}}.\\
\rowcolor{gray!15} ARG3 & starting point, benefactive, attribute & It moves \textcolor{c2}{\textbf{[$_{\text{ARG3}}$ from cities]}} to rural areas.\\
ARG4 & ending point & It moves from cities \textcolor{c2}{\textbf{[$_{\text{ARG4}}$ to rural areas]}}. \\
\rowcolor{gray!15} ARGM & modifier & \\
~~-COM & Comitative & I sang \textcolor{c2}{\textbf{[$_{\text{ARGM-COM}}$ with my sister]}}.\\
\rowcolor{gray!15} ~~-LOC & Locative & They are playing \textcolor{c2}{\textbf{[$_{\text{ARGM-LOC}}$ on the ground]}}. \\
~~-DIR & Directional & He walk \textcolor{c2}{\textbf{[$_{\text{ARGM-DIR}}$ forward]}} to the house.\\
\rowcolor{gray!15} ~~-GOL & Goal & The child fed the cat \textcolor{c2}{\textbf{[$_{\text{ARGM-GOL}}$ for her mother]}}. \\
~~-MNR & Manner & The plumber unclogged the sink \textcolor{c2}{\textbf{[$_{\text{ARGM-MNR}}$ with a drain snake]}}.\\
\rowcolor{gray!15} ~~-TMP & Temporal & Four of the five surviving workers have asbestos-related diseases, including three with \textcolor{c2}{\textbf{[$_{\text{ARGM-TMP}}$ recently diagnosed cancer]}}.\\
~~-EXT & Extent & He may care \textcolor{c2}{\textbf{[$_{\text{ARGM-EXT}}$ more]}} about Senior Olympic games.\\
\rowcolor{gray!15} ~~-REC & Reciprocals & If the stadium was such a good idea someone would build it \textcolor{c2}{\textbf{[$_{\text{ARGM-REC}}$ himself]}}. \\ 
~~-PRD & Secondary Predication & This wage inflation is bleeding the NFL \textcolor{c2}{\textbf{[$_{\text{ARGM-PRD}}$ dry]}}.\\
\rowcolor{gray!15} ~~-PRP & Purpose & Commonwealth Edison could
raise its electricity rates by \$49 million \textcolor{c2}{\textbf{[$_{\text{ARGM-PRP}}$ to pay for the plant]}}.\\
~~-CAU & Cause & However, five other countries will remain on that priority watch list \textcolor{c2}{\textbf{[$_{\text{ARGM-CAU}}$ because of an interim review]}}, U.S. Trade Representative Carla Hills announced. \\
\rowcolor{gray!15} ~~-DIS & Discourse & The notification \textcolor{c2}{\textbf{[$_{\text{ARGM-DIS}}$ also]}} clarifies the requirements of the evaluation.\\
~~-ADV & Adverbials & The notification recognizes the company and \textcolor{c2}{\textbf{[$_{\text{ARGM-ADV}}$ also]}} clarifies the requirements of the evaluation.\\
\rowcolor{gray!15} ~~-ADJ & Adjectival & We get a \textcolor{c2}{\textbf{[$_{\text{ARGM-ADJ}}$ different]}} excuse for this every time.\\
~~-MOD & Modal & But voters decided that if the stadium was such a good idea someone \textcolor{c2}{\textbf{[$_{\text{ARGM-MOD}}$ would]}} build it himself, and rejected it 59\% to 41\%.\\
\rowcolor{gray!15} ~~-NEG & Negation & I had \textcolor{c2}{\textbf{[$_{\text{ARGM-NEG}}$ no]}} right to print that.\\
~~-DSP & Direct Speech & Among other things, they said \textcolor{c2}{\textbf{[$_{\text{ARGM-DSP}}$ [*?*]]}}, Mr. Azoff would develop musical acts for a new record label. // \textit{[*?*] placeholder for ellipsed material}\\
\rowcolor{gray!15} ~~-LVB & Light Verb & Yesterday, Mary \textcolor{c2}{\textbf{[$_{\text{ARGM-LVB}}$ made]}} an accusation of duplicity against John because she was enraged with jealousy.\\
~~-CXN & Construction & Hillary Clinton is \textcolor{c2}{\textbf{[$_{\text{ARGM-CXN}}$ about as damaging to the Dem Party as Jeremiah Wright]}}. \\
\bottomrule
\end{tabularx}
\end{center}
\caption{
Description of SRL arguments based on English PropBank annotation guidelines as a case study.
}
\label{tab:args}
\end{table*}

\begin{figure*}[ht]
    \centering
    \begin{subfigure}[t]{0.48\linewidth}
        \centering
        \includegraphics[width=\linewidth]{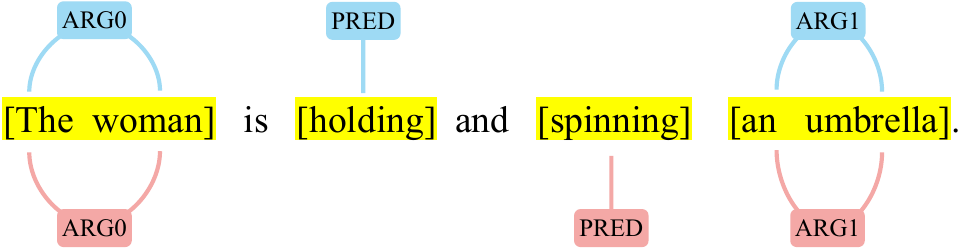}
        \caption{Span-based SRL.}
    \end{subfigure}
    \hfill
    \begin{subfigure}[t]{0.48\linewidth}
        \centering
        \includegraphics[width=\linewidth]{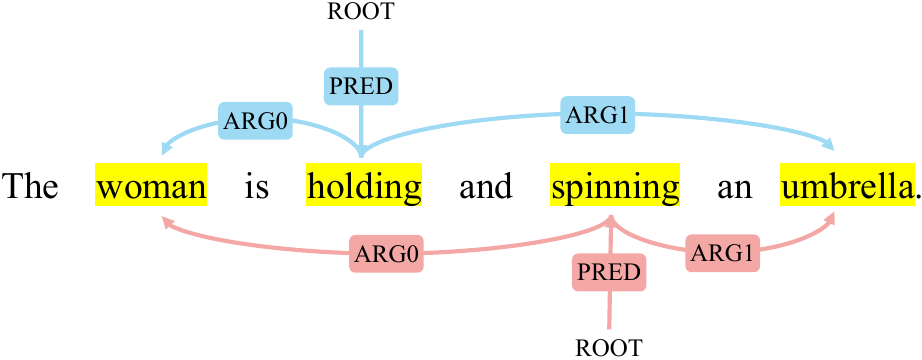}
        \caption{Dependency-based SRL.}
    \end{subfigure}
    \begin{subfigure}[t]{0.48\linewidth}
        \centering
        \includegraphics[width=\linewidth]{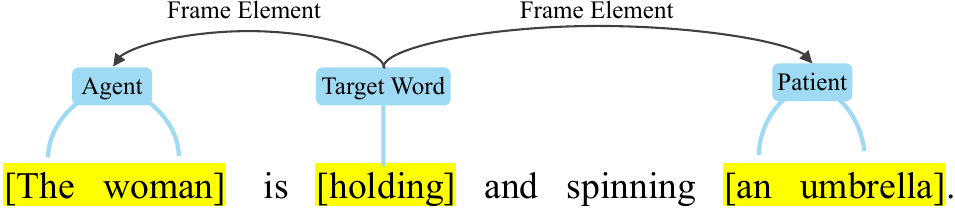}
        \caption{Frame SRL.}
    \end{subfigure}
    \hfill
    \begin{subfigure}[t]{0.48\linewidth}
        \centering
        \includegraphics[width=\linewidth]{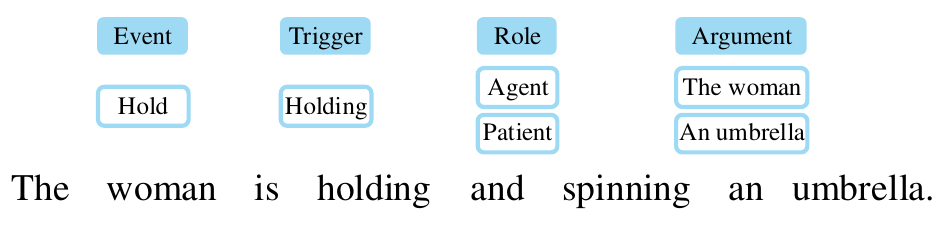}
        \caption{Event extraction (EE).}
    \end{subfigure}
    \begin{subfigure}[t]{0.6\linewidth}
        \centering
        \includegraphics[width=\linewidth]{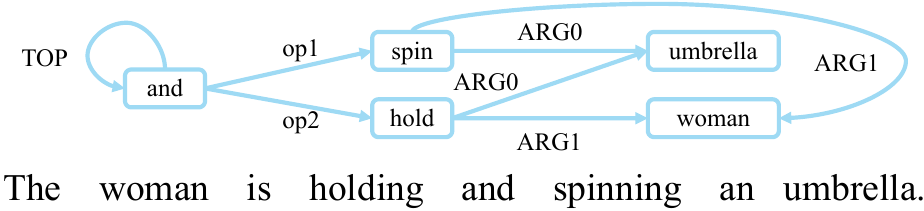}
        \caption{Abstract meaning representation (AMR).}
    \end{subfigure}
    \caption{Illustration of different semantic role labeling formulations and related semantic parsing tasks.}
    \label{fig:1}
\end{figure*}

Table~\ref{tab:args} provides examples of PropBank-style semantic roles, while Figure~\ref{fig:1} visually illustrates the difference between span-based and dependency-based SRL.
In the span-based case, arguments are explicitly marked as text spans using bracket notation, such as \textbf{[The woman]} for ARG0 and \textbf{[an umbrella]} for ARG1, where the brackets denote the boundaries of each argument span.
In the dependency-based case, only the syntactic head word of each argument phrase is linked to the predicate, without marking span boundaries.
Note that the ARG1 example in Table~\ref{tab:args}, namely ``He got a sense \textbf{[$_{\text{ARG1}}$ of his soul]}'', involves a light verb construction in which \textit{get} functions as the predicate and \textit{sense} is treated as a nominal complement rather than an independent predicate.
In PropBank annotation, \textit{sense} in this construction does not receive a semantic role tag, and the ARG1 label is assigned to the prepositional phrase \textit{of his soul}, which fills the patient role of the predicate \textit{get}.
This example illustrates that the semantic interpretation of argument labels such as ARG1 depends on the sense of the predicate, as the roleset associated with each predicate sense determines which arguments are licensed and how they are interpreted.

It should be noted that PropBank-style SRL is designed to capture predicate-argument structure at the semantic role level, and does not resolve fine-grained identity distinctions among entities of the same type.
When multiple same-type entities co-occupy a single argument position of one predicate, such as two distinct engineering components jointly serving as ARG0, they are annotated as a single argument span without further differentiation.
In application scenarios where individual entity identity must be distinguished, SRL should be used in conjunction with complementary tasks such as fine-grained named entity recognition or entity coreference resolution.
For domain-specific settings that require richer semantic constraints tied to entity types, frame-semantic approaches or ontology-grounded annotation schemes provide a more suitable foundation.
For further details on the datasets and annotation schemes underlying these paradigms, please refer to \S\ref{datasets}.

\subsection{Other Definitions}

\subsubsection{Frame SRL}
Frame SRL (FSRL) aims to identify arguments and label them with frame elements for frame-evoking targets in a sentence, as shown in Figure~\ref{fig:1}.
For a sentence $S=\{w_1,...,w_n\}$ and a target word $w_t \in S$ that evokes a frame $f$, the predicate is $p = w_t \in \mathcal{P}$.
The arguments for predicate $p$ are $a_1,...,a_k \in \mathcal{A}$, and each argument $a_i$ is assigned a semantic role label $r_i \in \mathcal{R}_f$, where $\mathcal{R}_f \subseteq \mathcal{R}$ denotes the set of frame elements defined for frame $f$.

While FSRL shares similar task formulation with traditional SRL, it differs significantly in semantic depth and granularity.
FSRL is grounded in Frame Semantics and goes beyond just assigning semantic roles to arguments.
It involves identifying the larger conceptual frame or situation that a predicate evokes.
The roles of FSRL are more detailed depending on the frame definition and the frame elements (FE).
Under this constraint, FSRL can be seen as a \textbf{Slot Filling} task.
FSRL emphasizes semantic frames and conceptual structures, providing richer semantic representations compared to traditional syntactic-oriented SRL approaches.

\begin{figure*}[t]
    \centering
    \includegraphics[width=\linewidth]{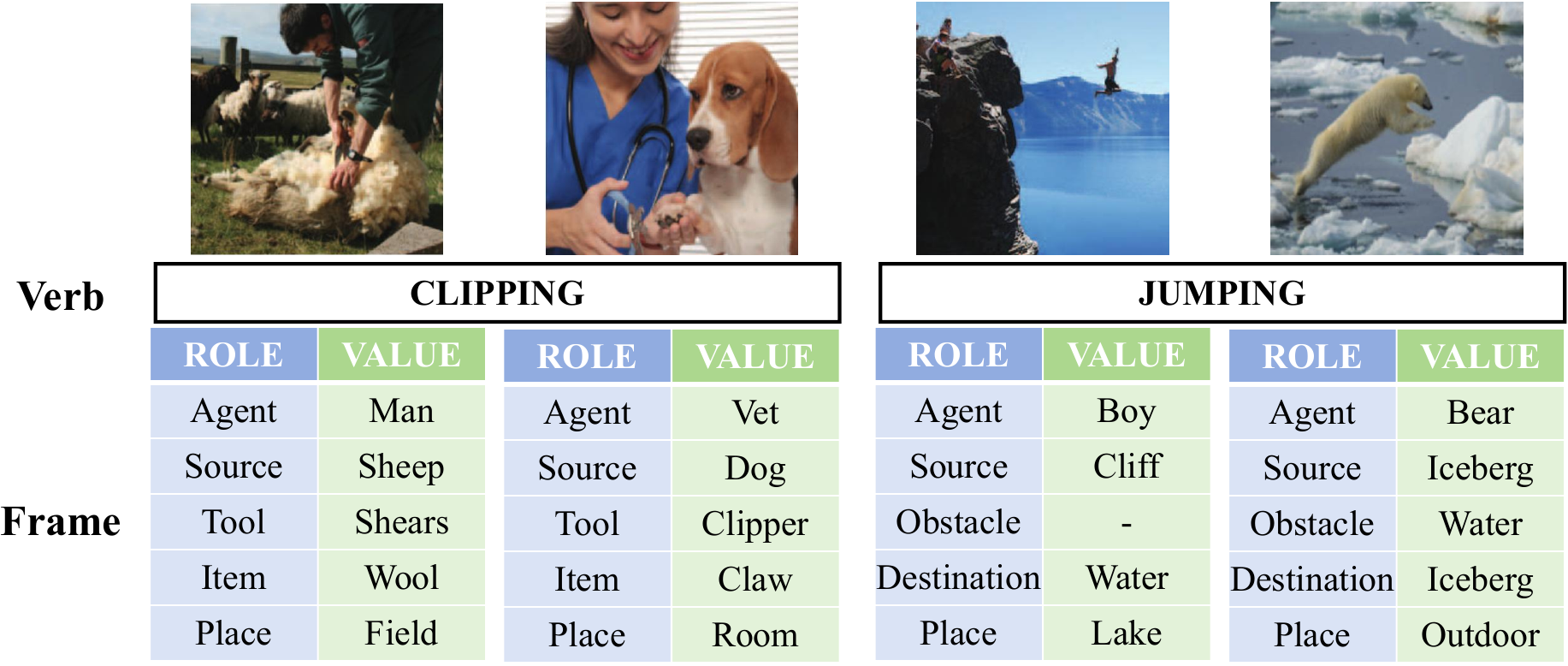}
    \caption{
    Examples of Visual SRL from the SituNet dataset.
    Each image is annotated with a verb and a realized frame consisting of role-value pairs.
    For the verb \textit{clipping}, roles such as \textit{Agent}, \textit{Source}, \textit{Tool}, \textit{Item}, and \textit{Place} are filled with scene-specific noun values.
    For the verb \textit{jumping}, the same frame structure is instantiated with different noun values across two distinct scenes, illustrating how the same verb frame can accommodate diverse visual contexts.
    The ``-'' symbol indicates that a role value is not present or not applicable in the given scene.
    }
    \label{fig:vsrl_example}
\end{figure*}

\subsubsection{Visual SRL}
VSRL is also known as situation recognition.
We follow the definition of~\citet{DBLP:conf/cvpr/YatskarZF16}.
In situation recognition, we assume a discrete set of verbs $\mathcal{V}$, a discrete set of nouns $\mathcal{N}$, and a discrete set of frames $\mathcal{F}$, where each verb $v \in \mathcal{V}$ determines a frame $f_v \in \mathcal{F}$ and its associated set of semantic roles $\mathcal{R}_{f_v} \subseteq \mathcal{R}$.
\begin{compactitem}
    \item Each semantic role $r \in \mathcal{R}_{f_v}$ is assigned a noun value $n_r \in \mathcal{N} \cup \{\varnothing\}$, where $\varnothing$ indicates the value is not known or does not apply.
    \item The set of role-value pairs forms a realized frame, $\mathcal{F}_r = \{(r, n_r) : r \in \mathcal{R}_{f_v}\}$.
    \item A realized frame is valid if and only if each role $r \in \mathcal{R}_{f_v}$ is assigned exactly one noun $n_r$.
\end{compactitem}
Given an image, the VSRL task is to predict a situation $\mathcal{S} = (v, \mathcal{F}_r)$, specified by a verb $v \in \mathcal{V}$ and a valid realized frame $\mathcal{F}_r$.
Figure~\ref{fig:vsrl_example} illustrates two representative VSRL examples with different verbs and their corresponding realized frames.
A key distinction from textual SRL lies in argument identification: textual SRL identifies argument boundaries over discrete token sequences, whereas VSRL predicts discrete noun labels for each semantic role from a fixed vocabulary.

\begin{figure*}[t]
    \centering
    \includegraphics[width=\linewidth]{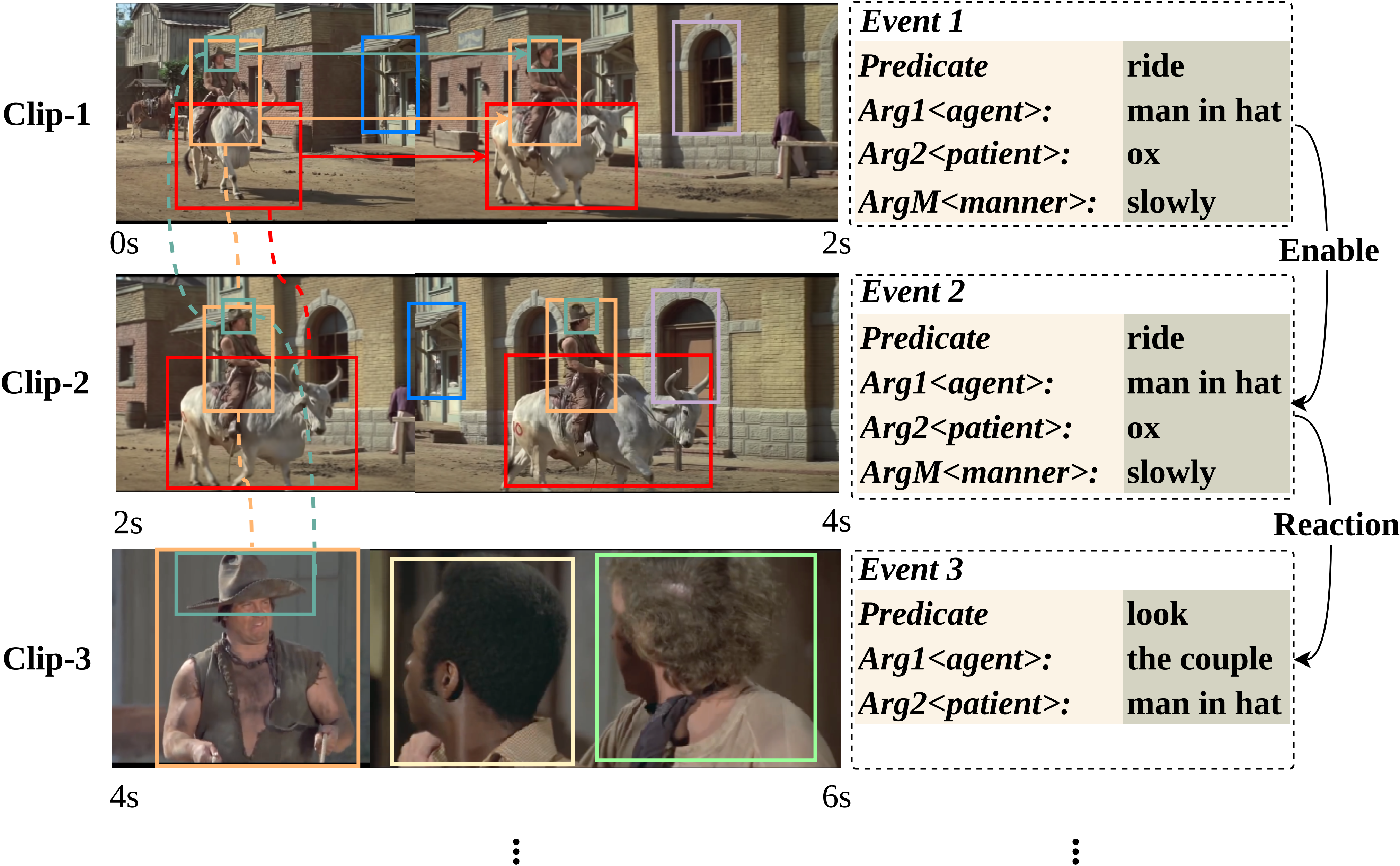}
    \caption{
    An example of Video SRL from the VidSitu dataset. Three consecutive clips are annotated with distinct events and their semantic role structures. Event~1 and Event~2 share the same predicate \textit{ride} with consistent argument assignments across clips, connected by an \textit{Enable} relation. Event~3 introduces a new predicate \textit{look}, whose arguments overlap with those of the preceding events, connected by a \textit{Reaction} relation. Colored bounding boxes indicate the spatial grounding of each argument within the respective frames.
    }
    \label{fig:vidsrl_example}
\end{figure*}

\subsubsection{Video SRL}
Video SRL (VidSRL) is proposed by~\citet{DBLP:conf/cvpr/SadhuGYNK21}.
Given a video $\mathcal{X}$, VidSRL requires a model to predict a set of related salient events $\{E_i\}_{i=1}^k$ constituting a situation.
Each event $E_i$ consists of a predicate $p_i \in \mathcal{P}$ and a set of argument-role pairs, where each argument $a_{i,j} \in \mathcal{A}$ is assigned a semantic role label $r_{i,j} \in \mathcal{R}$.
We denote the roles or arguments of a predicate $p$ as $\{a^p_j\}_{j=1}^m$, and $a^p_j \leftarrow a$ implies that the $j$-th argument of predicate $p$ is assigned the value $a$.
The relationship between any two events $E$ and $E'$ is denoted by $l(E, E') \in \mathcal{L}$, where $\mathcal{L}$ is an event-relations label set.
Compared to VSRL, VidSRL performs semantic role extraction for multiple events in a video and predicts the relationships between these events.
Figure~\ref{fig:vidsrl_example} illustrates a representative VidSRL example across three consecutive video clips.
Beyond the spatial grounding challenge shared with VSRL, VidSRL must additionally track argument entities across frames, requiring temporal consistency in role assignment that has no analogue in either textual or image-based SRL.

\subsubsection{Speech SRL}
Speech SRL aims to perform SRL directly from speech input, identifying semantic relationships between predicates and corresponding arguments in spoken utterances without intermediate text transcription.
Following the end-to-end framework of~\citet{DBLP:conf/acl/ChenLZZ24}, we assume discrete sets of predicates $\mathcal{P}$, arguments $\mathcal{A}$, and semantic role labels $\mathcal{R}$, consistent with the general SRL formulation in \S\ref{sec:formal}.
\begin{compactitem}
\item Given a speech signal $\mathcal{X}$, the task involves identifying predicates $p_i \in \mathcal{P}$ and their associated arguments $a_{i,j} \in \mathcal{A}$.
\item Each predicate-argument pair $(p_i, a_{i,j})$ is assigned a semantic role label $r_{i,j} \in \mathcal{R}$, where $r_{i,j}$ indicates the relationship between predicate $p_i$ and argument $a_{i,j}$.
\item A predicate-argument structure is defined as $\text{PAS}_i = \{(a_{i,j}, r_{i,j}) \mid j \in [1, |\mathcal{A}_i|]\}$, where $\mathcal{A}_i$ represents the set of arguments associated with predicate $p_i$.
\item A complete speech SRL output is represented as $\text{SRL}(\mathcal{X}) = \{(p_i, \text{PAS}_i) \mid i \in [1, |\mathcal{P}|]\}$, capturing all semantic relationships in the spoken utterance.
\end{compactitem}
The Speech SRL task jointly optimizes automatic speech recognition and SRL to predict the complete semantic structure $\text{SRL}(\mathcal{X})$, leveraging both acoustic and linguistic features for robust semantic understanding from speech.
Unlike textual SRL where argument boundaries correspond to token-level spans, speech SRL must resolve argument boundaries from continuous acoustic signals, where word boundaries are not explicitly marked and ASR errors may corrupt predicate-argument alignment.

\tikzstyle{my-box}=[
    rectangle,
    draw=gray,
    rounded corners,
    text opacity=1,
    minimum height=1.5em,
    minimum width=5em,
    inner sep=2pt,
    align=center,
    fill opacity=.5,
    line width=0.8pt,
]
\tikzstyle{leaf}=[my-box, minimum height=1.5em,
    fill=pink!10, text=black, align=left,font=\small,
    inner xsep=2pt,
    inner ysep=4pt,
    line width=0.8pt,
]

\definecolor{c1}{RGB}{93,191,237} 
\definecolor{c2}{RGB}{237,110,106} 
\definecolor{c3}{RGB}{240,154,69} 
\definecolor{c4}{RGB}{8,153,68} 
\definecolor{c5}{RGB}{205,180,243} 
\definecolor{c6}{RGB}{97,218,184} 

\begin{figure*}[!pt]
    \centering
    \resizebox{0.9\textwidth}{!}{
        \begin{forest}
            forked edges,
            for tree={
                grow=east,
                reversed=true,
                anchor=base west,
                parent anchor=east,
                child anchor=west,
                base=center,
                font=\large,
                rectangle,
                draw=gray,
                rounded corners,
                align=left,
                text centered,
                minimum width=4em,
                edge+={darkgray, line width=1pt},
                s sep=3pt,
                inner xsep=2pt,
                inner ysep=3pt,
                line width=0.8pt,
                ver/.style={rotate=90, child anchor=north, parent anchor=south, anchor=center},
            },
            where level=1{text width=12em,font=\normalsize,}{},
            where level=2{text width=12em,font=\normalsize,}{},
            where level=3{text width=10em,font=\normalsize,}{},
            where level=4{text width=35em,font=\normalsize,}{},
            where level=5{text width=18em,font=\normalsize,}{},
            [
                \textbf{Semantic Role Labeling}, ver, line width=0.7mm
                [
                    \textbf{Benchmarks}, fill=c1!60, draw=c1, line width=0mm
                    [
                        \textbf{FrameNet}, fill=c1!60, draw=c1, line width=0mm, edge={c1}
                        [
                            FrameNet \citep{DBLP:conf/acl/BakerFL98};
                            HuRIC \citep{DBLP:journals/ai/VanzoCBBN20}, leaf, text width=32em, draw=c1, line width=0.7mm, edge={c1}
                        ]
                    ]
                    [
                        \textbf{PropBank}, fill=c1!60, line width=0mm, edge={c1}
                        [
                            PropBank \citep{DBLP:journals/coling/PalmerKG05};
                            Universal PropBank \citep{DBLP:conf/lrec/JindalRULNT0022}, leaf, text width=32em, draw=c1, line width=0.7mm, edge={c1}
                        ]
                    ]
                    [
                        \textbf{NomBank}, fill=c1!60, line width=0mm, edge={c1}
                        [
                            NomBank \citep{DBLP:conf/naacl/MeyersRMSZYG04}; , leaf, text width=32em, draw=c1, line width=0.7mm, edge={c1}
                        ]
                    ]
                    [
                        \textbf{CoNLL Shared Tasks}, fill=c1!60, line width=0mm, edge={c1}
                        [
                            CoNLL-2005 \citep{DBLP:conf/conll/CarrerasM05};
                            CoNLL-2009 \citep{DBLP:conf/conll/HajicCJKMMMNPSSSXZ09};\\
                            CoNLL-2012 \citep{DBLP:conf/conll/PradhanMXUZ12}; , leaf, text width=32em, draw=c1, line width=0.7mm, edge={c1}
                        ]
                    ]
                    [
                        \textbf{Non-text}, fill=c1!60, draw=c1, line width=0mm, edge={c1}
                        [
                            Image: V-COCO \citep{DBLP:journals/corr/GuptaM15}; SituNet \citep{DBLP:conf/cvpr/YatskarZF16};\\
                            SWiG \citep{DBLP:conf/eccv/PrattYWFK20}; ExHVV \citep{DBLP:conf/aaai/SharmaASNA023};\\
                            Video: VidStu \citep{DBLP:conf/cvpr/SadhuGYNK21};
                            Speech: AS-SRL \citep{DBLP:conf/acl/ChenLZZ24}, leaf, text width=32em, draw=c1, line width=0.7mm, edge={c1}
                        ]
                    ]
                ]
                [
                    \textbf{Method},
                    fill=c2!60, draw=c2, line width=0mm
                    [
                        \textbf{Statistical}\\ \textbf{Machine Learning}, align=center, fill=c2!60, draw=c2, line width=0mm, edge={c2}
                        [
                            Statistical pipeline \citep{DBLP:conf/acl/GildeaJ00,DBLP:conf/conll/CarrerasM05};\\
                            SVM \citep{DBLP:conf/naacl/PradhanWHMJ04};
                            MaxEnt \citep{DBLP:conf/emnlp/XueP04};\\
                            CRFs \citep{DBLP:journals/coling/ToutanovaHM08,DBLP:conf/acl/WatanabeAM10}
                            ,leaf, text width=32em, draw=c2, line width=0.7mm, edge={c2}
                        ]
                    ]
                    [
                        \textbf{Neural Network}, fill=c2!60, draw=c2, line width=0mm, edge={c2}
                        [
                            CNN \citep{DBLP:journals/jmlr/CollobertWBKKK11};
                            RNN \citep{DBLP:conf/acl/ZhouX15};\\
                            LSTM \citep{DBLP:conf/emnlp/FitzGeraldTG015};
                            GRU \citep{DBLP:conf/paclic/OkamuraTITIIAU18};
                            BiLSTM;\\
                            Transformer-based \citep{DBLP:conf/emnlp/StrubellVAWM18};
                            PLMs \citep{DBLP:conf/coling/DannellsJB24}
                        ,leaf, text width=32em, draw=c2, line width=0.7mm, edge={c2}
                        ]
                    ]
                    [
                        \textbf{Graph-based Methods}, fill=c2!60, draw=c2, line width=0mm, edge={c2}
                          [
                            GCN \citep{DBLP:conf/emnlp/MarcheggianiT17,DBLP:conf/emnlp/LiHCZZLLS18,DBLP:conf/aaai/0001LLJ21};\\
                            GAT \citep{DBLP:conf/iclr/VelickovicCCRLB18}
                            , leaf, text width=32em, draw=c2, line width=0.7mm, edge={c2}
                          ]
                    ]
                    [
                        \textbf{GLM \& LLM}, fill=c2!60, draw=c2, line width=0mm, edge={c2}
                        [
                            HMM \citep{DBLP:conf/semeval/ThompsonPA04};
                            Seq2seq \citep{DBLP:conf/rep4nlp/DazaF18,DBLP:conf/emnlp/DazaF19};\\
                            LLMs \citep{DBLP:journals/corr/abs-2306-09719,li-etal-2025-llms-also,DBLP:conf/icic/ChengYWLYZTXH24}
                            ,leaf, text width=32em, draw=c2, line width=0.7mm, edge={c2}
                        ]
                    ]
                ]
                [
                    \textbf{Paradigm}, fill=c3!60, draw=c3, line width=0mm
                    [
                        \textbf{Span-based}, fill=c3!60, draw=c3, line width=0mm, edge={c3}
                        [                                           \cite{DBLP:conf/conll/HaciogluPWMJ04,DBLP:journals/tacl/TackstromG015,DBLP:conf/emnlp/FitzGeraldTG015};\\
                        \cite{DBLP:conf/acl/HeLLZ17a,DBLP:conf/acl/HeLLZ18,DBLP:conf/emnlp/OuchiS018,DBLP:conf/coling/ZhouXLZHZ22};
                        , leaf, text width=32em, draw=c3, line width=0.7mm, edge={c3}
                        ]
                    ]
                    [
                        \textbf{Dependency-based}, fill=c3!60, draw=c3, line width=0mm, edge={c3}
                        [
                        \cite{DBLP:conf/acl/PradhanWHMJ05,DBLP:conf/acl/SwansonG06,DBLP:conf/acl/HeLLZ18};\\
                        \cite{DBLP:conf/emnlp/JohanssonN08,DBLP:conf/emnlp/MarcheggianiT17};\\
                        \cite{DBLP:conf/coling/CaiHLZ18,DBLP:conf/emnlp/StrubellVAWM18,DBLP:conf/aaai/LiHZZZZZ19,DBLP:conf/emnlp/LiZWP20,DBLP:conf/emnlp/ShiMI20};
                        , leaf, text width=32em, draw=c3, line width=0.7mm, edge={c3}
                        ]
                    ]
                ]
                [
                    \textbf{Syntax Feature Modeling}, fill=c4!60, draw=c4, line width=0mm
                    [
                        \textbf{Syntax-aided SRL}, fill=c4!60, draw=c4, line width=0mm, edge={c4}
                        [
                        \cite{DBLP:conf/acl/GildeaP02,DBLP:conf/acl/PradhanWHMJ05,DBLP:conf/conll/KoomenPRY05};\\
                        \cite{DBLP:conf/ijcai/PunyakanokRY05,DBLP:journals/coling/PunyakanokRY08,DBLP:journals/coling/ToutanovaHM08,DBLP:conf/acl/ZhaoHLB18};\\
                        \cite{DBLP:conf/emnlp/MarcheggianiT17,DBLP:conf/emnlp/StrubellVAWM18,DBLP:conf/acl/WangJWSW19};\\
                        \cite{DBLP:conf/acl/FeiWRLJ21,DBLP:conf/acl/ChenHM22};
                         , leaf, text width=32em, draw=c4, line width=0.7mm, edge={c4}
                        ]
                    ]
                    [
                        \textbf{Syntax-free SRL}, fill=c4!60, draw=c4, line width=0mm, edge={c4}
                        [
                        \cite{DBLP:journals/jmlr/CollobertWBKKK11,DBLP:conf/acl/ZhouX15,DBLP:conf/conll/MarcheggianiFT17};\\
                        \cite{DBLP:conf/acl/HeLLZ17a,DBLP:conf/emnlp/OuchiS018,DBLP:conf/coling/CaiHLZ18,DBLP:conf/aaai/LiHZZZZZ19};\\
                        \cite{DBLP:conf/emnlp/StrubellVAWM18,DBLP:journals/coling/LiZHC21};
                        , leaf, text width=32em, draw=c4, line width=0.7mm, edge={c4}
                        ]
                    ]
                ]
                [
                    \textbf{Scenario}, fill=c5!60, draw=c5, line width=0mm
                    [
                        \textbf{Beyond Single Sentence}, fill=c5!60, draw=c5, line width=0mm, edge={c5}
                        [
                        Discourse: \cite{fillmore2001frame,burchardt2005building,DBLP:conf/ijcnlp/DoBM17};\\
\cite{DBLP:journals/lre/RuppenhoferLSM13,DBLP:journals/coling/RothF15,DBLP:conf/acl/LaparraR13};\\
                        Dialogue:
\cite{DBLP:journals/taslp/XuWSZSY21,DBLP:conf/icmlc2/HeWLSC21,DBLP:conf/naacl/WuT0LWS22,DBLP:conf/ijcai/0001WZRJ22};\\
                        \cite{DBLP:journals/taslp/WuXS24}
                        , leaf, text width=32em, draw=c5, line width=0.7mm, edge={c5}
                        ]
                    ]
                    [
                        \textbf{Multilingual}, fill=c5!60, draw=c5, line width=0mm, edge={c5}
                        [
                        Chinese \citep{DBLP:conf/acl-sighan/XueP03};
                        German \citep{DBLP:conf/acl/PadoL06};\\
                        French \citep{DBLP:conf/taln/PadoP07,DBLP:conf/acl/PlasMH11};\\
                        Italian \citep{DBLP:conf/lrec/TonelliP08,DBLP:conf/cicling/BasiliCCCM09,DBLP:conf/cicling/AnnesiB10};\\
                        Global Approach: \cite{DBLP:conf/coling/PlasAC14,DBLP:conf/acl/KozhevnikovT13};\\
                        \cite{DBLP:conf/emnlp/DazaF19,DBLP:conf/acl/FeiZJ20,DBLP:conf/emnlp/DeviantiM24};\\
                        \cite{DBLP:conf/acl/MulcaireSS18,DBLP:conf/emnlp/JohannsenAS15,DBLP:conf/acl/FeiZJ20,DBLP:journals/taslp/FeiZLJ20}
                        , leaf, text width=32em, draw=c5, line width=0.7mm, edge={c5}
                        ]
                    ]
                    [
                        \textbf{Other Modalities}, fill=c5!60, draw=c5, line width=0mm, edge={c5}
                        [
                        Image: VSRL \citep{DBLP:journals/corr/GuptaM15};
                        SituNet \citep{DBLP:conf/cvpr/YatskarZF16};\\
                        GSR \citep{DBLP:conf/eccv/PrattYWFK20};
                        VGG+RNN \citep{DBLP:conf/iccv/MallyaL17};\\
                        GNN \citep{DBLP:conf/iccv/LiTLJUF17,DBLP:conf/iccv/SuhailS19};\\
                        Noisy Caption \citep{DBLP:conf/emnlp/SilbererP18};\\
                        Query-based VR \citep{DBLP:conf/cvpr/CoorayCL20};
                        ViT \citep{DBLP:conf/bmvc/ChoYLK21};\\
                        Video: VidSRL \citep{DBLP:conf/cvpr/SadhuGYNK21};
                        Grounding \citep{DBLP:conf/nips/KhanJT22};\\
                        State Tracking \citep{DBLP:conf/aaai/YangLZ0JC23};
                        Spatio-Temporal \citep{DBLP:conf/mm/Zhao00LZWZC23};\\
                        Speech: End-to-End SpeechSRL \citep{DBLP:conf/acl/ChenLZZ24}
                        , leaf, text width=32em, draw=c5, line width=0.7mm, edge={c5}
                        ]
                    ]
                ]
                [
                    \textbf{Applications}, fill=c6!60, draw=c6, line width=0mm
                    [
                        \textbf{NLP Tasks},  fill=c6!60, draw=c6, line width=0mm, edge={c6}
                        [
                        Information Extraction \citep{DBLP:conf/ranlp/EvansO19};\\
                        Machine Translation \citep{DBLP:conf/acl/ShiLRFLZSW16,DBLP:conf/naacl/MarcheggianiBT18};\\
                        Question Answering \citep{DBLP:conf/emnlp/BerantCFL13,DBLP:conf/emnlp/HeLZ15,DBLP:conf/acl/YihRMCS16};\\
                        Summarization \citep{DBLP:journals/asc/KhanSK15,DBLP:journals/ipm/MohamedO19}
                        , leaf, text width=32em, draw=c6, line width=0.7mm, edge={c5}
                        ]
                    ]
                    [
                        \textbf{Language Modeling},  fill=c6!60, draw=c6, line width=0mm, edge={c6}
                        [
                        BERT Enhancement \citep{DBLP:conf/aaai/0001WZLZZZ20};\\
                        Dialog Rewriting \citep{DBLP:conf/emnlp/XuTSWZSY20};\\
                        ACO Integration \citep{DBLP:journals/jksucis/Onan23a};\\
                        Feature Extraction \citep{DBLP:conf/coling/ZouG0LCLAS24}
                        , leaf, text width=32em, draw=c6, line width=0.7mm, edge={c6}
                        ]
                    ]
                    [
                        \textbf{Robotics},  fill=c6!60, draw=c6, line width=0mm, edge={c6}
                        [
                        NL Instruction Mapping \citep{DBLP:conf/ecai/BastianelliCCBN14};\\
                        Dependency Features \citep{DBLP:journals/ieeejas/LuC17};\\
                        Scene Understanding \citep{DBLP:conf/cvpr/0016J0021}
                        , leaf, text width=32em, draw=c6, line width=0.7mm, edge={c6}
                        ]
                    ]
                    [
                        \textbf{Advanced Embodied AI},  fill=c6!60, draw=c6, line width=0mm, edge={c6}
                        [
                        Real-time Environmental SRL \citep{DBLP:conf/acl-jssp/BastianelliCCB13};\\
                        Visual-Linguistic Framework \citep{DBLP:conf/naacl/YangGLXZC16};\\
                        Language-Independent Command Interpretation \citep{DBLP:journals/ai/VanzoCBBN20};\\
                        SRL-based Action Sequence Validation \citep{DBLP:journals/kbs/ZhangTZD21}
                        , leaf, text width=32em, draw=c6, line width=0.7mm, edge={c6}
                        ]
                    ]
                ]
            ]
        \end{forest}
    }
    \caption{Comprehensive taxonomy of SRL research.}
    \label{fig: taxonomy}
\end{figure*}

\subsection{Discrimination}
\subsubsection{SRL vs. AMR}
AMR and SRL are two foundational yet distinct approaches to semantic parsing in natural language processing.
SRL primarily focuses on identifying predicate-argument structures within sentences, directly labeling semantic roles on text spans or syntactic heads and maintaining a close connection to surface syntax.
In contrast, AMR provides a more abstract, graph-based semantic representation, where nodes correspond to semantic concepts and edges encode a wide range of semantic relations—including modality, negation, and cross-sentence links—thus capturing deeper meaning beyond surface form~\citep{DBLP:conf/acllaw/BanarescuBCGGHK13}, as illustrated in Figure~\ref{fig:1}.

To facilitate comparison and interoperability among different semantic representations, the meaning representation parsing (MRP) shared task~\citep{DBLP:conf/conll/OepenAHHKOXCSU19,DBLP:conf/conll/OepenAABHHLOXZ20} introduced a unified evaluation framework.
The MRP framework organizes five major semantic representations into two main categories (``flavors'') based on their anchoring to surface text:
\footnote{Notably, this differs from the MRP 2019 task, which also included Flavor 0 representations (e.g., DM and PSD with direct lexical correspondences), later removed in 2020 to lower the participation barrier.}
\begin{compactitem}
    \item Flavor 1 (including EDS, PTG, and UCCA): allows flexible anchoring, where nodes may correspond to arbitrary spans of the sentence.
    \item Flavor 2 (represented by AMR and DRG): provides fully abstract representations, with nodes not explicitly anchored to surface tokens.
\end{compactitem}

\subsubsection{SRL vs. Event Extraction}
Event extraction (EE) is another task closely related to SRL.
As shown in Figure~\ref{fig:1}, EE involves identifying event triggers (typically verbs, nouns, or phrases evoking an event) and extracting their associated arguments or participants from the text.
Both SRL and EE aim to uncover the relational structure of text, identifying who or what is involved, how, and under what circumstances.
They share the core principle of detecting ``participants'' and assigning them specific roles.

However, there are key differences.
SRL provides a domain-independent, predicate-centric analysis of semantic structure, aiming for broad linguistic coverage.
In contrast, EE is typically domain- or scenario-specific, focusing on a predefined set of event types relevant to particular domains, such as conflict, legal, and biomedical domains.
EE often seeks to capture a more comprehensive event structure, potentially spanning multiple sentences or documents.
A primary distinction lies in their labeling schemes.
EE uses descriptive, natural-language role names such as ``BUYER'' and ``PLACE'', while SRL employs generalized, abstract labels such as ``ARG0'' and ``ARGM-LOC''.
Furthermore, SRL role definitions may be inconsistent or insufficiently specific for direct use as event roles or frames in EE.
For example, in ACE-style event extraction, the ``Attack'' event type requires a dedicated ``Attacker'' role, whereas PropBank assigns the same participant the generic label ``ARG0'', which denotes the agent of any predicate regardless of event type.
Similarly, the ``ARG2'' label in PropBank conflates instrument, benefactive, and attribute readings depending on the predicate, making it difficult to map directly onto the fine-grained role distinctions required by domain-specific EE schemas.
In FrameNet-based SRL, frame elements are frame-specific and carry richer semantic constraints, yet the frame inventory does not align with the event ontologies used in EE benchmarks such as ACE or ERE, which limits direct cross-task transfer.
Additionally, standard SRL datasets rarely annotate distant arguments, those lacking explicit syntactic links, whereas EE may require capturing such long-range dependencies.

In summary, while SRL and EE are intertwined tasks that both seek to extract structured knowledge from text, each has a distinct focus and set of conventions.
Understanding their respective strengths and limitations is essential for designing robust semantic parsing systems and for their effective application in real-world scenarios.

\section{Overview of the SRL Taxonomy}
\label{Taxonomy of SRL Methodology}

In recent years, SRL has undergone significant evolution, reflecting increasing complexity and a broadening scope of applications.
To systematically capture this progress, we present a comprehensive taxonomy that organizes SRL research along six key dimensions, as illustrated in Figure~\ref{fig: taxonomy}.
\begin{compactitem}
    \item First, we introduce major benchmarks and shared tasks, including FrameNet, PropBank, CoNLL shared tasks, and datasets for non-text modalities, which have played a pivotal role in standardizing SRL evaluation and driving progress in the field.
    \item Second, we review the core methodological advances, covering statistical machine learning, neural networks, graph-based methods, and the recent integration of generative and large language models (GenLMs \& LLMs), which collectively underpin the development of modern SRL systems.
    \item Third, we discuss different task paradigms, such as span-based and dependency-based approaches, which define the granularity and structure of semantic role annotation and inference.
    \item Fourth, we examine syntax feature modeling strategies, distinguishing between syntax-aided and syntax-free SRL, and analyzing the conditions under which syntactic information provides measurable gains over purely data-driven approaches.
    \item Fifth, we explore diverse scenarios in which SRL is deployed, including discourse-level and conversational SRL (beyond single sentence), multilingual and cross-lingual transfer, as well as the extension to other modalities such as images, videos, and speech.
    \item Finally, we summarize the broadening landscape of applications enabled by SRL, ranging from traditional NLP tasks (e.g., information extraction, machine translation, question answering, summarization) to emerging domains such as language modeling, robotics, and advanced embodied AI.
\end{compactitem}

\section{SRL Benchmarks}
\label{Benchmarks}

\subsection{Datasets}
\label{datasets}

\begin{table*}[ht]
    \fontsize{9}{10}\selectfont
    \setlength{\tabcolsep}{1.3mm}
    \begin{center}
    \resizebox{1\textwidth}{!}{
    \begin{tabular}{lccccc}
    \Xhline{0.08em}
    \rowcolor{blue!15}
    \multicolumn{6}{c}{\textit{\textbf{Textual SRL}}}\\
    \bf Dataset & \bf Style & \bf Corpus & \bf Scale & \bf Scenario & \bf Languages \\ \hline
    FrameNet \citep{DBLP:conf/acl/BakerFL98} & FrameNet & British National Corpus & \textgreater 200,000 & sentence & En \\
    \rowcolor{gray!15}
    PropBank \citep{DBLP:journals/coling/PalmerKG05} & PropBank & PTB, robotic surgery books & \textgreater 100,000 & sentence & En \\
    
    \multirow{2}{*}{CoNLL 2005 \citep{DBLP:conf/conll/CarrerasM05}} & \multirow{2}{*}{PropBank} & Wall Street Journal in PTB, & \multirow{2}{*}{44,020} & \multirow{2}{*}{sentence} & \multirow{2}{*}{En}\\
    & & Brown corpus in PTB & & & \\
    
    \cellcolor{gray!15} & \cellcolor{gray!15} & \cellcolor{gray!15}PTB 3, BBN's NE, PropBank 1, NomBank & \cellcolor{gray!15}41,678 & \cellcolor{gray!15}sentence & \cellcolor{gray!15}En \\
    \cellcolor{gray!15} & \cellcolor{gray!15} & \cellcolor{gray!15}CPB & \cellcolor{gray!15}24,833 & \cellcolor{gray!15}sentence & \cellcolor{gray!15}Zh \\
    \cellcolor{gray!15} & \cellcolor{gray!15} & \cellcolor{gray!15}AnCora & \cellcolor{gray!15}31,116 & \cellcolor{gray!15}sentence & \cellcolor{gray!15}Ca,Es \\
    \cellcolor{gray!15} & \cellcolor{gray!15} & \cellcolor{gray!15}Prague Dependency Treebank 2.0 & \cellcolor{gray!15}42,940 & \cellcolor{gray!15}sentence & \cellcolor{gray!15}Cs \\
    \cellcolor{gray!15} & \cellcolor{gray!15} & \cellcolor{gray!15}SALSA & \cellcolor{gray!15}38,020 & \cellcolor{gray!15}sentence & \cellcolor{gray!15}De \\
    \multirow{-6}{*}{\cellcolor{gray!15}CoNLL 2009 \citep{DBLP:conf/conll/HajicCJKMMMNPSSSXZ09}}& \multirow{-6}{*}{\cellcolor{gray!15}PropBank} & \cellcolor{gray!15}Kyoto University & \cellcolor{gray!15}4,893 & \cellcolor{gray!15}sentence & \cellcolor{gray!15}Ja \\
    NomBank \citep{DBLP:conf/naacl/MeyersRMSZYG04} & NomBank & NOMLEX-PLUS,PTB & 114,576 & sentence & En\\

    \cellcolor{gray!15} & \cellcolor{gray!15} & \cellcolor{gray!15} & \cellcolor{gray!15}94,269 & \cellcolor{gray!15}sentence & \cellcolor{gray!15}En \\
    \cellcolor{gray!15} & \cellcolor{gray!15} & \cellcolor{gray!15} & \cellcolor{gray!15}47,042 & \cellcolor{gray!15}sentence & \cellcolor{gray!15}Zh \\
    \multirow{-3}{*}{\cellcolor{gray!15}CoNLL 2012 \citep{DBLP:conf/conll/PradhanMXUZ12}}& \multirow{-3}{*}{\cellcolor{gray!15}PropBank} & \multirow{-3}{*}{\cellcolor{gray!15}OntoNotes} & \cellcolor{gray!15}9,395 & \cellcolor{gray!15}sentence & \cellcolor{gray!15}Ar \\
    
    HuRIC \citep{DBLP:journals/ai/VanzoCBBN20} & FrameNet & Human Robot Interaction & 897 & command & En,It\\
    \rowcolor{gray!15}
    ConSD \citep{DBLP:journals/talip/LiZZPH22} & PropBank & PTB,PropBank,NomBank & 44,020 & sentence & En\\
    
    \multirow{6}{*}{Universal PropBank \citep{DBLP:conf/lrec/JindalRULNT0022}} & \multirow{6}{*}{PropBank} & \multirow{6}{*}{UD} & \multirow{6}{*}{3,860,000} & \multirow{6}{*}{sentence} & Cs,De,El.Es \\
    & & & & & Fi,Fr,Hi,Hu \\
    & & & & & Id,It,Ja,Ko \\
    & & & & & Mr,Nl,Pr,Pt \\
    & & & & & Ro,Ru,Ta,Te \\
    & & & & & Uk,Vi,Zh \\ 
    \rowcolor{gray!15}
    NounAtlas \citep{navigli-etal-2024-nounatlas} & FrameNet & WordNet, SemCor & 10,086 & sentence & En\\

    \specialrule{.2em}{.05em}{0.05em} 
    \rowcolor{blue!15}
    \multicolumn{6}{c}{\textit{\textbf{Visual SRL}}} \\
    \bf Dataset & \bf Style & \bf Corpus & \bf Scale & \bf Scenario & \bf Languages \\ \hline
    V-COCO \citep{DBLP:journals/corr/GuptaM15} & & COCO & 10,000 & image & En \\
    \rowcolor{gray!15}
    SituNet \citep{DBLP:conf/cvpr/YatskarZF16} & & Google image & 126,102 & image & En \\
    
    SWiG \citep{DBLP:conf/eccv/PrattYWFK20} & & Google image & 126,102 & image & En \\
    \rowcolor{gray!15}
    ExHVV \citep{DBLP:conf/aaai/SharmaASNA023} & & HVVMemes & 4,680 & image & En \\
    
    HuRIC-G \citep{hromei-etal-2025-grounded} & FrameNet & HuRIC & \textgreater 11,000 & \makecell[c]{image \\ command} & En \\
    
    \specialrule{.2em}{.05em}{0.05em} 
    \rowcolor{blue!15}
    \multicolumn{6}{c}{\textit{\textbf{Video SRL}}} \\
    \bf Dataset & \bf Style & \bf Corpus & \bf Scale & \bf Scenario & \bf Languages \\ \hline
    VidStu \citep{DBLP:conf/cvpr/SadhuGYNK21} & & MovieClips & 3,037 & video & En \\ 
    \specialrule{.2em}{.05em}{0.05em} 
    \rowcolor{blue!15}
    \multicolumn{6}{c}{\textit{\textbf{Speech SRL}}} \\
    \bf Dataset & \bf Style & \bf Corpus & \bf Scale & \bf Scenario & \bf Languages \\ \hline
    AS-SRL \citep{DBLP:conf/acl/ChenLZZ24} & & AISHELL-1,CPB  & 9,000 & speech & Zh \\
    \bottomrule
    \end{tabular}
    }
    \end{center}
    \caption{
    Overview of SRL benchmarks across different modalities. The table is organized into four modality-specific sections: Textual SRL, Visual SRL, Video SRL, and Speech SRL. Within each section, the Scale column reports the number of instances, where the unit of instances varies by modality, including sentences for textual datasets, images for visual datasets, videos for video datasets, and utterances for speech datasets.
    }
    \label{tab:bench}
    \end{table*}

There are several important public corpora related to the text-only SRL task: FrameNet, PropBank, NomBank, CoNLL, as well as more recent resources such as VerbAtlas and NounAtlas.
In Table~\ref{tab:bench}, we list the mainstream SRL benchmarks and the statistics.

\subsubsection{FrameNet}
The FrameNet dataset~\citep{DBLP:conf/acl/BakerFL98} is a large lexical resource that provides semantic annotations based on the Frame Semantics theory~\citep{lowe-1997-frame,fillmore2006frame}.
Frame Semantics, developed by~\citet{fillmore1976frame} from his earlier Case Grammar work, holds that word meanings are intelligible only against the structured background knowledge, or frames, that they evoke, a view now central to cognitive linguistics.
The theoretical grounding of Frame Semantics draws on converging evidence from cognitive psychology and linguistics~\citep{marshall1995schemas,wagoner2013bartlett} that human conceptual knowledge is organized around event schemas and participant roles, lending the framework ontological coherence without fully dismissing feature-based accounts of meaning.
This foundation directly motivates FrameNet, where each frame encodes a conceptually coherent scenario and its associated frame elements define the roles that participants characteristically fill, providing the ontological backbone for SRL.
It consists of frames, which represent conceptual structures of events, situations, or activities, and frame elements, which are the roles participants play within these frames (such as Agent, Theme, Recipient, etc.).
Each word (verbs, nouns, and adjectives) is associated with one or more frames and the roles it fills in context.
FrameNet's annotations help in tasks like SRL, machine translation, and information extraction by offering rich, structured semantic information that connects syntax to meaning.
It is widely used in computational linguistics, with resources available for multiple languages and formats for research and application development.
Currently, FrameNet contains over 13,000 lexical units annotated with more than 1,200 hierarchically related semantic frames~\citep{ruppenhofer2016framenet}.
However, FrameNet's frame-specific role labels limit cross-frame generalizability, and its coverage of the verbal lexicon remains incomplete.
Inspired by FrameNet's frame semantics, \citet{di-fabio-etal-2019-verbatlas} introduced VerbAtlas, a manually-crafted resource that clusters all 13,767 WordNet verbal synsets into 466 semantically-coherent frames with a compact, cross-frame inventory of 25 explicit semantic roles, achieving full coverage of the English verbal lexicon while preserving role interpretability.
Building on VerbAtlas, \citet{navigli-etal-2024-nounatlas} further extended this framework to nominal predicates by introducing NounAtlas, which maps over 10,000 nominal synsets from WordNet into the same VerbAtlas frames, enabling unified SRL over both verbal and nominal predicates within a single coherent inventory.

\subsubsection{PropBank}
The PropBank (Proposition Bank) dataset is a linguistic corpus first developed by~\citet{DBLP:journals/coling/PalmerKG05} that provides semantic role annotations for the Wall Street Journal portion of the Penn Treebank, focusing on verb predicates and their arguments.
PropBank enhances syntactic treebanks by adding semantic information, where each verb is annotated with specific sense numbers and associated rolesets that describe the arguments a verb can take, including numbered core arguments (ARG0-ARG5) whose semantic interpretations (e.g., Agent, Theme, or Goal) are defined on a per-predicate basis within each roleset, as well as modifier arguments (ArgM) for temporal, locative, and other adjunct information.
For example, \textit{leave.01} corresponds to the sense ``move away from'' while \textit{leave.02} corresponds to ``give'', and the semantic interpretation of each numbered argument such as ARG1 differs accordingly across these senses.
This sense-dependent nature of argument roles means that predicate sense identification is a prerequisite for accurately determining the semantic content of each argument, and it constitutes one of the four fundamental subtasks of SRL as described in \S\ref{Paradigm Modeling in SRL}.
The corpus has been extended to multiple languages, including Arabic, Chinese, Finnish, Hindi, Portuguese, and Turkish.

\subsubsection{NomBank}
NomBank~\citep{DBLP:conf/naacl/MeyersRMSZYG04} extends PropBank annotation to nominal predicates, providing argument structure annotation for approximately 5,000 common nouns in the Penn Treebank II corpus.
The project annotates verbal nominalizations (e.g., ``destruction'', ``investment''), adjectival nominalizations, and relational nouns with numbered arguments (ARG0, ARG1, etc.) following PropBank's roleset approach.
NomBank enables pattern generalization across related constructions, allowing systems to recognize equivalent semantic relationships in expressions like ``IBM appointed John'', ``IBM's appointment of John'', and ``John is the IBM appointee''.
The resource contains annotations for approximately 240,000 noun instances, making it valuable for comprehensive SRL systems handling both verbal and nominal predicates.

\subsubsection{CoNLL Shared Task}
The competitive open tasks in the Conference on Computational Natural Language Learning (CoNLL) held in 2005, 2009, and 2012 are related to the SRL task.
The CoNLL 2005 dataset consists of English texts annotated with semantic roles based on both PropBank and FrameNet resources, providing a comprehensive evaluation framework for SRL systems.
In CoNLL 2009, the dataset was expanded to six languages, including Catalan, Chinese, Czech, German, Japanese, and Spanish, with annotations following the PropBank-style approach.
The CoNLL 2012 data is built on the OntoNotes v5.0 corpus, containing multi-task annotations, including Part-of-Speech, syntactic parsing, Named Entity Recognition, coreference resolution, and SRL based on PropBank-style annotations.

\subsubsection{Annotation Practices, Costs, and Crowdsourcing Challenges}
\label{annotation_practices}

The construction of high-quality SRL corpora depends critically on the recruitment of qualified annotators, and this challenge has been addressed in different ways across the literature.

Major SRL benchmarks have historically relied on trained linguistic experts rather than naive annotators.
FrameNet was developed over more than two decades at the International Computer Science Institute at Berkeley, with annotation carried out by computational linguists who received extensive training in Frame Semantics theory~\citep{DBLP:conf/acl/BakerFL98,ruppenhofer2016framenet}.
PropBank annotation relied on trained annotators working with detailed predicate-specific roleset guidelines, and inter-annotator agreement for core argument roles has been reported at approximately 0.9 kappa, though agreement drops noticeably for adjunct roles such as ARGM-CAU and ARGM-DIS~\citep{DBLP:journals/coling/PalmerKG05}.
The CoNLL shared task datasets extended these conventions to multiple languages through cross-institutional collaborations, each contributing annotators with native-language expertise, though harmonizing guidelines across typologically diverse languages introduced additional consistency challenges~\citep{DBLP:conf/conll/HajicCJKMMMNPSSSXZ09}.

To reduce annotation costs, researchers have explored crowdsourcing platforms such as Amazon Mechanical Turk \cite{mohammad-etal-2014-semantic}.
However, the abstract nature of PropBank role labels and the sense-dependent interpretation of numbered arguments make SRL particularly difficult for non-expert workers.
Question-answer driven SRL~\citep{DBLP:conf/emnlp/HeLZ15} addresses this by reformulating role labeling as a natural language question answering task, yielding reasonable inter-annotator agreement among non-experts, though the resulting labels require post-hoc mapping to standard role inventories.
Crowdsourcing also faces practical limitations in multilingual settings, where the available worker pool for low-resource languages is too small to support large-scale annotation campaigns.

Expert SRL annotation is substantially more costly than simpler NLP labeling tasks, requiring several minutes per sentence and running into hundreds of thousands of dollars for corpora of tens of thousands of sentences.
These cost pressures have motivated annotation-efficient approaches such as active learning, semi-supervised training, and cross-lingual projection, and they underscore the importance of releasing annotation tools and guidelines alongside completed corpora to support future efforts.

\subsection{Evaluations}

Evaluation metrics for SRL differ slightly between span-based and dependency-based formulations.
Nevertheless, both approaches adopt precision, recall, and F$_1$ score as standard metrics, as established in the CoNLL shared tasks~\citep{DBLP:conf/conll/CarrerasM05,DBLP:conf/conll/SurdeanuJMMN08}.

In \textbf{span-based SRL}, a prediction is considered correct only when both the predicted argument span boundaries and the semantic role label exactly match the gold annotation~\citep{DBLP:journals/coling/PalmerKG05}.
The metrics are defined as:
\begin{align*}
    Precision = \frac{|C|}{|P|},~~~ Recall = \frac{|C|}{|G|}, \\ F_1 = \frac{2 \times Precision \times Recall}{Precision + Recall},
\end{align*}
where $C$ is the set of correctly predicted arguments, $P$ is the set of all predicted arguments, and $G$ is the set of gold standard arguments.

In \textbf{dependency-based SRL}, as formalized in the CoNLL-2008 shared task~\citep{DBLP:conf/conll/SurdeanuJMMN08}, evaluation is conducted over predicate-argument-role triples.
The corresponding metrics are:
\begin{align*}
    Precision = \frac{C_{par}}{P_{par}},~~~ Recall = \frac{C_{par}}{G_{par}}, \\ F_{1} = \frac{2 \times Precision \times Recall}{Precision + Recall},
\end{align*}
where $C_{par}$ is the number of correctly identified $(p,a,r)$ tuples, $P_{par}$ is the number of predicted tuples, and $G_{par}$ is the number of gold tuples.
Here, $(p,a,r)$ denotes a predicate, an argument head word, and its role label.

To ensure standardized evaluation, the community uses several scorers.
The CoNLL-2005 scorer~\citep{DBLP:conf/conll/CarrerasM05} is standard for span-based SRL on English datasets.
The SemEval-2007 scorer\footnote{\href{http://www.ark.cs.cmu.edu/SEMAFOR/eval/}{http://www.ark.cs.cmu.edu/SEMAFOR/eval/}}~\citep{DBLP:conf/semeval/BakerEE07} evaluates frame SRL on FrameNet datasets with fine-grained matching.
The CoNLL-2009 scorer\footnote{\href{https://ufal.mff.cuni.cz/conll2009-st/scorer.html}{https://ufal.mff.cuni.cz/conll2009-st/scorer.html}}~\citep{DBLP:conf/conll/HajicCJKMMMNPSSSXZ09} handles dependency-based SRL with multilingual support.

While F$_1$ score has served as the dominant evaluation metric in SRL research, its adequacy as a measure of semantic understanding warrants critical examination.
F$_1$ treats all role types equally, assigning identical weight to core arguments such as ``ARG0'' and ``ARG1'' and to adjunct modifiers such as ``ARGM-CAU'' and ``ARGM-DIS'', despite the fact that these role types differ substantially in their linguistic complexity and their contribution to sentence meaning.
A system that achieves high F$_1$ by accurately predicting frequent and structurally simple roles may still fail to capture the deeper semantic relationships that are most relevant for downstream applications.
Furthermore, the exact-match criterion used in span-based evaluation penalizes predictions that are semantically near-correct but differ slightly in boundary, offering no credit for partially correct spans even when the core semantic content is preserved.
In dependency-based evaluation, the reliance on head-word matching similarly obscures cases where the predicted argument captures the correct semantic participant through a different syntactic realization.

A further limitation of F$_1$ as the sole reporting metric is that it conveys no information about result stability.
The performance figures compiled in this survey are drawn directly from the original publications, each of which reports a single-point score without accompanying variance estimates or confidence intervals.
Since different studies use different random seeds, data preprocessing pipelines, and PLM checkpoints, the observed score differences between systems are often smaller than the variability that would be expected across multiple runs of the same system.
Readers should therefore treat small F$_1$ differences between adjacent entries in the performance tables as indicative rather than conclusive, and cross-system comparisons should be interpreted with this limitation in mind.

Beyond the metric itself, the dominant evaluation paradigm carries several structural biases that deserve attention.
Standard benchmarks such as CoNLL-2005 and CoNLL-2012 are constructed from a limited set of genres and domains, with a substantial portion drawn from edited written text such as newswire and broadcast news, which means that high F$_1$ scores on these datasets may not reflect genuine generalization to informal, spoken, or domain-specific text.
The consistent performance gap between in-domain and out-of-domain test sets, as observed in the Brown corpus results reported in Tables~\ref{tab:result-05} and~\ref{tab:result-09}, confirms that benchmark-driven progress does not always translate to robust semantic understanding across varied text types.
Role label distributions in existing corpora are also highly skewed, with core arguments far outnumbering adjunct roles, which can lead evaluation metrics to favor systems that perform well on common labels while underperforming on rarer but semantically informative ones.
In multilingual and cross-lingual settings, the use of English-centric annotation guidelines introduces additional bias, as role definitions developed for English may not align well with the argument structure conventions of typologically distant languages.
These dataset-level biases carry direct consequences for downstream applications in socially sensitive domains.
When SRL models trained on skewed corpora are deployed in areas such as legal document analysis, clinical text processing, or automated content moderation, the inherited annotation biases can lead to systematically unequal treatment of different demographic groups or linguistic communities~\citep{siddharth2025data}.
Addressing such biases therefore requires not only improved evaluation protocols but also greater attention to data governance and fairness considerations during corpus construction.

A related concern is benchmark saturation on well-established test sets.
On the CoNLL-2005 WSJ test set, the strongest reported systems now exceed 90 F$_1$, and score differences among top systems have narrowed to less than one point.
At this level of performance, the benchmark provides limited signal for distinguishing genuinely superior systems from those that benefit from favorable preprocessing choices or PLM version differences.
This saturation effect motivates the development of more challenging evaluation conditions, including out-of-domain test sets, low-resource language benchmarks, and task-oriented evaluation protocols that measure the downstream utility of predicted predicate-argument structures rather than surface label matching alone.

Despite these standardized scorers, reproducibility remains a notable challenge in SRL research.
Direct comparison across systems is frequently complicated by inconsistencies in data splits, preprocessing pipelines, and input feature settings, such as whether gold or predicted syntactic annotations are used, and whether predicate identification is assumed or performed end-to-end.
Differences in pre-trained language model (PLM) versions further obscure meaningful comparisons.
Because the performance figures reported in this survey are aggregated from heterogeneous experimental setups across independent publications, they should not be treated as the basis for fine-grained ranking of systems.
We therefore encourage future work to release preprocessing scripts and model checkpoints, and to report results under clearly documented experimental conditions, ideally including multiple runs with variance estimates.
Beyond reproducibility, the field would benefit from developing evaluation protocols that go beyond aggregate F$_1$, such as role-type-stratified reporting, out-of-domain evaluation as a standard requirement, and task-oriented metrics that measure the downstream utility of predicted predicate-argument structures directly.

\section{Methods in SRL}
\label{Methods in SRL}

\begin{figure*}[ht]
    \centering
    \includegraphics[width=0.8\linewidth]{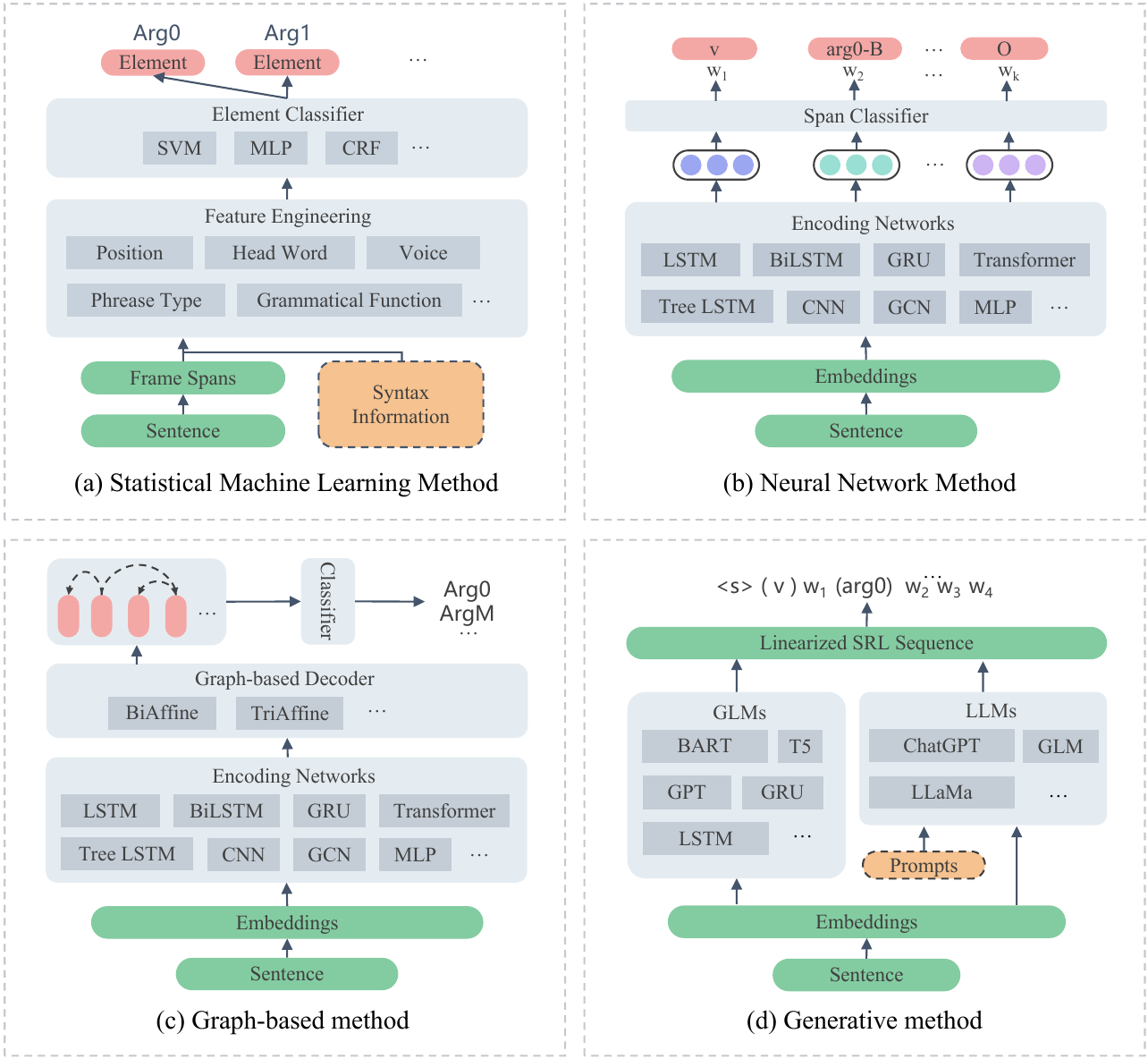}
    \caption{
        Overview of four major SRL task modeling paradigms. (a) Statistical machine learning methods rely on feature engineering and classical classifiers such as SVM, MLP, and CRF to predict semantic roles. (b) Neural network methods employ encoding networks including LSTM, BiLSTM, CNN, GCN, and Transformer to learn representations automatically. (c) Graph-based methods introduce structured decoders such as BiAffine and TriAffine on top of encoding networks to model predicate-argument dependencies explicitly. (d) Generative methods recast SRL as a sequence generation task, leveraging GenLMs such as BART, T5, and GPT, as well as LLMs such as ChatGPT and LLaMA, with prompt-based inputs.
    }
    \label{fig:paradigm}
\end{figure*}

The methodological evolution of SRL can be broadly divided into four major paradigms, as illustrated in Figure~\ref{fig:paradigm}.
Each paradigm transition was driven not merely by the availability of new tools, but by the accumulation of unresolved limitations in the preceding approach.
Statistical methods reached a ceiling imposed by the expressiveness of hand-crafted feature templates and the propagation of parser errors through rigid pipelines.
Neural methods overcame feature engineering bottlenecks but introduced new dependencies on large annotated corpora and opaque representations.
Graph-based methods addressed structural modeling gaps left by sequential encoders, yet their standalone advantage diminished as pretrained language models grew stronger, with many competitive systems adopting hybrid graph-and-PLM architectures.
Generative methods offered greater output flexibility, but at the cost of structural consistency and inference efficiency.
Understanding these transitions as responses to specific unresolved problems, rather than as purely chronological progress, provides a more principled basis for evaluating current methods and anticipating future directions.
We begin by reviewing traditional statistical machine learning approaches (\S\ref{Statistical Machine Learning Methods}), which rely on manually designed features and classical classifiers to identify semantic roles.
We then discuss neural network-based methods (\S\ref{Neural Network Methods}), covering CNNs, RNNs, Transformers, and their advanced variants, which enable automatic representation learning from raw text.
Graph-based approaches (\S\ref{Graph-based Methods}) are examined next, as they explicitly model the structural dependencies between predicates and arguments through graph encoding and structured decoding.
Finally, we cover generative language models and LLMs (\S\ref{Generative Methods}), which reframe SRL as a sequence generation task or natural language definition task and leverage large-scale pretraining for flexible role prediction.
Each paradigm represents a distinct stage in the methodological progression of SRL research, yet the boundaries between them are not always sharp: later paradigms frequently incorporate components from earlier ones, and no single paradigm dominates across all deployment conditions.
The conditions under which each paradigm is most suitable are discussed in \S\ref{Practical Guidance for Method Selection}.

\begin{figure*}[t]
    \centering
    \includegraphics[width=\linewidth]{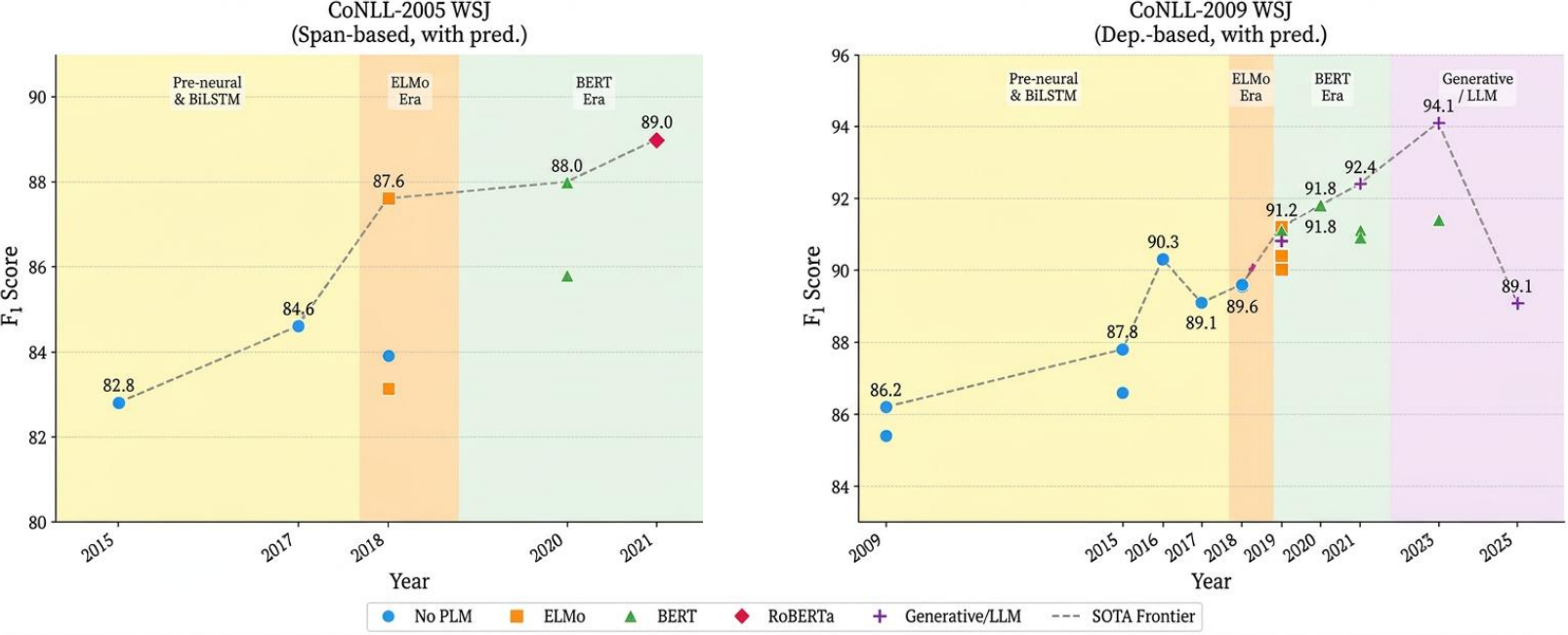}
    \caption{
    Performance trajectory of representative SRL systems on CoNLL-2005 (span-based, WSJ) and CoNLL-2009 (dependency-based, WSJ) test sets.
    Each point represents a published system colored by its primary representation type; the dashed line traces the SOTA frontier.
    Background shading indicates the dominant paradigm of each period.
    A consistent pattern of diminishing marginal returns is observed across successive paradigm transitions on both benchmarks.
    Results are not directly comparable across systems due to differences in experimental configurations; the figure is intended to illustrate broad performance trends rather than precise system rankings.
    }
\label{fig:srl_trajectory}
\end{figure*}

To provide a higher-level view of the field's trajectory, it is instructive to trace the quantitative progression of SRL performance across paradigm transitions on shared benchmarks.
Figure~\ref{fig:srl_trajectory} plots the F$_1$ scores of representative systems on CoNLL-2005 and CoNLL-2009 WSJ test sets from 2015 to 2024, organized by the dominant representation paradigm of each period.
On the CoNLL-2005 WSJ test set, early statistical systems reported F$_1$ scores in the range of 77 $\sim$ 80 \citep{DBLP:conf/acl/PradhanWHMJ05, DBLP:conf/conll/KoomenPRY05}.
The introduction of deep BiLSTM architectures pushed this figure to approximately 82 $\sim$ 83 \citep{DBLP:conf/acl/ZhouX15, DBLP:conf/acl/HeLLZ17a}.
The addition of ELMo contextual embeddings produced a further gain of roughly 2 $\sim$ 3 F$_1$ points, lifting top systems to the 84 $\sim$ 86 range \citep{DBLP:conf/emnlp/StrubellVAWM18, DBLP:conf/emnlp/OuchiS018}.
Subsequent adoption of BERT and RoBERTa representations brought leading systems to approximately 88 $\sim$ 89 F$_1$ \citep{DBLP:conf/acl/FeiWRLJ21}.
A similar trajectory is observed on CoNLL-2009: pre-neural systems achieved F$_1$ scores around 85 $\sim$ 87 \citep{DBLP:conf/emnlp/ZhaoCK09, DBLP:conf/naacl/LeiZVMB15}, BiLSTM-based models advanced this to roughly 89 $\sim$ 90 \citep{DBLP:conf/emnlp/MarcheggianiT17, DBLP:conf/coling/CaiHLZ18}, and BERT-augmented systems reached the low nineties \citep{DBLP:conf/emnlp/LiZWP20}.
These figures reveal a consistent pattern of diminishing marginal returns across successive paradigm shifts.
The gain from transitioning from statistical to neural methods was approximately 5 $\sim$ 6 F$_1$ points on CoNLL-2005 WSJ.
The gain from adding ELMo was approximately 2 $\sim$ 3 points.
The gain from BERT over ELMo-augmented systems narrowed to approximately 1 $\sim$ 2 points.
This pattern of saturation on in-domain benchmarks suggests that further progress on standard test sets is unlikely to reflect meaningful advances in semantic understanding, and that out-of-domain generalization and low-resource performance represent more informative axes of evaluation for future work.

\subsection{Statistical Machine Learning Methods}
\label{Statistical Machine Learning Methods}

The emergence of statistical machine learning methods in the early 2000s marked a paradigm shift in SRL, introducing probabilistic frameworks and data-driven feature learning.
These approaches addressed the scalability and generalization limitations of earlier rule-based systems, which relied heavily on manual feature engineering and domain expertise.
By recasting SRL as a supervised classification task, statistical models enabled systematic feature extraction from annotated corpora, greatly reducing manual effort and improving model robustness.

The seminal work of \citet{DBLP:conf/acl/GildeaJ00} established a two-stage statistical pipeline for SRL based on the FrameNet lexical resource, decomposing the task into frame element boundary identification and semantic role classification.
While this work operated within the FrameNet paradigm, its core pipeline architecture of separating constituent identification from role assignment proved broadly influential.
Subsequent research extended this framework to the PropBank annotation scheme, forming the basis for the CoNLL-2004/2005 shared tasks~\citep{DBLP:conf/conll/CarrerasM05}, which standardized large-scale evaluation using PropBank-style argument labels and accelerated progress in the field.

A major advancement came with the application of support vector machines by~\citet{DBLP:conf/naacl/PradhanWHMJ04}, who systematically explored the combination of predicate properties, syntactic paths, phrase types, and positional features.
Their results demonstrated that machine learning models could effectively exploit complex feature interactions when provided with carefully engineered linguistic features, reinforcing the central role of manual feature design in statistical SRL.

Subsequent research explored alternative classifiers.
\citet{DBLP:conf/emnlp/XueP04} demonstrated, using a maximum entropy classifier, that systematic calibration and selection of syntactic features could yield strong SRL performance, emphasizing that the quality and coverage of the feature set were often more critical to final performance than the choice of classifier architecture itself.
This finding reveals an important trade-off in the statistical era: classifier architecture contributed less to final performance than the quality and coverage of the feature set, which placed the burden of improvement squarely on linguistic expertise rather than algorithmic innovation.

The importance of leveraging diverse syntactic information was further highlighted by~\citet{DBLP:conf/acl/PradhanWHMJ05}, who demonstrated that combining multiple syntactic views, including constituency parses, dependency parses, and shallow chunks, could significantly improve performance by compensating for individual parser errors.
This established multi-view learning as a key paradigm for robust statistical SRL and underscored the value of automated feature selection and balancing model complexity with performance.
However, this multi-view strategy also introduced a practical limitation: performance gains were contingent on the availability of multiple high-quality parsers, and errors from any single parser could propagate through the pipeline in ways that were difficult to diagnose or correct.

As the field matured, structured prediction methods such as conditional random fields gained prominence.
\citet{DBLP:journals/coling/ToutanovaHM08} leveraged CRFs to jointly model argument structures, capturing global dependencies and long-distance relationships.
Similarly, \citet{bjorkelund-etal-2009-multilingual} developed integrated models for joint predicate sense disambiguation and role labeling, further improving the modeling of interdependencies between subtasks.
This line of structured prediction research has continued into the neural era: \citet{Ai_Tu_2024} extended CRF-based modeling to FSRL by introducing arbitrary-order CRFs that capture global interactions among all frame elements of a predicate, going beyond the pairwise or local dependencies modeled by earlier CRF approaches.

The scope of statistical SRL was further extended by the CoNLL-2008 shared task~\citep{DBLP:conf/conll/SurdeanuJMMN08}, which introduced joint syntactic-semantic parsing.
While this task fostered advances in English SRL, it was initially limited to monolingual evaluation.
The subsequent CoNLL-2009 shared task~\citep{DBLP:conf/conll/HajicCJKMMMNPSSSXZ09} expanded evaluation to seven languages, revealing that for most participating languages, sequential pipeline architectures with rich feature sets were competitive with or outperformed joint learning methods, though results varied across languages and metrics, with joint models showing advantages in certain morphologically rich settings.
This result was somewhat counterintuitive: joint modeling was expected to benefit from shared syntactic and semantic supervision, yet the added complexity of joint inference appeared to hinder optimization in practice, particularly for morphologically rich languages with smaller training sets.
These findings provided valuable insights for designing robust multilingual statistical SRL systems, but also exposed a persistent tension between model expressiveness and trainability that would recur in later neural approaches.

These developments established the statistical learning era as a crucial foundation for subsequent neural and generative approaches in SRL.
However, the heavy reliance on manual feature engineering and limited ability to capture complex, long-range dependencies eventually motivated the transition to neural network-based methods.
From a data efficiency perspective, statistical methods require relatively modest amounts of annotated data, as hand-crafted features encode strong linguistic priors that reduce dependence on large corpora.
However, feature templates designed for one domain or language typically require substantial redesign when applied elsewhere, limiting their reusability across diverse settings.
A further unresolved challenge from this era concerns the interaction between parser quality and SRL performance: because statistical SRL systems depend heavily on upstream syntactic parsers, errors introduced at the parsing stage propagate directly into role labeling, and no principled mechanism existed to quantify or bound this error propagation across the pipeline.
Crucially, the problems that statistical methods left unsolved did not disappear with the transition to neural approaches.
The sensitivity to parser quality resurfaced in graph-based methods, which similarly depend on dependency trees of varying reliability.
The domain generalization problem persisted into the neural era and remains only partially addressed by large pretrained models.
The tension between model expressiveness and trainability, prominently highlighted in joint syntactic-semantic models during the CoNLL-2009 shared task, recurred in later neural architectures that attempted to jointly model multiple SRL subtasks.
Recognizing these continuities across paradigms is important: they suggest that certain challenges in SRL are not architectural in nature but reflect deeper properties of the task itself, including its dependence on implicit world knowledge, its sensitivity to annotation conventions, and the structural complexity of predicate-argument interactions in natural language.

\subsection{Neural Network Methods}
\label{Neural Network Methods}

The advent of neural network methods marked a paradigm shift in SRL, moving from feature engineering to automated representation learning.
This subsection reviews the evolution of neural architectures in SRL, encompassing early CNNs and RNNs, advanced recurrent variants, and attention-based Transformer models.
These neural approaches have been extensively applied to both PropBank and FrameNet SRL benchmarks, facilitating cross-paradigm advances.

Early neural SRL architectures leveraged convolutional and recurrent neural networks.
\citet{DBLP:journals/jmlr/CollobertWBKKK11} pioneered the application of CNNs to PropBank-style SRL, introducing a unified neural architecture that learned features automatically from raw text, significantly reducing reliance on hand-crafted features.
CNN-based models are computationally efficient due to their parallelizable convolution operations, making them well-suited for deployment in latency-sensitive settings.
However, CNNs capture only local context within a fixed receptive field, which limits their ability to model long-range predicate-argument dependencies that span many tokens.
Motivated by the sequential nature of language, researchers soon adopted RNNs to better capture long-range dependencies.
\citet{DBLP:conf/acl/ZhouX15} introduced one of the first RNN-based models for SRL, demonstrating improved capability over traditional approaches.
To address the vanishing gradient problem, more advanced recurrent architectures such as long short-term memory~\citep[LSTM,][]{hochreiter1997long} and gated recurrent units~\citep[GRU,][]{cho-etal-2014-properties} were adopted for SRL~\citep{DBLP:conf/emnlp/FitzGeraldTG015, DBLP:conf/emnlp/WangJCS15, DBLP:conf/paclic/OkamuraTITIIAU18, DBLP:conf/emnlp/XiaLZ19}.
However, recurrent architectures process tokens sequentially, which limits parallelism during training and increases inference latency on long sequences.
This sequential bottleneck became a practical concern as SRL was applied to longer documents and conversational contexts, where batch processing efficiency matters for deployment.
Notably, \citet{DBLP:conf/emnlp/FitzGeraldTG015} demonstrated the effectiveness of neural models on both PropBank and FrameNet datasets.
\citet{DBLP:conf/acl/HeLLZ17a} presented a BiLSTM-based model for PropBank SRL, achieving strong performance even without explicit syntactic input, while \citet{DBLP:conf/acl/ZhaoHLB18} further integrated CNNs and BiLSTMs for enhanced feature extraction.

The introduction of attention mechanisms and Transformer-based models brought further breakthroughs.
\citet{DBLP:conf/aaai/TanWXCS18} were among the first to incorporate self-attention into BiLSTM-based SRL, demonstrating that attention mechanisms could enhance global dependency modeling beyond what recurrent layers alone could capture.
\citet{DBLP:conf/emnlp/StrubellVAWM18} proposed the LISA model, which incorporated syntactic knowledge directly into the self-attention mechanism by training specific attention heads to attend to syntactic parents, enabling the model to jointly perform dependency parsing and SRL within a multi-task learning framework.
This design demonstrated that syntax-injected self-attention could effectively capture both local and global predicate-argument dependencies without relying on separate syntactic preprocessing.
Subsequent studies, such as~\citet{DBLP:journals/corr/abs-1904-05255}, showed that simple BERT-based models could achieve outstanding results on both span-based SRL and dependency-based SRL without explicit syntactic features.
These advances highlighted the generalizability of Transformer models across different SRL paradigms.
Beyond NLP-centric applications, Transformer-based span classification techniques have also been adopted in engineering domains for entity and relation labeling without relying on tree-based schemes~\citep{siddharth2024retrieval}.
Nevertheless, the self-attention mechanism in standard Transformers has quadratic complexity with respect to sequence length, which poses practical challenges for processing long inputs in real-time or resource-constrained environments.
A further concern is that strong benchmark performance of BERT-based models does not always translate to robust out-of-domain generalization, as evidenced by the persistent WSJ-to-Brown performance gap reported in Tables~\ref{tab:result-05} and~\ref{tab:result-09}, even for systems using large PLMs.

Recent trends emphasize leveraging large PLMs for SRL in diverse languages and resource settings.
\citet{DBLP:conf/coling/DannellsJB24} explored transfer learning with pre-trained Transformers for Swedish SRL.
Researchers are investigating hybrid architectures that integrate Transformer components with lightweight convolutional or recurrent layers to balance performance and efficiency.
Such hybrid designs aim to retain the representational power of Transformers while reducing memory footprint and inference time, which is particularly important for deployment on edge devices or in streaming applications.
Neural network methods spanning from CNNs and RNNs to Transformers have become the backbone of modern SRL, supporting both PropBank and FrameNet paradigms.
The data requirements of neural SRL systems vary considerably across architectures.
CNN and RNN models trained from scratch typically require tens of thousands of annotated examples to reach competitive performance, whereas PLM-based approaches substantially reduce this requirement by transferring knowledge from large unlabeled corpora.
Nevertheless, fine-tuning large PLMs on small SRL datasets remains prone to overfitting, motivating continued research into data augmentation and semi-supervised training for annotation-scarce settings.
The interaction between PLMs and explicit syntactic supervision is discussed in detail in Section~\ref{Syntax Feature Modeling in SRL}.

\subsection{Graph-based Methods}
\label{Graph-based Methods}

Graph-based methods have emerged as a significant paradigm in SRL, primarily due to their ability to explicitly model the complex structural relationships between predicates and arguments.
Early evidence, such as the work by~\citet{DBLP:conf/acl/RothL16}, demonstrated that embedding dependency paths can significantly improve the representation of predicate-argument structures, paving the way for graph-based neural approaches.

The introduction of graph convolutional networks (GCNs) into SRL by~\citet{DBLP:conf/emnlp/MarcheggianiT17} marked an important step in this direction.
This foundational work showed that syntactic information, especially dependency parses, can be effectively encoded via GCNs, which operate over graphs constructed from dependency trees by converting directed dependency arcs into bidirectional edges.
By combining GCN layers with LSTM encoders, their model was able to capture both local and non-local dependencies, achieving notable improvements in role labeling accuracy.
This approach established a natural framework for modeling intricate predicate-argument interactions, surpassing the limitations of purely sequential architectures.
It should be noted, however, that simpler sequential models have in certain settings achieved performance comparable to GCN-based systems, suggesting that the structural inductive bias of graph methods is not universally decisive.
This inconsistency across experimental conditions points to an unresolved question: whether the gains from graph-based encoding reflect genuine structural reasoning or are instead an artifact of specific dataset characteristics and evaluation protocols.

Subsequent research expanded upon this foundation.
\citet{DBLP:conf/emnlp/LiHCZZLLS18} proposed a unified framework that integrates multiple types of syntactic encoders, including syntactic GCNs, syntax-aware LSTMs, and tree-structured LSTMs, on top of deep BiLSTM encoders.
Their empirical study highlighted the complementary strengths of different structural representations, demonstrating that hybrid models leveraging multiple views of syntax can yield more robust and accurate semantic role predictions.
However, combining multiple syntactic encoders also increases model complexity and training cost, and the marginal gains from each additional encoder type diminish as the base encoder grows stronger, raising questions about the practical value of such combinations in high-resource settings.

\citet{DBLP:conf/aaai/0001LLJ21} introduced a novel encoder-decoder framework for unified SRL, incorporating a label-aware graph convolutional network (LA-GCN) to encode both syntactic dependency arcs and role labels into BERT-based word representations.
They also designed a high-order interacted attention mechanism~\citep{DBLP:conf/ijcnlp/FeiRJ20} that leverages previously recognized predicate-argument-role triplets to inform current decisions, enabling dynamic modeling of semantic dependencies.
This design represents a shift from static syntactic structures toward more flexible, context-sensitive graph-based representations.

The influence of related graph-based architectures, such as graph attention networks~\cite[GATs,][]{DBLP:conf/iclr/VelickovicCCRLB18}, which were originally developed for general graph learning, has also been evident in attention-based variants applied to SRL and related tasks.
Other works, such as~\citet{DBLP:conf/coling/CaiHLZ18} and \citet{DBLP:conf/emnlp/LiZWP20}, further explored the integration and unification of span- and dependency-based SRL within graph-based frameworks.

Beyond GCN-based architectures, Graph Transformer models have recently attracted growing attention in the SRL community.
\citet{mohammadshahi-henderson-2023-syntax} proposed the Syntax-aware Graph-to-Graph Transformer (SynG2G-Tr), which inputs syntactic dependency relations as embeddings directly into the Transformer self-attention mechanism, introducing a soft structural bias while retaining the flexibility to learn alternative attention patterns.

Graph-based methods have made substantial contributions to SRL by enabling the explicit modeling of both syntactic and semantic dependencies, and continue to serve as a foundation for ongoing innovations in the field.
The explicit structural inductive bias of dependency graphs can improve data efficiency in learning predicate-argument structures compared to purely sequential models, but this advantage presupposes the availability of high-quality parse trees, which themselves require substantial annotation effort.
For low-resource languages where both labeled SRL data and reliable parsers are scarce, the potential efficiency gains of graph-based methods are therefore considerably diminished.
A further limitation concerns scalability: graph construction over full dependency trees incurs overhead that grows with sentence length, and the benefit of graph encoding has been shown to decrease as PLM-based encoders become stronger, suggesting that the structural information captured by GCNs may be partially redundant with what large pretrained models already encode implicitly.

\subsection{Generative Methods}
\label{Generative Methods}
\begin{table*}[ht]
\fontsize{7}{8}\selectfont
\setlength{\tabcolsep}{0.9mm}

\begin{center}
\begin{NiceTabular*}{\textwidth}{@{\extracolsep{\fill}}lccccccccc}[
    code-before = \rowcolor{blue!15}{1} \rowcolor{gray!15}{5}
]
\Xhline{0.08em}
\multicolumn{10}{c}{\emph{\textbf{Generative SRL}}} \\
\multirow{2}{*}{\bf Method} & \multicolumn{3}{c}{\bf CoNLL 2009 WSJ} & \multicolumn{3}{c}{\bf CoNLL 2009 Brown} & \multicolumn{3}{c}{\bf CoNLL 2012} \\
\cmidrule{2-4}\cmidrule{5-7}\cmidrule{8-10}
& \bf P & \bf R & \bf F$_1$ & \bf P & \bf R & \bf F$_1$ & \bf P & \bf R & \bf F$_1$ \\
\hline
\citet{DBLP:conf/emnlp/DazaF19} & - & - & 90.8 & - & - & 84.1 & - & - & 75.4\\
\citet{DBLP:conf/ijcai/BlloshmiCTN21} & 92.9 & 92.0 & 92.4 & 85.8 & 84.5 & 85.2 & 87.8 & 86.8 & 87.3 \\
\end{NiceTabular*}

\begin{NiceTabular*}{\textwidth}{@{\extracolsep{\fill}}lcccccccccc}[
    code-before = \rowcolor{blue!15}{1} \rowcolor{gray!15}{6,8,10,12,14}
]
\specialrule{.2em}{.05em}{0.05em} 
\multicolumn{11}{c}{\textit{\textbf{LLM-based SRL}}} \\
\multirow{2}{*}{\bf Method} & \multirow{2}{*}{\bf Shot} & \multicolumn{3}{c}{\bf CoNLL 2009 WSJ} & \multicolumn{3}{c}{\bf CoNLL 2009 Brown} & \multicolumn{3}{c}{\bf CoNLL 2012} \\
\cmidrule{3-5}\cmidrule{6-8}\cmidrule{9-11}
& & \bf P & \bf R & \bf F$_1$ & \bf P & \bf R & \bf F$_1$ & \bf P & \bf R & \bf F$_1$ \\
\hline
ChatGPT+FT~\cite{DBLP:journals/corr/abs-2306-09719} & Full & - & - & 94.1 & - & - & - & - & - & 88.6 \\
\citet{DBLP:conf/icic/ChengYWLYZTXH24} & \\
\quad Davinci+CoT & 3 & 6.29 & 26.06 & 10.13 & 4.01 & 18.70 & 6.60 & 2.50 & 16.13 & 4.33 \\
\quad Davinci+PromptSRL & 3 & 12.07 & 14.79 & 13.29 & 7.13 & 21.64 & 10.73 & 4.08 & 15.70 & 6.48 \\
\quad Llama2-7B-Chat+PromptSRL & 3 & 5.49 & 16.46 & 8.23 & 5.34 & 14.58 & 7.82 & 1.83 & 10.67 & 3.13\\
\quad ChatGLM2-6B+PromptSRL & 3 & 12.72 & 34.12 & 18.53 & 8.94 & 22.83 & 12.85 & 5.68 & 29.58 & 9.53 \\
\quad Ada+PromptSRL & 3 & 2.00 & 1.41 & 1.65 & 0.98 & 0.75 & 0.85 & 0.47 & 0.83 & 0.60 \\
\quad Babbage+PromptSRL & 3 & 3.45 & 1.41 & 2.00 & 3.94 & 3.73 & 3.83 & 4.64 & 7.44 & 5.72 \\
\quad Curie+PromptSRL & 3 & 3.74 & 5.63 & 4.49 & 6.36 & 11.19 & 8.11 & 3.67 & 7.44 & 4.92 \\
\quad ChatGPT+PromptSRL & 3 & 39.19 & 41.73 & 40.42 & 37.59 & 41.32 & 39.37 & 36.57 & 40.83 & 38.58 \\
\citet{li-etal-2025-llms-also} & Full & 89.92 & 88.23 & 89.07 & 83.36 & 79.04 & 81.14 & 85.64 & 85.59 & 85.61 \\
\bottomrule
\end{NiceTabular*}
\end{center}
\caption{
Performance comparison of generative and LLM-based SRL methods on CoNLL 2009 (WSJ and Brown test sets) and CoNLL 2012 datasets.
P, R, and F$_1$ denote precision, recall, and F$_1$ score respectively.
A dash (-) indicates that the corresponding metric was not reported in the original paper.
The Shot column indicates whether the method uses full supervised training (Full) or $k$-shot in-context learning.
All scores are taken directly from the respective original publications and reflect heterogeneous experimental conditions, including differences in predicate identification strategy, PLM version, and output post-processing procedures.
Direct numerical comparison across rows should therefore be treated with caution.
}
\label{tab:res_glm}
\end{table*}

Traditional discriminative methods in SRL, including those based on neural sequence labeling~\citep{DBLP:conf/acl/ZhouX15} and span classification~\citep{DBLP:journals/jmlr/CollobertWBKKK11}, have achieved strong results but face inherent limitations in modeling complex semantic dependencies and managing uncertainty in predictions.
To address these challenges, generative methods have emerged as a significant paradigm shift, providing a probabilistic framework that naturally models the joint distribution of predicates and their semantic arguments.
Unlike discriminative approaches that directly model the conditional probability of outputs given inputs, generative models attempt to capture the joint probability of both inputs and outputs, offering a more holistic view of semantic structure.
Table~\ref{tab:res_glm} compares the performance of generative and LLM-based SRL methods on CoNLL datasets, highlighting the effectiveness and evolution of this paradigm.
The scores reported in Table~\ref{tab:res_glm} are taken directly from the respective original publications and reflect heterogeneous experimental conditions, including differences in predicate identification strategy, PLM version, and output post-processing procedures.
Direct numerical comparison across rows should therefore be treated with caution, as observed score differences may reflect experimental setup variation rather than genuine differences in model capability.

\subsubsection{Early Generative Models}

The earliest generative approaches to SRL were motivated by the need to better capture sequential dependencies and uncertainty in semantic role assignments.
\citet{DBLP:conf/semeval/ThompsonPA04} introduced an early generative framework for SRL, modeling the task using a first-order hidden Markov model conditioned on frame assignments, where a target generates a frame which in turn generates a linearized role sequence over sentence constituents.
Subsequently, \citet{DBLP:conf/conll/YuretYU08} presented a more sophisticated generative model that considered the joint probability of semantic dependencies, enabling interaction between prediction stages and improving the coherence of role assignments.
These early models demonstrated that generative formulations could improve output coherence compared to independent classification, but they were limited by the expressive capacity of shallow probabilistic models and did not scale well to the full diversity of predicate types and argument structures found in standard benchmarks.

\subsubsection{GenLMs}

The emergence of GenLMs marked a transformative shift in SRL methodology.
\citet{DBLP:conf/rep4nlp/DazaF18} recast SRL as a sequence-to-sequence generation task, aligning semantic role annotation with human language generation.
This approach was further extended to cross-lingual settings by~\citet{DBLP:conf/emnlp/DazaF19}, who proposed a translate-and-label encoder-decoder model that jointly translates sentences and generates SRL annotations in multiple languages.
\citet{DBLP:conf/ijcai/BlloshmiCTN21} introduced the first end-to-end unified sequence-to-sequence model for SRL, integrating predicate sense disambiguation, argument identification, and classification into a single framework.
This unified generation paradigm achieved strong performance on both span- and dependency-based SRL tasks, demonstrating that generative modeling can surpass traditional sequence labeling methods by offering a more flexible approach to semantic role prediction.
From a computational perspective, sequence-to-sequence models incur higher inference costs than discriminative classifiers because output length is variable and decoding proceeds autoregressively, which can limit throughput in latency-sensitive applications.
A further concern is output validity: autoregressive decoding does not guarantee that the generated sequence forms a well-formed predicate-argument structure, and post-hoc constraint enforcement is often required to handle malformed or incomplete outputs.

\subsubsection{LLMs}

With the success of LLMs in a variety of NLP tasks, recent research has explored their application to SRL.
\citet{DBLP:journals/corr/abs-2306-09719} investigated the use of ChatGPT for SRL, demonstrating the feasibility of leveraging LLMs' semantic knowledge for argument labeling.
\citet{DBLP:conf/icic/ChengYWLYZTXH24} systematically evaluated LLMs' capabilities in capturing structured semantics through SRL tasks, and proposed a prompt-based framework that further enhances LLM performance in both zero-shot and few-shot scenarios.
Their evaluation protocol covers both zero-shot and few-shot settings, using standard CoNLL benchmarks as the test bed and F$_1$ score as the primary metric, consistent with supervised SRL evaluation.
However, this evaluation design itself carries limitations that deserve attention.
F$_1$-based comparison between prompt-based LLMs and supervised systems conflates two fundamentally different output regimes: supervised systems produce structured label sequences with deterministic decoding, whereas prompt-based systems generate free-form text that must be post-processed into role-argument pairs, introducing an additional parsing step that is sensitive to output format variation.
The findings of \citet{DBLP:conf/icic/ChengYWLYZTXH24} reveal both the potential and limitations of LLMs in semantic understanding, particularly in handling long-distance dependencies and complex structures.

Prompt-based methods exhibit several recurring failure patterns in SRL-specific evaluations.
First, LLMs frequently produce argument spans that are semantically plausible but do not align with gold annotation boundaries, resulting in low exact-match scores even when the core semantic content is correctly identified.
Second, role label consistency degrades on predicates with multiple closely related senses, where the prompt does not supply sufficient sense-disambiguating context.
Third, chain-of-thought prompting, while improving reasoning coherence on open-ended tasks, does not reliably enforce the structural constraints of SRL, such as the requirement that each predicate-argument-role triple be complete and non-overlapping.
\citet{spaulding-etal-2025-role} provide further evidence of this limitation: even when LLMs are supplied with semantic proto-role labels as additional context, SRL accuracy does not improve consistently, and in high-performing models such as GPT-4o, the additional context can actively degrade performance.
Their results suggest that LLMs implicitly encode multi-dimensional event role knowledge, but this internal representation does not translate reliably into the structured categorical outputs required by SRL.
Fourth, few-shot exemplar selection has a disproportionate effect on output quality.
Exemplars drawn from a different predicate type or syntactic construction than the test instance lead to systematic role confusion, particularly for adjunct roles such as ARGM-CAU and ARGM-PRP.
\citet{raghav-jana-2025-llms} corroborate this finding across a broader set of models including Llama, Mistral, Qwen, OpenChat, and Gemini, showing that few-shot prompting consistently improves over zero-shot across all evaluated models, yet the best-performing model still falls substantially short of supervised systems.
Their study also demonstrates that reformulating SRL as a question answering task partially mitigates output format inconsistency, though it does not resolve the underlying difficulty of assigning precise argument boundaries.

As shown in Table~\ref{tab:res_glm}, even with prompt engineering and chain-of-thought strategies, LLM-based methods still fall considerably short of specialized systems on standard benchmarks under fully supervised settings.
Several patterns in the table warrant closer attention.
Chain-of-thought prompting does not improve structured prediction: Davinci+CoT achieves an F$_1$ of 10.13 on CoNLL 2009 WSJ, below the 13.29 of Davinci+PromptSRL, suggesting that free-form reasoning chains do not improve and may hinder the production of structured role-argument outputs.
A related pattern appears in recall-precision asymmetry: ChatGLM2-6B+PromptSRL achieves a recall of 34.12 but a precision of only 12.72 on CoNLL 2009 WSJ, indicating that the model over-generates argument spans rather than making precise assignments.
Smaller legacy models largely fail to produce meaningful outputs, with Ada and Babbage reaching F$_1$ scores below 6.0 across all test sets, suggesting that a minimum model capacity threshold must be exceeded before prompt-based SRL yields usable results.
The gap between few-shot and fully supervised performance is particularly striking: ChatGPT under 3-shot prompting achieves an F$_1$ of 40.42 on CoNLL 2009 WSJ, whereas a fully supervised ChatGPT-based system \citep{DBLP:journals/corr/abs-2306-09719} reaches 94.1 on the same benchmark.
This gap reflects three structural sources of difficulty that prompt engineering alone cannot resolve.
First, PropBank role labels are predicate-specific and defined by roleset entries in the PropBank lexicon, which are not accessible to a prompt-based model unless explicitly supplied, making reliable role assignment for low-frequency predicates particularly difficult.
Second, a 3-shot prompt exposes the model to only a small fraction of the full label inventory, leaving the majority of role types ungrounded.
\citet{raghav-jana-2025-llms} conduct a comprehensive error analysis across five models, identifying imprecise span boundary detection, inaccurate argument extraction, and formatting deviations as the dominant failure modes.
Third, output format deviation introduces evaluation penalties that are independent of semantic accuracy, as post-processing free-form text into structured role-argument pairs is sensitive to minor wording variations in model output.
Supervised fine-tuning addresses all three sources of difficulty within a single training objective, which accounts for the larger few-shot to fully supervised gap in SRL relative to most other NLP tasks.

An additional reproducibility concern specific to LLM-based evaluation is that prompt-based systems are sensitive to minor variations in prompt wording, exemplar selection, and model version, which means that reported scores may not be stable across runs or replication attempts.
To overcome these challenges, \citet{li-etal-2025-llms-also} proposed a retrieval-augmented framework that equips LLMs with external linguistic knowledge and self-correction mechanisms, achieving the first instance of an LLM-based method surpassing traditional encoder-decoder approaches on complete SRL tasks across both Chinese and English benchmarks.
Despite these advances, deploying LLMs for SRL in production settings raises significant practical concerns.
Models such as GPT-4 and LLaMA-2 require substantial GPU memory and exhibit inference latency that is orders of magnitude higher than compact discriminative models, making them difficult to integrate into real-time pipelines or resource-constrained environments.
Prompt-based approaches also introduce non-determinism and sensitivity to prompt formulation, which complicates reproducibility and quality control in downstream applications.

The limitations of LLMs in SRL are not merely empirical but reflect a fundamental architectural mismatch.
LLMs are trained with a next-token prediction objective, which optimizes for local fluency rather than global structural consistency.
SRL requires that all predicate-argument-role triples within a sentence form a coherent and complete structure, where each assignment is constrained by the others.
Autoregressive decoding generates each token without explicit awareness of the full output structure, and therefore provides no mechanism to enforce the global interdependence among predicate-argument-role assignments.
LLMs consequently produce outputs that are locally coherent but globally inconsistent, and this problem is most visible in sentences with multiple predicates or overlapping argument spans.
This architectural mismatch manifests differently depending on argument structure complexity.
For simple sentences with a single predicate and two or three core arguments (e.g., ARG0 and ARG1), LLMs generally produce outputs that are structurally well-formed, as the local context is sufficient to determine role boundaries and the output space is small enough to be navigated by pattern matching from pre-training data.
Performance degrades substantially, however, as structural complexity increases.
In sentences with multiple predicates sharing overlapping argument spans, LLMs frequently misattribute arguments to the wrong predicate or produce duplicate role assignments across predicates.
For adjunct roles such as ARGM-TMP, ARGM-LOC, and ARGM-CAU, which are not lexically anchored to a specific predicate type, LLMs show inconsistent labeling behavior even within the same document.
\citet{spaulding-etal-2025-role} observe that SRL errors do not overwhelmingly co-occur with SPRL errors, but do overwhelmingly co-occur with SPRL properties annotated as ``not applicable'' to the argument. They further find that GPT-4o performs poorly on prompt variants requiring the most reasoning steps, suggesting that LLM performance degrades as argument-predicate analysis increases in complexity.
LLM performance on SRL is therefore strongest on high-frequency, structurally simple configurations that are well-represented in pre-training corpora, and weakest on complex and low-frequency structures that pre-training data cannot adequately cover.
\citet{dukic2025supervised} further show that long-context inputs consistently degrade generative sequence labeling performance in both ICL and supervised fine-tuning settings, and that this deficiency can be mitigated by removing task instructions from the prompt, as instructions prove largely unnecessary for strong performance under their proposed framework.

Whether LLMs can fully substitute for specialized SRL systems remains an open question.
The evidence so far suggests that LLMs and specialized systems occupy distinct and complementary positions.
LLMs demonstrate broad coverage and flexibility, but they remain unreliable in scenarios that demand precise argument boundary detection, consistent handling of nested or discontinuous spans, and stable performance under low-resource or domain-specific conditions.
Specialized SRL systems, by contrast, offer controllable outputs, interpretable decision processes, and robust performance on structured prediction tasks where exact label assignment matters.
Beyond performance, specialized systems serve a distinct function in the NLP pipeline: they produce explicit predicate-argument structures that can be directly consumed by downstream tasks such as information extraction, question answering, and machine translation, without relying on the implicit and sometimes inconsistent reasoning of LLMs.
Rather than rendering specialized SRL systems obsolete, LLMs are better understood as providing stronger contextual representations that specialized systems can build upon.
The most productive direction is therefore not to replace one with the other, but to develop hybrid approaches in which LLMs supply rich semantic representations while specialized architectures enforce structural constraints and ensure output reliability.
The SIFT framework of \citet{dukic2025supervised} and the QA-based reformulation of \citet{raghav-jana-2025-llms} represent early steps in this direction, combining the flexibility of generative models with task-specific structural constraints.

Generative methods have substantially transformed SRL research, evolving from early probabilistic models to advanced sequence-to-sequence and LLM-based approaches.
These innovations enable more natural and flexible modeling of semantic roles, and open up directions for future research integrating external knowledge, prompt engineering, and multilingual adaptation.
Unified sequence-to-sequence models improve data efficiency by sharing parameters across predicate detection, sense disambiguation, and argument classification, making better use of limited annotations than pipeline systems.
LLMs further reduce annotation dependence through in-context learning, though as shown in Table~\ref{tab:res_glm}, few-shot performance still falls substantially short of fully supervised specialized systems.
A key unresolved challenge is how to reconcile the flexibility of generative decoding with the structural rigidity required by SRL, as current approaches either sacrifice output validity for fluency or impose hard constraints that limit generalization to unseen predicate types.

\subsection{Error Analysis}
\label{Error Analysis}

Despite substantial progress across all four methodological paradigms, SRL systems continue to exhibit systematic failure patterns that reveal the boundaries of current approaches.

\textbf{Challenging semantic role types.}
Adjunct roles such as ``ARGM-CAU'' (cause), ``ARGM-PRP'' (purpose), and ``ARGM-DIS'' (discourse) remain consistently difficult across methods, as their realization is highly context-dependent and their boundaries are often ambiguous even for human annotators.
Roles requiring world knowledge or pragmatic inference, such as ``ARGM-ADV'' (adverbials) and ``ARGM-PRD'' (secondary predication), also show lower performance compared to core arguments like ``ARG0'' and ``ARG1''.
In FrameNet-style SRL, rare or fine-grained frame elements present additional difficulty due to limited training instances per frame.
Across all four paradigms, performance on these adjunct and knowledge-intensive roles lags behind core argument performance by a consistent margin, indicating that the challenge is not primarily architectural but reflects a deeper gap between surface distributional patterns and the pragmatic knowledge required for accurate role assignment.

\textbf{Long-distance and discontinuous arguments.}
Arguments that are syntactically distant from their predicates pose persistent challenges for both span-based and dependency-based systems.
Discontinuous argument spans, which occur in FrameNet annotations, are particularly hard to recover with standard sequence labeling or span enumeration approaches.
Even Transformer-based models with global attention struggle when the predicate and its arguments are separated by long intervening clauses.
This failure mode is especially consequential for downstream applications such as information extraction and question answering, where missing a distal argument can lead to an incomplete or incorrect semantic interpretation.

\textbf{Predicate disambiguation errors.}
Incorrect predicate sense assignment propagates errors to downstream argument classification, since role definitions are sense-specific in PropBank.
Polysemous verbs with closely related senses are frequent sources of cascading mistakes in pipeline systems.
A notable inconsistency in the literature is that some systems report strong overall F$_1$ scores while using gold predicate senses as input, which artificially inflates performance and makes cross-system comparisons unreliable when sense identification is handled differently across experimental setups.

\textbf{Domain shift and low-resource settings.}
Out-of-domain evaluation consistently reveals a significant performance gap relative to in-domain results, as reflected in the Brown corpus results in Tables~\ref{tab:result-05} and~\ref{tab:result-09}.
Statistical and early neural models are especially sensitive to domain shift, while PLM-based systems show improved but still imperfect generalization.
For low-resource languages, the scarcity of annotated data amplifies all of the above error types.
Beyond performance degradation, domain shift in socially critical applications raises fairness concerns, as models trained predominantly on newswire text may systematically misassign roles in legal, medical, or conversational contexts, producing outputs that disadvantage underrepresented populations~\citep{siddharth2025data}.
The gap between in-domain and out-of-domain performance has narrowed with the adoption of large PLMs, yet it has not been eliminated, suggesting that benchmark-driven progress does not straightforwardly translate to robust generalization across text types.

\textbf{Cross-sentence and implicit arguments.}
Standard SRL systems trained on sentence-level data fail to recover arguments that are realized in a different sentence or left implicit in discourse.
As shown in Table~\ref{tab:res_diag}, null instantiation accuracy remains low across all evaluated systems, indicating that cross-sentence argument resolution is far from solved.
This limitation is particularly pronounced for FrameNet-style annotation, where implicit arguments are more frequently marked, yet current models are almost entirely trained and evaluated on explicit, within-sentence arguments.

\textbf{LLM-specific error patterns.}
LLM-based methods exhibit failure modes that differ structurally from those of encoder-based systems.
The most consistent failure is argument span boundary imprecision: LLMs frequently identify the correct semantic participant but assign a span that does not match gold annotation boundaries, incurring evaluation penalties under exact-match F$_1$ even when the core semantic content is correct.
Role label inconsistency on polysemous predicates constitutes a second failure mode, as prompt-based models without access to the PropBank lexicon cannot reliably distinguish closely related senses, leading to systematic misassignment on low-frequency predicates.
A third failure mode is structural incompleteness under complex argument configurations: LLMs perform adequately on simple single-predicate sentences but degrade on sentences with multiple predicates or overlapping argument spans, where adjunct roles such as ARGM-TMP and ARGM-CAU show inconsistent labeling even within the same document.
These failure modes are not randomly distributed: errors concentrate on low-frequency predicates, adjunct roles, and structurally complex configurations that are underrepresented in pre-training corpora and cannot be adequately grounded through in-context exemplars alone.

\textbf{Multimodal grounding errors.}
In visual and video SRL, role-filler grounding remains a major source of error.
Systems frequently predict the correct verb and role structure but fail to localize the corresponding visual entity, leading to low grounded accuracy scores as reflected in Table~\ref{tab:result-visual}.
Speech SRL introduces additional errors from ASR transcription mistakes, which propagate into the semantic role assignment stage despite end-to-end training strategies.
A structural challenge specific to multimodal settings is that the evaluation metrics used across modalities are not directly comparable.
Textual SRL is evaluated by span-based F$_1$ scoring or dependency-based labeled attachment F$_1$ depending on the formalism, visual SRL by grounded bounding box accuracy, and video SRL by grounded accuracy over temporal segments or action tubes, making it difficult to assess whether progress in one modality transfers to another.

The recurring error patterns documented above indicate that improvements in context modeling, cross-sentence reasoning, and multimodal grounding are among the most pressing needs for future SRL research.
Many of these failure modes are shared across paradigms, suggesting that they reflect fundamental limitations of current training data and evaluation protocols rather than deficiencies of any single architectural choice.

\begin{table*}[ht]
\centering

\definecolor{clHigh}{RGB}{56,142,60}
\definecolor{clMid}{RGB}{245,124,0}
\definecolor{clLow}{RGB}{198,40,40}
\definecolor{clRange}{RGB}{30,100,180}

\newcommand{\High}{\textcolor{clHigh}{\textbf{$\bullet\bullet\bullet$}}}
\newcommand{\Mid} {\textcolor{clMid} {\textbf{$\bullet\bullet$}}}
\newcommand{\Low} {\textcolor{clLow} {\textbf{$\bullet$}}}

\newcommand{\RangeLH}{\textcolor{clRange}{\textbf{$\bullet\!\sim\!\bullet\bullet\bullet$}}}
\newcommand{\RangeMH}{\textcolor{clRange}{\textbf{$\bullet\bullet\!\sim\!\bullet\bullet\bullet$}}}

\newcommand{\DualLH}{\Low\,/\,\High}
\newcommand{\DualHM}{\High\,/\,\Mid}

\resizebox{\textwidth}{!}{%
\begin{tabular}{lcccc}
\toprule
\textbf{Dimension}
    & \textbf{Statistical}
    & \textbf{Neural}
    & \textbf{Graph-based}
    & \textbf{Generative} \\
\midrule
Annotation requirement
    & \Low
    & \RangeLH\textsuperscript{a}
    & \RangeLH\textsuperscript{a}
    & \DualLH\textsuperscript{b}              \\[4pt]
Inference efficiency
    & \High
    & \DualHM\textsuperscript{c}
    & \Mid
    & \Low                                    \\[4pt]
Out-of-domain generalization
    & \Low
    & \RangeMH
    & \Mid\textsuperscript{$\dagger$}
    & \Mid                                    \\[4pt]
Dependence on syntactic input
    & \High
    & \Low
    & \High
    & \Low                                    \\[4pt]
Output controllability
    & \High
    & \High
    & \High
    & \Low\textsuperscript{d}                 \\[4pt]
Interpretability
    & \High
    & \Mid
    & \DualHM\textsuperscript{e}
    & \Low                                    \\[4pt]
Multilingual transferability
    & \Low
    & \RangeMH
    & \Mid\textsuperscript{$\dagger$}
    & \Mid                                    \\
\bottomrule
\multicolumn{5}{l}{\footnotesize
    \textsuperscript{a}~\Low~(feature-based, limited data) to \High~(with pretrained representations).}\\
\multicolumn{5}{l}{\footnotesize
    \textsuperscript{b}~Few-shot prompting $\to$ \Low;\enspace fine-tuning $\to$ \High.}\\
\multicolumn{5}{l}{\footnotesize
    \textsuperscript{c}~CNN/BiLSTM $\to$ \High;\enspace PLM-based $\to$ \Mid.}\\
\multicolumn{5}{l}{\footnotesize
    \textsuperscript{d}~Output format is difficult to strictly constrain; structured generation requires careful prompt engineering.}\\
\multicolumn{5}{l}{\footnotesize
    \textsuperscript{e}~Explicit graph structure $\to$ \High;\enspace internal GNN computation $\to$ \Mid~(opaque).}\\
\end{tabular}
}
\caption{%
    Cross-paradigm comparison of four major SRL methodological paradigms
    across key practical dimensions. Ratings are relative and intended to
    support method selection under different deployment constraints.%
    \\[2pt]%
    \textcolor{clHigh}{\textbf{$\bullet\bullet\bullet$}}~High,\enspace
    \textcolor{clMid}{\textbf{$\bullet\bullet$}}~Moderate,\enspace
    \textcolor{clLow}{\textbf{$\bullet$}}~Low,\enspace
    \textcolor{clRange}{\textbf{$\bullet\!\sim\!\bullet\bullet\bullet$}}~architecture-dependent range,\enspace
    \Low\,/\,\High~discrete dual-mode.%
    \\[1pt]%
    $\dagger$~Ratings assume a high-quality parser is available for the
    target language and domain.%
}
\label{tab:paradigm_comparison}
\end{table*}

\subsection{Cross-Paradigm Synthesis}
\label{Cross-Paradigm Synthesis}

The four paradigms reviewed above are not independent lines of development but form an interconnected progression in which each stage inherits unresolved problems from its predecessor while introducing new trade-offs of its own.
This subsection extracts several cross-cutting lessons from the literature that are not visible when each paradigm is examined in isolation.

\textbf{What inductive bias does SRL minimally require?}
A recurring question across paradigms is how much structural prior knowledge a model needs in order to perform SRL reliably.
Statistical methods encoded this prior explicitly through syntactic feature templates, dependency path representations, and phrase-type indicators.
Graph-based methods encoded it explicitly through the topology of dependency trees.
Neural sequence models, particularly BiLSTM-based systems, demonstrated that strong SRL performance is achievable without explicit structural supervision, provided sufficient annotated data is available.
Transformer-based models further reduced the apparent need for structural priors by learning positional and relational patterns from large unlabeled corpora.
However, the persistent performance gap on out-of-domain data and on adjunct roles suggests that distributional learning alone does not fully substitute for structured linguistic knowledge.
The evidence across paradigms points to a conditional answer: structural inductive bias improves data efficiency and robustness under low-resource or domain-shifted conditions, but its marginal contribution decreases as model scale and pretraining data increase.
This implies that the value of explicit structural supervision is not constant but depends on the deployment context, a finding with direct implications for resource allocation in annotation projects.

\textbf{Why do graph-based methods plateau in certain settings?}
Graph-based methods consistently improve over sequential baselines when gold-standard parse trees are available and when the base encoder is relatively weak.
However, their advantage diminishes or disappears in two conditions.
First, when large pretrained language models serve as encoders, the structural information provided by dependency graphs appears to be partially redundant with what the PLM already encodes through self-attention over large corpora.
Second, when parse quality is low, as is common for morphologically rich or low-resource languages, graph construction introduces noise rather than useful structure, and the model may perform better without graph encoding altogether.
This plateau pattern suggests that graph-based methods do not provide a universally superior inductive bias but rather serve as a useful supplement in a specific operating regime defined by encoder strength and parse quality.
A practical implication is that the decision to include graph-based components should be conditioned on an empirical assessment of parse reliability in the target setting, rather than adopted as a default architectural choice.

\textbf{The recurring tension between flexibility and structural consistency.}
Each paradigm shift introduced greater output flexibility at the cost of structural consistency.
Statistical pipeline systems produced outputs that were structurally well-formed by construction, since each stage operated within a predefined label space, though error propagation across stages remained a known source of structural inconsistency in the final output.
Neural sequence labeling models relaxed this constraint by learning label transitions implicitly, which improved coverage but occasionally produced invalid label sequences.
Generative models and LLMs extended this trend further, enabling open-ended output formats that can express complex semantic structures but frequently violate the completeness and non-overlap constraints that SRL requires.
This progression reveals a fundamental tension that has not been resolved by any existing paradigm: the more flexible the output space, the harder it is to enforce the global structural constraints that define a valid predicate-argument structure.
Current approaches address this tension through post-hoc constraint enforcement, constrained decoding, or re-ranking, but none of these mechanisms is fully satisfactory.
A principled solution would require architectures that jointly model local generation decisions and global structural validity, which remains an open research problem.

\textbf{Shared failure modes across paradigms.}
As documented in \S\ref{Error Analysis}, several error types recur across all four paradigms with only modest variation in severity.
Adjunct role labeling, long-distance argument recovery, predicate sense disambiguation, and out-of-domain generalization remain challenging regardless of whether the underlying model is a CRF, a BiLSTM, a GCN, or a large language model.
This cross-paradigm persistence suggests that these failure modes are not primarily architectural in origin.
They reflect instead the limits of current training data, which is concentrated in newswire domains and does not adequately cover the pragmatic and world-knowledge requirements of adjunct role assignment.
They also reflect the limits of current evaluation protocols, which measure exact-match performance on explicit within-sentence arguments and do not reward partial credit for semantically correct but boundary-misaligned predictions.
Progress on these persistent challenges is therefore unlikely to come from architectural innovation alone and will require advances in data collection, annotation methodology, and evaluation design.

\subsection{Practical Guidance for Method Selection}
\label{Practical Guidance for Method Selection}

The four methodological paradigms reviewed above differ substantially in their suitability across common real-world deployment conditions.
Table~\ref{tab:paradigm_comparison} provides a structured overview of how the paradigms compare across seven practical dimensions.
The following guidance translates these comparisons into actionable recommendations for three recurring deployment scenarios.

\textbf{When labeled data is limited.}
Statistical methods remain a viable baseline when only a few thousand annotated sentences are available, as hand-crafted features encode strong linguistic priors that reduce dependence on large corpora.
Graph-based methods with gold-standard parse trees offer a useful structural inductive bias that partially compensates for the absence of large training sets, though this advantage presupposes the availability of reliable parsers for the target language.
PLM-based neural methods are generally preferred when a moderate amount of task-specific data is available for fine-tuning, as they transfer broad linguistic knowledge from large unlabeled corpora.
Generative and LLM-based methods are most attractive when labeled data is extremely scarce, though as shown in Table~\ref{tab:res_glm}, few-shot LLM performance still falls substantially short of fully supervised specialized systems.
The relative ranking of paradigms under low-resource conditions is sensitive to the quality of available parse trees and the typological distance between the target language and the pretraining corpus of the PLM, so no single recommendation applies universally.

\textbf{When the target domain differs from training data.}
PLM-based neural methods show the strongest out-of-domain generalization among all paradigms, as reflected in the WSJ-to-Brown transfer results in Tables~\ref{tab:result-05} and~\ref{tab:result-09}.
In specialized domains such as biomedical or legal text, explicit syntactic features serve as a domain-agnostic inductive bias that can partially compensate for distribution mismatch, making graph-based methods a useful complement to PLM-based encoders.
Generative methods with retrieval-augmented prompting offer a promising direction for domain adaptation without retraining, though their structured output reliability remains lower than that of specialized systems, as discussed in \S\ref{Error Analysis}.
Statistical methods are least suitable for cross-domain deployment, as feature templates designed for one domain typically require substantial redesign when applied elsewhere.

\textbf{When inference efficiency is the primary constraint.}
CNN-based models offer the lowest inference latency among neural approaches due to their parallelizable convolution operations, making them well-suited for latency-sensitive pipelines.
Compact discriminative models based on BiLSTM or lightweight Transformer encoders provide a practical balance between accuracy and throughput for production deployments.
Graph-based methods introduce additional overhead from graph construction and message passing, which grows with sentence length and can become a bottleneck for long-document processing.
Sequence-to-sequence generative models and large LLMs are generally unsuitable for real-time applications without significant model compression, as discussed in \S\ref{Efficient and Deployable SRL}.

A recurring finding across all three scenarios is that no paradigm dominates on all dimensions simultaneously, as summarized in Table~\ref{tab:paradigm_comparison}.
Gains in accuracy typically come at the cost of efficiency or data requirements, and gains in generalization often require sacrificing output controllability or interpretability.
Practitioners should therefore treat method selection as a multi-objective decision that requires explicit prioritization of the constraints most relevant to their deployment context.

\section{Paradigm Modeling in SRL}
\label{Paradigm Modeling in SRL}

SRL can be decomposed into four fundamental subtasks: predicate detection, predicate disambiguation, argument identification, and argument classification.
Among these, predicate disambiguation refers to the identification of the correct sense of a predicate in PropBank, or the correct frame in FrameNet, which is relevant because the semantic interpretation of each argument label is conditioned on the predicate sense or frame respectively.
Two main paradigms for argument annotation have been developed: span-based and dependency-based SRL.

Span-based SRL (\S\ref{Span-based Methods}) identifies and labels text spans as complete semantic arguments.
Dependency-based SRL (\S\ref{Dependency-based Methods}) adopts a more concise representation by annotating only the syntactic head of each argument.
These paradigms differ not only in their representational granularity but also in the modeling assumptions they impose and the error types they are prone to.
Span-based annotation captures full argument boundaries, which is necessary for downstream tasks that require precise text extraction, but it introduces a harder search problem over all possible spans.
Dependency-based annotation reduces this search space by restricting predictions to head words, which simplifies decoding but loses boundary information that may be critical for certain argument types.
Neither paradigm strictly dominates the other across all evaluation conditions, and the choice between them should be guided by the requirements of the target application and the annotation scheme of the available training data.

\subsection{Span-based Methods}
\label{Span-based Methods}
\begin{table*}[ht]
\fontsize{9}{10}\selectfont
\setlength{\tabcolsep}{1.3mm}

\begin{center}
\resizebox{1\textwidth}{!}{
\begin{tabular}{lccccccccccccccccccccc}
\Xhline{0.08em}
\rowcolor{blue!15}
\multicolumn{22}{c}{\textit{\textbf{Span-based SRL}}} \\
\multirow{6}{*}{\bf Method} & \multirow{6}{*}{\bf PLM} & \multirow{6}{*}{\bf SYN} & \multirow{6}{*}{\bf E2E} & \multicolumn{9}{c}{\bf Without pre-identified predicates} & \multicolumn{9}{c}{\bf With pre-identified predicates}\\
\cmidrule(lr){5-13}\cmidrule(lr){14-22}
 &  &  &  & \multicolumn{6}{c}{\bf CoNLL05} & \multicolumn{3}{c}{\multirow{3}{*}{\bf CoNLL12}} & \multicolumn{6}{c}{\bf CoNLL05} & \multicolumn{3}{c}{\multirow{3}{*}{\bf CoNLL12}}\\
\cmidrule(lr){5-10}\cmidrule(lr){14-19}
& & & & \multicolumn{3}{c}{\bf WSJ} & \multicolumn{3}{c}{\bf Brown} & & & &  \multicolumn{3}{c}{\bf WSJ} & \multicolumn{3}{c}{\bf Brown} & & &\\
\cmidrule(lr){5-22}
 & & & & \bf P & \bf R & \bf F$_1$ & \bf P & \bf R & \bf F$_1$ & \bf P & \bf R & \bf F$_1$ & \bf P & \bf R & \bf F$_1$ & \bf P & \bf R & \bf F$_1$ & \bf P & \bf R & \bf F$_1$ \\
\hline
\citet{DBLP:conf/acl/ZhouX15} & N & & & 82.9 & 82.8 & 82.8 & 70.7 & 68.2 & 69.4 & - & - & 81.3 & - & - & - & - & - & - & - & - & - \\
\rowcolor{gray!15}
\citet{DBLP:conf/acl/HeLLZ17a} & N & N & Y & 82.0 & 83.4 & 82.7 & 69.7 & 70.5 & 70.1 & 80.2 & 76.6 & 78.4 & 85.0 & 84.3 & 84.6 & 74.9 & 72.4 & 73.6 & 83.5 & 83.3 & 83.4\\
\citet{DBLP:conf/aaai/TanWXCS18} & N & & & - & - & - & - & - & - & - & - & - & 85.9 & 86.3 & 86.1 & 74.6 & 75.0 & 74.8 & 83.3 & 84.5 & 83.9\\
\rowcolor{gray!15}
\citet{DBLP:conf/acl/HeLLZ18} & Y & N & & 84.8 & 87.2 & 86.0 & 73.9 & 78.4 & 76.1 & 81.9 & 84.0 & 82.9 & - & - & 87.4 & - & - & 80.4 & - & - & 85.5\\
\citet{DBLP:conf/emnlp/StrubellVAWM18} & Y & Y & & 84.0 & 83.2 & 83.6 & 73.3 & 70.6 & 71.9 & 81.9 & 79.6 & 80.7 & 84.7 & 84.2 & 84.5 & 73.9 & 72.4 & 73.1 & - & - & - \\
\quad+ELMo & Y & N & & 86.7 & 86.4 & 86.6 & 79.0 & 77.2 & 78.1 & 84.0 & 82.3 & 83.1 & 84.6 & 84.6 & 84.6 & 74.8 & 74.3 & 74.6 & - & - & -\\
\rowcolor{gray!15}
\citet{DBLP:conf/emnlp/OuchiS018} & & N & & - & - & - & - & - & - & - & - & - & 84.7 & 82.3 & 83.5 & 76.0 & 70.4 & 73.1 & 84.4 & 81.7 & 83.0\\
\rowcolor{gray!15}
\quad+ELMo & Y & N & & - & - & - & - & - & - & - & - & - & 88.2 & 87.0 & 87.6 & 79.9 & 77.5 & 78.7 & 87.1 & 85.3 & 86.2\\
\citet{DBLP:conf/emnlp/ZhouLZ20} & & Y & N & 83.7 & 85.5 & 84.6 & 72.0 & 73.1 & 72.6 & - & - & - & 85.9 & 85.8 & 85.8 & 76.9 & 74.6 & 75.7 & - & - & -\\
\quad+BERT & Y & Y & N & 85.3 & 87.7 & 86.5 & 76.1 & 78.3 & 77.2 & - & - & - & 87.8 & 88.3 & 88.0 & 79.6 & 78.6 & 79.1 & - & - & -\\
\rowcolor{gray!15}
\citet{DBLP:conf/acl/WangJWSW19} & Y & Y & & - & - & - & - & - & - & - & - & - & - & - & 88.2 & - & - & 79.3 & - & - & 86.4 \\
\citet{DBLP:conf/aaai/JiaYWT22} & & & & - & - & 84.5 & - & - & 72.7 & - & - & 81.6 & - & - & 86.2 & - & - & 75.6 & - & - & 84.9\\

\rowcolor{gray!15}
\citet{DBLP:conf/acl/FeiWRLJ21} &  & Y & & - & - & - & - & - & - & - & - & - & 87.2  & 87.6 & 87.3 & 78.7 &77.4 &78.1 & 86.5 & 85.9 & 86.2 \\
\rowcolor{gray!15} \quad+RoBERTa & Y & Y & & - & - & - & - & - & - & - & - & - & 88.8 & 89.3 & 89.0 & 83.5 & 83.8 & 83.7 & 88.1 & 88.8 & 88.6 \\

\bottomrule
\end{tabular}
}
\end{center}
\caption{
Performance comparison of span-based SRL systems on CoNLL-2005 English in-domain (WSJ), out-of-domain (Brown), and CoNLL-2012 test sets. A dash (-) indicates that the corresponding metric was not reported in the original paper. Results are not directly comparable across systems due to differences in experimental settings, including predicate identification strategy, syntactic feature source, and pretrained encoder version.
}
\label{tab:result-05}
\end{table*}

Span-based methods in SRL identify and classify continuous spans of text as semantic arguments.
The development has evolved from traditional sequence labeling approaches to advanced span-centric and graph-based architectures.
Table~\ref{tab:result-05} summarizes the performance of representative methods on standard benchmarks.
Results across systems are not directly comparable due to differences in predicate identification strategy, use of gold versus predicted syntax, and pretrained encoder version.
The scores reported in Table~\ref{tab:result-05} are drawn from the original publications and represent single-point estimates without variance information.
Given the heterogeneity of experimental configurations across studies, small numerical differences between systems should not be interpreted as reliable indicators of performance ranking.

The earliest approaches used sequence labeling.
\citet{DBLP:conf/conll/HaciogluPWMJ04} pioneered chunk-level sequence labeling for SRL, reducing the search space from word-level to chunk-level units.
\citet{DBLP:journals/tacl/TackstromG015} formulated span-based SRL as a constrained structured prediction problem with dynamic programming, while \citet{DBLP:conf/emnlp/FitzGeraldTG015} extended this within a factor graph framework by replacing hand-crafted linear feature functions with neural network factors.
A known limitation of BIO-based sequence labeling is that it encodes argument boundaries implicitly through tag transitions, which makes it difficult to recover non-contiguous spans and introduces label imbalance between boundary and interior tags.

Deep learning brought significant advances.
\citet{DBLP:conf/acl/HeLLZ17a} introduced a deep highway BiLSTM model with BIO tagging and constrained decoding.
\citet{DBLP:conf/acl/HeLLZ18} proposed jointly predicting all predicates and argument spans in a single end-to-end model via a span-graph formulation, removing the assumption of gold predicate input that constrained prior BIO-based systems.
BIO-based models remain sensitive to boundary errors: a single mispredicted tag can invalidate an entire argument span, and the constrained decoding required to enforce valid BIO sequences can add inference overhead in the worst case.

Recent research shifted towards direct span modeling.
\citet{DBLP:conf/emnlp/OuchiS018} proposed scoring all possible labeled spans directly, bypassing intermediate BIO tagging.
This approach avoids the cascading boundary error problem of sequence labeling, but it introduces a quadratic number of candidate spans with respect to sentence length, which creates a computational bottleneck for long sentences and requires aggressive pruning strategies that may discard correct arguments.
\citet{DBLP:conf/coling/ZhouXLZHZ22} reframed span-based SRL as word-based graph parsing, significantly reducing computational complexity while maintaining accuracy.

A consistent pattern across systems is the performance gap between WSJ and Brown corpus results, which persists even in strong PLM-based models and indicates that benchmark F$_1$ gains do not straightforwardly translate to robust out-of-domain performance.
This gap may be particularly pronounced for adjunct roles such as ARGM-TMP and ARGM-LOC, whose realization patterns vary more across domains than those of core arguments.
Tracing the quantitative progression within Table~\ref{tab:result-05} reveals a clear pattern of diminishing returns as representations grow stronger.
The base BiLSTM system of \citet{DBLP:conf/acl/HeLLZ17a} achieved 82.7 F$_1$ on CoNLL-2005 WSJ without any pretrained representations.
Adding ELMo to the LISA model of \citet{DBLP:conf/emnlp/StrubellVAWM18} lifted performance to 86.6, a gain of approximately 3 points over the same model without ELMo.
The transition from ELMo to BERT produced a further improvement of roughly 1 $\sim$ 2 points in comparable configurations, as seen in \citet{DBLP:conf/emnlp/ZhouLZ20} where adding BERT raised F$_1$ from 85.8 to 86.5 on the same system.
The adoption of RoBERTa over BERT in \citet{DBLP:conf/acl/FeiWRLJ21} yielded an additional gain of approximately 1.7 points on CoNLL-2005 WSJ, from 87.3 to 89.0.
These incremental gains stand in contrast to the larger jumps observed during the transition from statistical to neural methods, confirming that in-domain benchmark performance is approaching saturation.
On the CoNLL-2005 WSJ test set, top-performing systems have reached F$_1$ scores approaching 89, and the margin between leading systems has narrowed to a range where differences are unlikely to be meaningful given the variability introduced by different experimental configurations.
This saturation on in-domain evaluation further underscores the need to treat out-of-domain performance as a primary rather than secondary criterion when assessing system quality.
A further unresolved challenge concerns the handling of discontinuous spans, which occur most notably in FrameNet annotations but are not supported by standard span enumeration approaches.
Current span-based models uniformly assume contiguous argument boundaries, and no widely adopted method exists for recovering split or gapped argument spans within the span-based paradigm.
Span-based SRL has progressed from sequential labeling to span-centric and graph-based paradigms, yet the trade-off between boundary precision and computational cost remains an open problem without a universally satisfactory solution.

\subsection{Dependency-based Methods}
\label{Dependency-based Methods}

\begin{table*}[ht]
\fontsize{9}{10}\selectfont
\setlength{\tabcolsep}{1.3mm}

\begin{center}
\resizebox{1.0\textwidth}{!}{
\begin{tabular}{lccccccccccccccc}
\Xhline{0.08em}
\rowcolor{blue!15}
\multicolumn{16}{c}{\textit{\textbf{Dependency-based SRL}}} \\
\multirow{4}{*}{\bf Method} & \multirow{4}{*}{\bf PLM} & \multirow{4}{*}{\bf SYN} & \multirow{4}{*}{\bf E2E} & \multicolumn{6}{c}{\bf Without pre-identified predicates} & \multicolumn{6}{c}{\bf With pre-identified predicates}\\
\cmidrule(lr){5-10}\cmidrule(lr){11-16}
& & & & \multicolumn{3}{c}{\bf WSJ} & \multicolumn{3}{c}{\bf Brown} & \multicolumn{3}{c}{\bf WSJ} & \multicolumn{3}{c}{\bf Brown} \\
\cmidrule(lr){5-7}\cmidrule(lr){8-10}\cmidrule(lr){11-13}\cmidrule(lr){14-16}
 & & & & \bf P & \bf R & \bf F$_1$ & \bf P & \bf R & \bf F$_1$ & \bf P & \bf R & \bf F$_1$ & \bf P & \bf R & \bf F$_1$ \\
\hline
\citet{DBLP:conf/emnlp/ZhaoCK09} & & Y & & - & - & - & - & - & - & - & - & 85.4 & - & - & 73.3\\
\rowcolor{gray!15}
\citet{DBLP:conf/conll/ZhaoCKUT09} & & Y & & - & - & - & - & - & - & - & - & 86.2 & - & - & 74.6\\
\citet{DBLP:conf/naacl/LeiZVMB15} & & Y & & - & - & - & - & - & - & - & - & 86.6 & - & - & 75.6\\
\rowcolor{gray!15}
\citet{DBLP:conf/emnlp/FitzGeraldTG015} & & Y & & - & - & - & - & - & - & - & - & 87.8 & - & - & 75.5\\
\citet{DBLP:conf/acl/RothL16} & & Y & & - & - & - & - & - & - & 90.3 & 85.7 & 87.9 & 79.7 & 73.6 & 76.5 \\
\rowcolor{gray!15}
\citet{DBLP:conf/conll/SwayamdiptaBDS16} & & N & & - & - & 80.5 & - & - & - & - & - & 85.0 & - & - & -\\
\citet{DBLP:conf/acl/MulcaireSS18} & & N & & - & - & - & - & - & - & - & - & 87.2 & - & - & -\\
\rowcolor{gray!15}
\citet{DBLP:conf/naacl/KasaiFFRR19} & & Y & & - & - & - & - & - & - & 89.0 & 88.2 & 88.6 & 78.0 & 77.2 & 77.6\\
\rowcolor{gray!15}
\quad+ELMo & & Y & & - & - & - & - & - & - & 90.3 & 90.0 & 90.2 & 81.0 & 80.5 & 80.8\\
\citet{DBLP:conf/emnlp/ZhangWS19} & & Y & & - & - & - & - & - & - & 89.6 & 86.0 & 87.7 & - & - & - \\
\rowcolor{gray!15}
\citet{DBLP:conf/emnlp/LyuCT19}+ELMo & Y & N & & - & - & - & - & - & - & - & - & 91.0 & - & - & 82.2\\


\citet{DBLP:conf/acl/ZhaoHLB18} & & Y & N & 83.9 & 82.7 & 83.3 & - & - & - & 89.7 & 89.3 & 89.5 & 81.9 & 76.9 & 79.3\\
\rowcolor{gray!15}
\citet{DBLP:conf/emnlp/ZhouLZ20} & & Y & N & 84.2 & 87.5 & 85.9 & 76.5 & 78.5 & 77.5 & 88.7 & 89.8 & 89.3 & 82.5 & 83.2 & 82.8 \\
\rowcolor{gray!15}
\quad+BERT & Y & Y & N & 87.4 & 89.0 & 88.2 & 80.3 & 82.9 & 81.6 & 91.2 & 91.2 & 91.2 & 85.7 & 86.1 & 85.9 \\
\citet{DBLP:journals/taslp/MunirZL21}  & & Y & N & 85.8 & 84.4 & 85.1 & 74.6 & 74.8 & 74.7 & 91.2 & 90.6 & 90.9 & 83.1 & 82.6 & 82.8\\
 & & N & N & - & - & - & - & - & - & 88.7 & 86.8 & 87.7 & 79.4 & 76.2 & 77.7 \\
\rowcolor{gray!15}
\citet{DBLP:journals/coling/LiZHC21} & & Y & N & 86.2 & 86.0 & 86.1 & 73.8 & 74.6 & 74.2 & 90.5 & 91.7 & 91.1 & 83.3 & 80.9 & 82.1\\
\citet{DBLP:conf/coling/CaiHLZ18} & & N & Y & 84.7 & 85.2 & 85.0 & - & - & 72.5 & 89.9 & 89.2 & 89.6 & 79.8 & 78.3 & 79.0\\
\rowcolor{gray!15}
\citet{DBLP:conf/aaai/LiHZZZZZ19} + ELMo & & N & N & 84.5 & 86.1 & 85.3 & 74.6 & 73.8 & 74.2 & 89.6 & 91.2 & 90.4 & 81.7 & 81.4 & 81.5\\
\citet{DBLP:conf/emnlp/LiZWP20} & & Y & Y & 86.0 & 85.6 & 85.8 & 74.4 & 73.3 & 73.8 & 91.3 & 88.7 & 90.0 & 81.8 & 78.4 & 80.0\\
\quad+BERT & Y & Y & Y & 88.6 & 88.6 & 88.6 & 79.9 & 79.9 & 79.9 & 92.6 & 91.0 & 91.8 & 86.5 & 83.8 & 85.1\\
\rowcolor{gray!15}
\citet{DBLP:conf/aaai/XiaL0ZFWS19} & & Y & N & - & - & - & - & - & - & 90.3 & 85.7 & 87.9 & 79.7 & 73.6 & 76.5 \\
\citet{DBLP:conf/emnlp/MarcheggianiT17} & & Y & N & - & - & - & - & - & - & 90.5 & 87.7 & 89.1 & 80.8 & 77.1 & 78.9\\
\rowcolor{gray!15}
\citet{DBLP:conf/acl/ZhaoHLB18} & & Y & Y & - & - & - & - & - & - & 89.7 & 89.3 & 89.5 & 81.9 & 76.9 & 79.3\\ 
\citet{DBLP:journals/tacl/CaiL19} & & Y & N & - & - & - & - & - & - & 90.5 & 88.6 & 89.6 & 80.5 & 78.2 & 79.4\\
\quad+ELMo & & Y & N & - & - & - & - & - & - & 90.9 & 89.1 & 90.0 & 80.8 & 78.6 & 79.7\\
\rowcolor{gray!15}
\citet{DBLP:conf/emnlp/CaiL19} & & N & N & - & - & - & - & - & - & 91.1 & 90.4 & 90.7 & 82.1 & 81.3 & 81.6\\
\rowcolor{gray!15}
\quad+Semi & & N & N & - & - & - & - & - & - & 91.7 & 90.8 & 91.2 & 83.2 & 81.9 & 82.5\\
\citet{DBLP:conf/emnlp/HeLZ19} & & Y & N & - & - & - & - & - & - & 90.0 & 90.0 & 90.0 & – & – & –\\
\quad+BERT & Y & Y & N & - & - & - & - & - & - & 90.4 & 91.3 & 90.9 & – & – & –\\
\rowcolor{gray!15}
\citet{DBLP:conf/emnlp/ChenLT19} & Y & N & Y & - & - & - & - & - & - & 90.7 & 91.4 & 91.1 & 82.7 & 82.8 & 82.7\\
\citet{DBLP:journals/kbs/Fernandez-Gonzalez23} & & N & Y & 85.9 & 88.0 & 86.9 & 74.4 & 76.4 & 75.4 & 90.2 & 90.5 & 90.4 & 80.6 & 80.5 & 80.6\\
\quad+BERT & Y & N & Y & 87.2 & 89.8 & 88.5 & 79.9 & 81.8 & 80.4 & 91.3 & 91.6 & 91.4 & 84.5 & 84.2 & 84.4\\

\bottomrule
\end{tabular}
}
\end{center}
\caption{
Performance comparison of dependency-based SRL systems on CoNLL-2008/CoNLL-2009 English datasets.
Results are reported for in-domain (Wall Street Journal, WSJ) and out-of-domain (Brown) test sets, showing precision (P), recall (R), and F$_1$ scores.
A dash (-) indicates that the corresponding metric was not reported in the original paper.
Results are not directly comparable across systems due to differences in syntactic feature source, predicate identification strategy, and pretrained encoder version.
}
\label{tab:result-09}
\end{table*}

Dependency-based SRL methods have undergone significant evolution in modeling predicate-argument relationships.
Table~\ref{tab:result-09} summarizes performance of mainstream approaches on the CoNLL-2009 benchmark.
Results across systems are not directly comparable due to differences in syntactic feature source, predicate identification strategy, and pretrained encoder version.
As with the span-based results in Table~\ref{tab:result-05}, the scores reported here are single-point estimates compiled from independent publications with heterogeneous experimental setups, and variance information is not available from the original sources.
Numerical differences between systems should therefore be read as approximate indicators of relative standing rather than precise performance rankings.

Early systems adopted pipeline architectures treating syntactic parsing and SRL as separate tasks \citep{DBLP:conf/acl/PradhanWHMJ05}.
\citet{DBLP:conf/acl/SwansonG06} further showed that parse tree path features from dependency parsers were substantially less reliable than those from constituency parsers, highlighting the limitations of dependency-based feature extraction at the time.
\citet{DBLP:conf/emnlp/JohanssonN08} proposed a joint syntactic-semantic system that selects the final output via reranking over candidates generated by separate syntactic and semantic submodels.
This work demonstrated for the first time that dependency-based PropBank SRL could rival constituent-based systems in performance.
A persistent limitation of pipeline architectures is that parsing errors propagate directly into role labeling with no mechanism for downstream correction, and the severity of this error propagation varies across languages depending on parser quality and the degree of syntactic-semantic alignment in the target language.

Neural networks enabled end-to-end learning paradigms.
BiLSTM architectures captured long-range dependencies and enabled syntax-agnostic modeling, as demonstrated by \citet{DBLP:conf/conll/MarcheggianiFT17}, who showed that competitive dependency-based SRL performance could be achieved without any syntactic parse information.
This result challenged the prevailing assumption that syntactic parsing was indispensable for dependency-based SRL, and showed that competitive performance could be achieved through direct mapping from raw input to semantic dependencies.

Graph-based neural models introduced structural inductive biases that improved predicate-argument modeling beyond what sequential encoders could achieve.
\citet{DBLP:conf/emnlp/MarcheggianiT17} applied GCNs to encode syntactic dependency structures as labeled graphs.
\citet{DBLP:conf/coling/CaiHLZ18} proposed the first end-to-end model for dependency-based SRL that jointly handles predicate disambiguation and argument labeling in a single pass via a biaffine attentional scorer over all word pairs, without relying on any syntactic parse information.
Further innovations integrated self-attention with structural supervision, as represented by the SynG2G-Tr model \citep{mohammadshahi-henderson-2023-syntax} and high-order dependency modeling \citep{DBLP:conf/emnlp/LiZWP20}, where high-order dependencies refer to predicate-argument interactions that extend beyond direct head-argument pairs to capture relations among multiple arguments of the same predicate or across nested argument structures.
A notable tension in this line of work is that increasing model order improves coverage of complex argument structures but raises inference complexity, and the practical gains from high-order modeling diminish as the base encoder grows stronger through pretraining.
Recently, \citet{DBLP:conf/emnlp/ShiMI20} proposed modeling semantic roles directly within dependency structures, treating them as enriched syntactic dependencies in unified representations capturing both syntactic and semantic relationships.
The specific contributions of syntactic features across different model architectures are analyzed in Section~\ref{Syntax Feature Modeling in SRL}.

The consistent WSJ-to-Brown performance gap across all systems indicates that in-domain F$_1$ alone provides an incomplete picture of model capability, and out-of-domain evaluation should be treated as a standard requirement rather than an optional supplement.
A further unresolved challenge concerns the head-word representation assumption underlying dependency-based SRL: by annotating only the syntactic head of each argument, this paradigm loses the boundary information needed for downstream tasks that require full argument spans, and the mapping from head words back to full spans is non-trivial for arguments with complex internal structure.
Examining the quantitative progression in Table~\ref{tab:result-09} reveals a trend consistent with that observed in span-based evaluation.
Pre-neural systems with rich syntactic features achieved F$_1$ scores in the range of 85 $\sim$ 88 on CoNLL-2009 WSJ \citep{DBLP:conf/emnlp/ZhaoCK09, DBLP:conf/naacl/LeiZVMB15, DBLP:conf/emnlp/FitzGeraldTG015}.
BiLSTM and GCN-based neural models without pretraining advanced this to approximately 89 $\sim$ 91 \citep{DBLP:conf/emnlp/MarcheggianiT17, DBLP:conf/coling/CaiHLZ18, DBLP:conf/emnlp/CaiL19}.
Adding ELMo produced gains of roughly 1 $\sim$ 2 points in comparable configurations, as seen in \citet{DBLP:conf/naacl/KasaiFFRR19} where ELMo raised F$_1$ from 88.6 to 90.2 on CoNLL-2009 WSJ.
The introduction of BERT yielded a further improvement of approximately 1 $\sim$ 3 points depending on the base system, with \citet{DBLP:conf/emnlp/LiZWP20} reporting an increase from 85.8 to 88.6 in the end-to-end (without pre-identified predicates) setting when BERT was added to the same architecture.
On the CoNLL-2009 WSJ test set, leading systems have converged to F$_1$ scores in the low nineties, and the benchmark shows signs of saturation similar to those observed in span-based evaluation.
Progress on this benchmark is increasingly difficult to interpret without accompanying out-of-domain and cross-lingual results that can confirm whether gains reflect genuine improvements in semantic understanding.
The field has progressed from pipeline and joint modeling to neural, graph-based, and unified approaches, yet no existing system fully resolves the tension between representational conciseness and the completeness of semantic information required by real-world applications.

\subsection{Comparison of Span and Dependency-based Paradigms}
\label{Paradigm Comparison}

The two paradigms differ along several dimensions that have direct consequences for system design and downstream applicability.
Table~\ref{tab:result-05} and Table~\ref{tab:result-09} report results on CoNLL-2005 and CoNLL-2009 respectively, but direct cross-paradigm comparison from these tables is complicated by differences in annotation scheme, evaluation metric, and benchmark corpus.
CoNLL-2005 uses span-based PropBank annotations evaluated by exact span matching, while CoNLL-2009 uses dependency-based annotations across multiple languages evaluated by labeled head-word attachment, making F$_1$ scores across the two tables measure fundamentally different aspects of prediction quality.

Span-based methods are better suited to tasks that require precise argument boundaries, such as information extraction and reading comprehension, where the exact text span of an argument must be returned to the user.
Dependency-based methods are more appropriate when the goal is to identify semantic participants rather than their precise textual extent, such as in semantic parsing or knowledge graph construction, where head-word identification is sufficient.

A recurring finding in the literature is that unified models trained on both paradigms simultaneously can achieve competitive performance on each, suggesting that the two representations may capture complementary rather than purely redundant aspects of predicate-argument structure.
However, the conditions under which joint training helps versus hurts remain poorly understood, and results vary across languages and annotation schemes in ways that have not been systematically explained.
This represents an unresolved challenge for future work: a principled account of when and why the two paradigms benefit from joint modeling would provide clearer guidance for system design across diverse deployment settings.

\section{Syntax Feature Modeling in SRL}
\label{Syntax Feature Modeling in SRL}

The modeling of syntactic and linguistic features has been a central topic in SRL research for many years.
The core intuition is that syntactic structures provide essential cues for uncovering predicate-argument relationships within a sentence, and thus, effective SRL systems should leverage this structural information.
Early methods adopted a \textit{syntax-aided} approach, incorporating syntactic information through manually designed feature templates such as phrase types, syntactic paths, and positional relations to guide semantic role identification.
With neural network-based models, there has been a shift towards \textit{syntax-free} approaches that automatically learn relevant features from raw text, reducing the need for explicit syntax-level feature engineering.
This transition has sparked ongoing investigation into the optimal integration of syntactic information in modern SRL systems.
The central question is not simply whether syntax helps, but rather \textit{when} and \textit{why} it helps: the answer depends on factors including model architecture, data availability, language type, and the quality of available syntactic parsers.

\begin{figure}
    \centering
    \includegraphics[width=\linewidth]{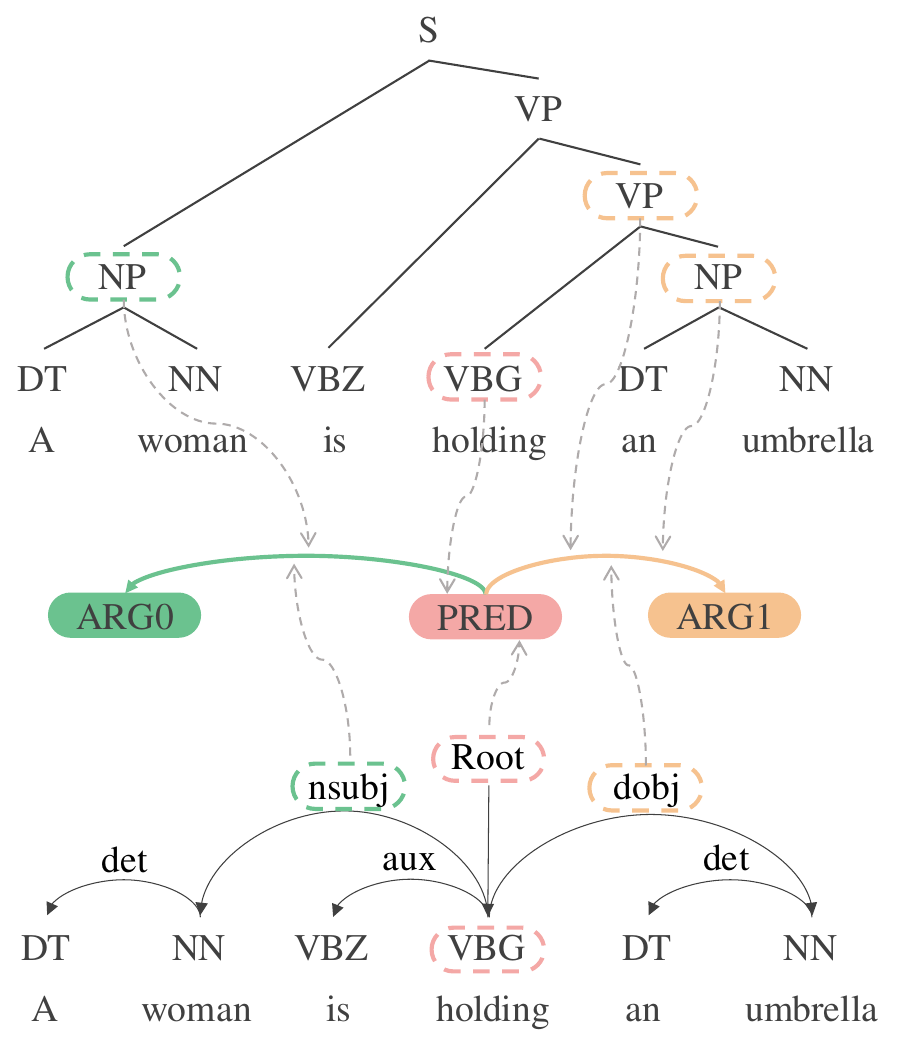}
    \caption{Illustration of how syntactic structural features contribute to SRL.}
    \label{fig:syntax}
\end{figure}

\subsection{Syntax-aided SRL}
Early research on SRL placed great emphasis on feature engineering, with syntactic information, especially syntactic tree features, proving to be particularly beneficial for SRL performance.
As illustrated in Figure~\ref{fig:syntax}, syntactic structures such as phrase structure trees and dependency relations provide valuable cues for aligning predicates with their arguments, motivating the integration of syntactic features into SRL systems.
\citet{DBLP:conf/acl/GildeaP02} pioneered the use of chunk-based systems, taking the last word of each chunk as the head word for role prediction.
Subsequently, \citet{DBLP:conf/acl/PradhanWHMJ05} systematically compared different syntactic feature views and showed that integrating multiple syntactic representations, including constituent, dependency, and chunk-based forms, significantly improves syntax-aided SRL performance by leveraging their complementary strengths.
Among these feature types, dependency-based features provide the most compact and direct encoding of predicate-argument relations in neural models, constituency-based features contribute complementary span-level boundary information for argument identification, and syntactic path features are particularly effective for capturing long-distance predicate-argument relations.
\citet{DBLP:conf/conll/KoomenPRY05} leveraged full parsing information to build more robust SRL systems, while~\citet{DBLP:conf/ijcai/PunyakanokRY05} and~\citet{DBLP:journals/coling/PunyakanokRY08} provided empirical evidence that syntactic parsing and inference are crucial for accurate SRL, proposing joint inference frameworks to integrate syntactic and semantic predictions.
\citet{DBLP:journals/coling/ToutanovaHM08} proposed global joint models for SRL that leverage syntactic structure to capture argument dependencies and enforce global consistency, demonstrating the crucial role of syntax in syntax-aided SRL.

With the advent of deep learning, researchers explored new ways to incorporate syntactic information into neural SRL models.
\citet{DBLP:conf/emnlp/MarcheggianiT17} introduced GCNs to encode syntactic dependency structures, demonstrating significant improvements in neural SRL.
\citet{DBLP:conf/acl/ZhaoHLB18} systematically quantified the contribution of syntax in neural SRL, and proposed a syntax-aware model using k-order argument pruning, demonstrating the effectiveness of syntax-aided SRL.
\citet{DBLP:conf/emnlp/StrubellVAWM18} presented LISA, which incorporates syntactic information by supervising a self-attention head to attend to syntactic parents within a multi-task learning framework, effectively improving SRL performance.
Building on these advances, \citet{DBLP:conf/acl/WangJWSW19} proposed a syntax-aided SRL method that leverages multiple candidate syntactic parses to improve robustness and performance.
\citet{DBLP:conf/acl/FeiWRLJ21} further explored the integration of heterogeneous syntactic information by combining both dependency and constituency representations, showing that jointly encoding complementary syntactic feature types yields stronger performance than relying on a single syntactic formalism.
More recently, \citet{DBLP:conf/acl/ChenHM22} advanced the field by modeling syntactic relations with relation-aware attention, showing that explicit modeling of syntactic dependencies can further enhance SRL performance.

The conditions under which these syntactic strategies provide the most reliable gains are examined systematically in Section~\ref{sec:when_and_why}.

\subsection{Syntax-free SRL}

The advent of neural networks, especially deep learning, has made syntax-free SRL possible by enabling automatic feature extraction without explicit syntactic input.
\citet{DBLP:journals/jmlr/CollobertWBKKK11} were the first to propose a syntax-free SRL system using a multilayer neural network, though their performance initially lagged behind state-of-the-art syntax-aided approaches.
Subsequent advances rapidly closed this gap: \citet{DBLP:conf/acl/ZhouX15} introduced a deep BiLSTM-based span SRL model, and \citet{DBLP:conf/conll/MarcheggianiFT17} developed a dependency-based syntax-free SRL model, both achieving results comparable to syntax-aided methods.
\citet{DBLP:conf/acl/HeLLZ17a} further advanced end-to-end SRL by showing that deep architectures such as multi-layer highway BiLSTMs can achieve strong performance even without explicit syntactic features, demonstrating the practical viability of syntax-free SRL.
Later, models such as the span selection approach by~\citet{DBLP:conf/emnlp/OuchiS018}, the BiAffine model by~\citet{DBLP:conf/coling/CaiHLZ18}, and the unified span-dependency framework by~\citet{DBLP:conf/aaai/LiHZZZZZ19} continued to push the performance of syntax-free SRL systems.
Moreover, \citet{DBLP:conf/emnlp/StrubellVAWM18} showed that self-attention mechanisms could support syntax-free modeling by capturing long-range dependencies directly.
Systematic studies by~\citet{DBLP:journals/coling/LiZHC21} revealed that the earlier underperformance of syntax-aided models was largely due to the limitations of automatic syntactic parsers, and that with gold-standard syntax, explicit syntactic features still offer clear improvements.

The rise of large-scale PLMs has changed the role of explicit syntax in SRL, but has not eliminated it.
PLMs trained on massive unlabeled corpora appear to encode substantial syntactic knowledge implicitly, which reduces the performance gap between syntax-aided and syntax-free systems in standard benchmarks.
However, this observation does not generalize uniformly across all settings.
When automatic parse quality is low, when labeled data is scarce, or when the target domain differs substantially from pretraining data, syntax-aided approaches continue to provide measurable gains.
End-to-end syntax-free SRL has become increasingly competitive in high-resource, standard-domain settings, yet the question of whether implicit syntactic knowledge in PLMs is sufficient for all SRL scenarios remains open and warrants continued investigation.

\subsection{When and Why Syntactic Features Help}
\label{sec:when_and_why}

The preceding discussion reveals that the value of syntactic features in SRL is neither absolute nor uniform, but is instead conditioned on several interacting factors.
We summarize the key conditions under which explicit syntax provides consistent benefits.

\textbf{Parser quality.}
The most direct factor is the quality of the syntactic parser used.
\citet{DBLP:journals/coling/LiZHC21} demonstrated that syntax-aided models consistently outperform syntax-free counterparts when gold-standard parse trees are available.
The performance gap narrows or reverses when automatic parsers introduce noise, which explains much of the inconsistency observed across studies.
This sensitivity is not uniform across role types: core arguments such as ARG0 and ARG1 tend to be relatively robust to parser noise due to their regular syntactic realization, whereas adjunct roles such as ARGM-TMP and ARGM-LOC, as well as semantically complex roles such as ARGM-CAU, tend to be more susceptible to parse errors because their boundaries rely more heavily on accurate structural paths.

\textbf{Model architecture.}
Syntax is most beneficial when the underlying model lacks the capacity to capture structural dependencies on its own.
Earlier CNN and RNN architectures benefited substantially from syntax-aided approaches because they could not model long-range structural relations directly.
Transformer-based models and PLMs reduce this dependency by attending across the full sequence, but they do not fully replicate the role of structured syntactic supervision in all cases.

\textbf{Data availability and domain.}
In low-resource settings or specialized domains such as biomedical or legal text, explicit syntactic features serve as a useful inductive bias that compensates for limited training data.
In high-resource, general-domain settings, the advantage of explicit syntax over strong PLM baselines is smaller but not absent.
The marginal value of syntactic supervision increases as labeled SRL data decreases, suggesting that syntax-aware models are most beneficial precisely in the scenarios where annotation is hardest to obtain.

\textbf{Language typology.}
Languages with flexible word order or rich morphology tend to benefit more from explicit syntactic features, as surface word order alone provides weaker cues for predicate-argument structure.
Cross-lingual transfer experiments have similarly shown that syntactic representations can improve robustness when transferring across typologically distant languages.

\textbf{Contradictions in reported results.}
A notable source of inconsistency across the SRL literature is the conflation of gold-standard and predicted syntactic input in experimental comparisons.
Several studies reporting strong gains from syntax-aided models used gold parse trees at test time, while syntax-free baselines operated on raw text.
This asymmetry inflates the apparent benefit of syntactic features and makes cross-study comparisons unreliable.
A further source of contradiction is the variation in which syntactic formalism is used: models incorporating constituency parses, dependency parses, and shallow chunks have each been reported as most beneficial in different studies, yet these claims are rarely evaluated under controlled conditions that hold all other variables constant.
Resolving these contradictions requires standardized evaluation protocols that distinguish clearly between gold and predicted syntax, and that report results under both conditions.

In summary, the syntax-aided versus syntax-free debate is best understood not as a binary choice but as a question of how to integrate structural information most effectively given the available resources and target conditions.
Future work should focus on developing adaptive integration strategies that can leverage syntactic information selectively, rather than treating it as uniformly beneficial or uniformly redundant.

\section{SRL under Various Scenarios}
\label{SRL under Various Scenarios}

As traditional SRL focuses on single-sentence, monolingual scenarios, the continuous development of natural language understanding requires more sophisticated approaches to deal with complex real-world applications.
This subsection explores three important extensions to SRL under various scenarios: SRL beyond single sentence (\S\ref{SRL Beyond Single Sentence}), multi-lingual/cross-lingual processing (\S\ref{Multi-lingual and Cross-lingual}),
and multi-modal type (\S\ref{Multi-modal SRL}).

\subsection{SRL Beyond Single Sentence}
\label{SRL Beyond Single Sentence}

\begin{table*}[ht]
\fontsize{6}{7}\selectfont
\setlength{\tabcolsep}{0.55mm}

\begin{center}
\resizebox{1.0\textwidth}{!}{
\begin{NiceTabular*}{\textwidth}{@{\extracolsep{\fill}}lccccccccc}[
    code-before = \rowcolor{blue!15}{1} \rowcolor{gray!15}{6,8,10,14,15}
]
\Xhline{0.08em}

\multicolumn{10}{c}{\textit{\textbf{Conversation SRL}}} \\

\multirow{2}{*}{\bf Method} & \multicolumn{3}{c}{\bf DuConv} & \multicolumn{3}{c}{\bf NewsDialog} & \multicolumn{3}{c}{\bf PersonalDialog} \\
\cmidrule{2-4}\cmidrule{5-7}\cmidrule{8-10}
& \bf F$_{1\_all}$ & \bf F$_{1\_cross}$ & \bf F$_{1\_intro}$ & \bf F$_{1\_all}$ & \bf F$_{1\_cross}$ & \bf F$_{1\_intro}$ & \bf F$_{1\_all}$ & \bf F$_{1\_cross}$ & \bf F$_{1\_intro}$\\
\hline
SimplePLM \citep{DBLP:conf/ijcai/0001WZRJ22}  & 86.54 & 81.62 & 87.02 & 77.68 & 51.47 & 80.99 & 66.53 & 30.48 & 70.00\\
\quad+CoDiaBERT & 88.40 & 82.96 & 88.25 & 79.42 & 53.46 & 82.77 & 68.86 & 33.75 & 72.23\\
CSRL \citep{DBLP:journals/taslp/XuWSZSY21}        & 88.46 & 81.94 & 89.46 & 78.77 & 51.01 & 82.48 & 68.46 & 32.56 & 72.02 \\
DAP \citep{DBLP:conf/acl/WuXSJZS20}        & 89.97 & 86.68 & 90.31 & 81.90 & 56.56 & 84.56 & - & - & - \\
CSAGN \citep{DBLP:conf/emnlp/WuXS21}      & 89.47 & 84.57 & 90.15 & 80.86 & 55.54 & 84.24 & 71.82 & 36.89 & 75.46\\
UE2E \citep{DBLP:conf/aaai/LiHZZZZZ19}       & 87.46 & 81.45 & 89.75 & 78.35 & 51.65 & 82.37 & 67.18 & 30.95 & 72.15\\
LISA \citep{DBLP:conf/emnlp/StrubellVAWM18}       & 89.57 & 83.48 & 91.02 & 80.43 & 53.81 & 85.04 & 70.27 & 32.48 & 75.70\\
SynGCN \\ \citep{DBLP:conf/emnlp/MarcheggianiT17}     & 90.12 & 84.06 & 91.53 & 82.04 & 54.12 & 85.35 & 70.65 & 34.85 & 76.96\\
\quad+CoDiaBERT & 91.34 & 86.72 & 91.86 & 82.86 & 56.75 & 85.98 & 72.06 & 37.76 & 77.41\\
POLar \citep{DBLP:conf/ijcai/0001WZRJ22}      & 92.06 & 90.75 & 92.64 & 83.45 & 60.68 & 87.96 & 73.46 & 40.97 & 78.02\\
\quad+CoDiaBERT & 93.72 & 92.86 & 93.92 & 85.10 & 63.85 & 88.23 & 76.61 & 45.47 & 78.55

\end{NiceTabular*}
}

\resizebox{\textwidth}{!}{
\begin{NiceTabular*}{\textwidth}{@{\extracolsep{\fill}}lccccccccc}[
    code-before = \rowcolor{blue!15}{1} \rowcolor{gray!15}{5,6,9,10,11}
]
\specialrule{.2em}{.05em}{0.05em} 
\multicolumn{10}{c}{\textit{\textbf{Discourse SRL}}} \\

\multirow{2}{*}{\bf Method} & \multirow{2}{*}{\bf NI Acc.} & \multirow{2}{*}{\bf DNI vs. INI Acc.} & \multirow{2}{*}{\bf absolute NI Acc.} & \multicolumn{3}{c}{\bf Full NI resolution} & \multicolumn{3}{c}{\bf DNI Linking}\\
\cmidrule{5-7}\cmidrule{8-10}
 & & & & \bf P & \bf R & \bf F$_1$ & \bf P & \bf R & \bf F$_1$\\
\hline
\citet{DBLP:journals/lre/RuppenhoferLSM13} & 8.0 & 64.2 & 5.0 & - & - & 1.4 & - & - & -\\
SEMAFOR \\ \citep{DBLP:journals/lre/RuppenhoferLSM13} & 63.4 & 54.7 & 35.0 & - & - & 1.2 & - & - & -\\
VENSES++ \\ \citep{DBLP:conf/semeval/TonelliD10} & 54.0 & 75.0 & 40.0 & 13.0 & 6.0 & 8.0 & - & - & -\\
S\&F \\ \citep{DBLP:conf/starsem/SilbererF12} & 58.0 & 68.0 & 40.0 & 6.0 & 8.9 & 7.1 & 25.6 & 25.1 & 25.3\\
\quad+bset heuristic data & 56.0 & 69.0 & 38.0 & 9.2 & 11.2 & 10.1 & 30.8 & 25.1 & 27.7\\
\citet{DBLP:conf/iwcs/MoorRF13} & - & - & - & - & - & - & 21.7 & 21.2 & 21.5 \\
\quad+feature selection & - & - & - & - & - & - & 33.3 & 22.0 & 26.5\\
\quad+full frame annotation & - & - & - & - & - & - & 34.3 & 26.3 & 29.8\\
\bottomrule

\end{NiceTabular*}
}
\end{center}
\caption{
Performance comparison of conversation SRL methods on DuConv, NewsDialog, and PersonalDialog datasets, and discourse SRL results on SemEval 2010 Task 10. ``NI Acc.'': null instantiation accuracy, ``DNI vs. INI Acc.'': relative accuracy of NI identification and interpretation, ``absolute NI Acc.'': absolute NI recognition accuracy, ``DNI Linking'': definite role antecedents linking. A dash (-) indicates that the corresponding metric was not reported in the original paper.
}

\label{tab:res_diag}
\end{table*}

SRL has traditionally been viewed as a sentence-level task.
However, this local view potentially misses important information that can only be recovered if local argument structures are linked across sentence boundaries.
From the very beginning, \citet{fillmore2001frame} had analyzed the frame semantics in articles.
Following, \citet{burchardt2005building} provided a detailed analysis of links between the local semantic argument structures in a short text.
\citet{DBLP:journals/lre/RuppenhoferLSM13} for the first time provided a new task to expand the SRL to the discourse dimension, which was published in the SemEval 2010 (Task-10).
\citet{DBLP:journals/coling/RothF15} pointed out and studied the phenomenon of implicit arguments and their respective antecedents in discourse, introducing an annotated corpus and a novel SRL framework.
Similar research on implicit SRL had also been conducted by~\citet{DBLP:conf/acl/LaparraR13,DBLP:conf/naacl/SchenkC16,DBLP:conf/ijcnlp/DoBM17}.

On the other hand, conversation is another cross-sentence SRL scenario.
The ellipsis and anaphora frequently occur in dialogues and the predicate-argument structures must be handled over the entire conversation.
\citet{DBLP:journals/taslp/XuWSZSY21} proposed the Conversational SRL (CSRL) for the first time.
They manually collected a CSRL dataset on the DuConv corpus and presented a BERT-based model to achieve the CSRL parsing.
\citet{DBLP:conf/icmlc2/HeWLSC21} further proposed to incorporate external knowledge into CSRL model to help capture and correlate the entities.
\citet{DBLP:conf/naacl/WuT0LWS22} explored the CSRL on few-shot and cross-lingual setting.
\citet{DBLP:conf/ijcai/0001WZRJ22} investigated the integration of a latent graph for CSRL, enhancing structural information integration, near-neighbor influence.
\citet{DBLP:journals/taslp/WuXS24} proposed to build structure-aware features to model the inter-speaker dependency and correlation of the predicates and the context utterances.
In Table~\ref{tab:res_diag}, we compare the SRL results on discourse and conversation scenarios.
A consistent finding in discourse SRL systems is that null instantiation and implicit argument recovery remain far below the performance levels achieved on explicit, within-sentence arguments, as reflected in Table~\ref{tab:res_diag}.
In contrast, conversational SRL systems achieve substantially higher overall F$_1$ scores on explicit argument identification, though cross-sentence argument recovery remains challenging.
This gap in discourse SRL is not primarily attributable to model capacity: even strong PLM-based systems fail to reliably resolve implicit arguments, suggesting that the challenge lies in the absence of explicit surface cues rather than in representational limitations.
A further unresolved challenge concerns evaluation methodology: most discourse SRL benchmarks are small and domain-specific, making it difficult to distinguish genuine generalization from overfitting to annotation idiosyncrasies.
Conversational SRL introduces the additional complication that ellipsis and anaphora resolution are prerequisites for accurate role labeling, yet these subtasks are typically handled by separate upstream modules whose errors propagate into the SRL output.
Whether end-to-end training can effectively learn to resolve these interdependencies without explicit intermediate supervision remains an open question.

\begin{table*}[ht]
\fontsize{9}{10}\selectfont
\setlength{\tabcolsep}{1.3mm}

\begin{center}
\begin{NiceTabular*}{\textwidth}{@{\extracolsep{\fill}}lcccccc}[
    code-before = \rowcolor{blue!15}{1} \rowcolor{gray!15}{4,6,8}
]
\Xhline{0.08em}
\multicolumn{7}{c}{\textit{\textbf{Cross-lingual SRL (from English to others)}}} \\

\bf Method & \bf DE & \bf FR & \bf IT & \bf ES & \bf PT & \bf FI \\
\hline
\citet{DBLP:conf/acl/FeiZJ20} & 65.0 & 64.8 & 58.7 & 62.5 & 56.0 & 54.5\\
\citet{DBLP:conf/emnlp/ZhangSH21} & 63.8 & 56.4 & 62.0 & 56.8 & 59.8 & 55.3 \\
\citet{DBLP:journals/taslp/FeiZLJ20} & 58.9 & 69.4 & 61.1 & 59.5 & 53.8 & 46.5\\
\citet{DBLP:conf/acl/HeLLZ17a} & 56.1 & 66.1 & 57.5 & 56.1 & 51.3 & 42.6 \\
\citet{DBLP:conf/emnlp/StrubellVAWM18} & 57.9 & 68.0 & 59.9 & 58.2 & 50.7 & 44.3 \\
\citet{DBLP:conf/acl/HeLLZ18} & 58.3 & 68.4 & 58.8 & 58.0 & 51.9 & 45.2 \\
\end{NiceTabular*}

\begin{NiceTabular*}{\textwidth}{@{\extracolsep{\fill}}lccccccc}[
    code-before = \rowcolor{blue!15}{1} \rowcolor{gray!15}{4,6,8}
]
\specialrule{.2em}{.05em}{0.05em} 
\multicolumn{8}{c}{\textit{\textbf{Multi-lingual SRL}}} \\

\bf Method & \bf CA & \bf CS & \bf DE & \bf EN & \bf JA & \bf ES & \bf ZH \\
\hline
\citet{DBLP:conf/conll/MarcheggianiFT17} & - & 86.0 & - & 87.6 & - & 80.3 & 81.2 \\
\citet{DBLP:conf/emnlp/MarcheggianiT17} & - & - & - & 89.1 & - & - & 82.5  \\
\citet{DBLP:conf/conll/ZhaoCKUT09} & 80.3 & 85.2 & 76.0 & 86.2 & 78.2 & 80.5 & 77.7 \\
\citet{DBLP:conf/acl/RothL16} & - & - & 80.1 & 87.7 & - & 80.2 & 79.4\\
\citet{DBLP:conf/naacl/KasaiFFRR19} & - & - & - & 90.2 & - & 83.0 & \\
\citet{DBLP:conf/acl/MulcaireSS18} & 77.31 & 84.87 & 66.71 & 86.54 & 74.99 & 75.98 & 81.26 \\
\citet{DBLP:conf/emnlp/HeLZ19} & 85.1 & 89.7 & 80.9 & 90.9 & 83.8 & 84.6 & 86.4 \\
\bottomrule
\end{NiceTabular*}
\end{center}
\caption{
Performance comparison of cross-lingual and multilingual semantic role labeling systems. A dash (-) indicates that the corresponding metric was not reported in the original paper.
}
\label{tab:res_multilingual}
\end{table*}

\subsection{Multi-lingual and Cross-lingual SRL}
\label{Multi-lingual and Cross-lingual}
Early SRL benchmarks such as FrameNet and PropBank-v1 were built with only English corpus.
To facilitate SRL studies on other languages, several attempts to build non-English SRL datasets, including Chinese~\citep{DBLP:conf/acl-sighan/XueP03}, French~\citep{DBLP:conf/taln/PadoP07,DBLP:conf/acl/PlasMH11}, German~\citep{DBLP:conf/acl/PadoL06}, and Italian~\citep{DBLP:conf/lrec/TonelliP08, DBLP:conf/cicling/BasiliCCCM09, DBLP:conf/cicling/AnnesiB10}, via cross-lingual annotation projection.

Cross-lingual SRL approaches can be broadly categorized into three main strategies: annotation projection, model transfer, and translation-based approaches.
For annotation projection, \citet{DBLP:conf/coling/PlasAC14} introduced a global approach that aggregates information at the corpus level, showing improved robustness to non-literal translations and alignment errors.
However, annotation projection heavily relies on word-aligned parallel data and is sensitive to both parallel data quality and source-language model accuracy.

In model transfer approaches, \citet{DBLP:conf/acl/KozhevnikovT13} pioneered an unsupervised transfer method that achieved competitive performance against annotation projection baselines.
Recent work by~\citet{DBLP:conf/emnlp/DeviantiM24} has further advanced this direction by investigating the transferability of SRL across typologically diverse languages, providing insights into which linguistic features facilitate or hinder cross-lingual transfer.

Translation-based approaches represent the third major strategy for cross-lingual SRL.
These approaches offer the potential for high cross-lingual transfer capabilities.
The primary objective of translation-based approaches is to leverage high-quality source-language annotations by translating them directly into the target language, thereby eliminating the need for additional target-language labeling.
\citet{DBLP:conf/emnlp/DazaF19} introduced a cross-lingual encoder-decoder model that simultaneously translates and generates sentences with SRL annotations in the target language.
Similar translation-based methods were also presented by~\citet{DBLP:conf/acl/FeiZJ20}.

Building on the work from CoNLL 2009 \citep{DBLP:conf/conll/HajicCJKMMMNPSSSXZ09}, multilingual benchmarks have since been developed at scale.
\citet{DBLP:conf/emnlp/JohannsenAS15} presented a multilingual FrameNet SRL dataset and a multilingual semantic parser with truly interlingual representation.
With the mixed multilingual resources, \citet{DBLP:conf/acl/MulcaireSS18} further built a polyglot SRL system combining data from multiple languages.
In Table~\ref{tab:res_multilingual}, we compare the SRL results on cross-lingual and multilingual scenarios.
Across the three cross-lingual strategies, data efficiency varies considerably: annotation projection requires large parallel corpora, model transfer requires no target-language annotations but is sensitive to typological distance, and translation-based methods allow reuse of source-language annotations without additional target-language labeling.
The development of Universal PropBank \citep{DBLP:conf/lrec/JindalRULNT0022} across 23 languages provides a valuable resource for studying how annotation volume affects cross-lingual transfer across typologically diverse language pairs.
A recurring contradiction in the cross-lingual SRL literature concerns the relative ranking of the three transfer strategies.
Annotation projection has been reported as superior in some studies but inferior in others, and the direction of this inconsistency correlates with parallel corpus size and source-language SRL model quality rather than with any intrinsic property of the projection method itself.
Model transfer results are similarly inconsistent across language pairs: transfer works well between typologically similar languages but degrades substantially for morphologically rich or low-resource target languages, yet the threshold at which typological distance becomes prohibitive has not been systematically established.
These inconsistencies make it difficult to provide universal guidance on strategy selection, and they point to the need for controlled comparative studies that hold training data size and model architecture constant across strategies.
Extending SRL coverage to low-resource languages therefore remains one of the most data-intensive open challenges in the field.

\subsection{Multi-modal SRL}
\label{Multi-modal SRL}

\begin{figure*}[ht]
    \centering
    \includegraphics[width=.9\linewidth]{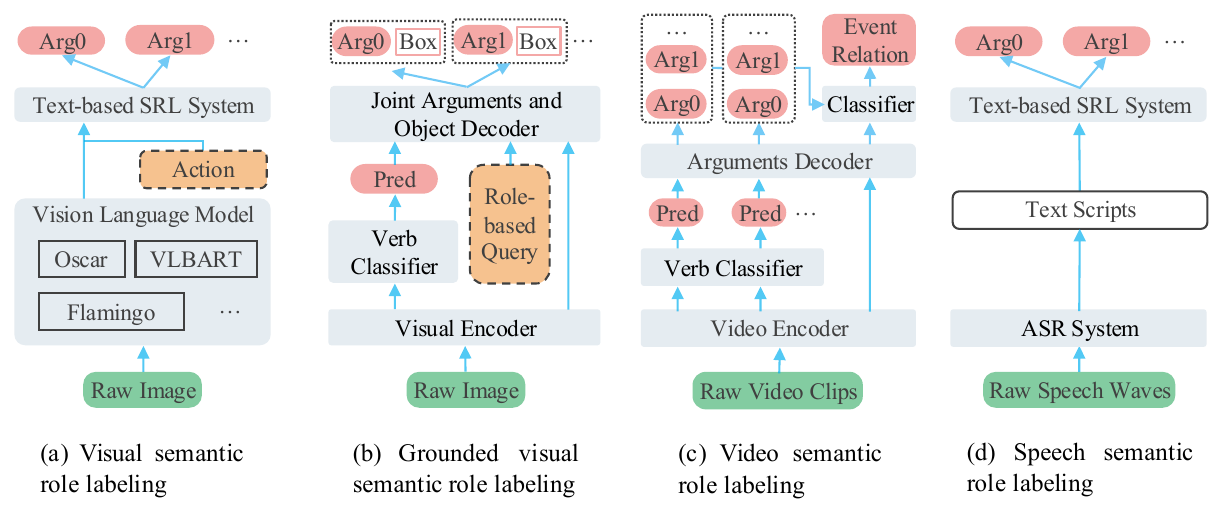}
    \caption{Pipeline architectures for multimodal SRL.}
\label{fig:multimodal-label}
\end{figure*}

While traditional SRL has predominantly focused on textual data, recent years have witnessed a growing interest in extending SRL to non-text modalities, including images, videos, and speech.
Non-text SRL presents unique challenges and opportunities that distinguish it from conventional text-based approaches.
A fundamental conceptual difference lies in the nature of argument identification: textual SRL locates arguments as discrete token spans within a sentence, whereas visual and video SRL must ground semantic roles onto spatial regions or temporal segments in continuous perceptual inputs, where no explicit boundary markers exist.
This grounding requirement introduces challenges that are absent in textual SRL, including visual ambiguity in role-filler assignment, sensitivity to viewpoint and occlusion, and the need to resolve co-referential visual entities across frames in video.
A central difficulty shared across all non-text modalities is the modality gap: visual and acoustic features occupy fundamentally different representational spaces from the linguistic semantic roles defined in schemes such as PropBank and FrameNet, and bridging this gap requires learning a cross-modal alignment that is not directly supervised by any unimodal objective.
A further challenge is annotation scalability: unlike textual SRL corpora constructed through expert linguistic annotation, multimodal SRL datasets rely on crowdsourced or semi-automatic procedures that introduce label noise and yield corpora orders of magnitude smaller than their textual counterparts.
Regarding evaluation, the metrics used across modalities are not directly comparable.
Textual SRL is evaluated by exact span or dependency head matching against gold annotations, yielding precision, recall, and F$_1$ scores over discrete label sets.
Visual SRL introduces additional grounding metrics such as \textit{Grounded-Value} and \textit{Grounded-Value-All}, which jointly assess role label correctness and the spatial accuracy of the predicted bounding box, and a prediction is counted as correct only when both conditions are satisfied simultaneously.
Video SRL further incorporates temporal evaluation criteria, including verb accuracy, CIDEr scores for argument generation, and event relation accuracy, reflecting the richer structured output required in the video setting.
These differences mean that F$_1$ scores reported on textual benchmarks and grounding accuracy reported on visual benchmarks measure fundamentally different aspects of semantic role prediction, and cross-modal performance comparisons should be interpreted with caution.
In Table~\ref{tab:result-visual}, we compare the results for multimodal SRL tasks.
The following paragraphs examine these non-lingual approaches.

\subsubsection{Image-related SRL}
\citet{DBLP:journals/corr/GuptaM15} coined the VSRL and built a benchmark based on the COCO images.
Almost at the same period, \citet{DBLP:conf/cvpr/YatskarZF16} also gave a similar definition for VSRL and introduced a dataset SituNet.
\citet{DBLP:conf/eccv/PrattYWFK20} proposed the idea of grounding nouns in the image.
Afterwards, mainstream researches used the term situation recognition (SR) or grounded situation recognition (GSR).
The key of VSRL is to model visual features which ground the semantic roles of the presented event.
Unlike textual argument identification, which operates over a finite sequence of tokens with clear positional boundaries, visual grounding must associate each semantic role with a spatial region in a continuous image, where candidate role-fillers are not pre-segmented and may overlap, occlude one another, or vary substantially in scale and appearance.
This spatial grounding requirement means that image SRL models must solve two coupled subproblems simultaneously: predicting the correct role label for each frame element, and localizing the corresponding visual entity within the image, making image SRL strictly harder than its textual counterpart even when the role inventory is held constant.
\citet{DBLP:conf/iccv/MallyaL17} used VGG as the visual feature encoder and a RNN as semantic role labeler.
\citet{DBLP:conf/iccv/LiTLJUF17} and \citet{DBLP:conf/iccv/SuhailS19} used GNN as graph based models to capture semantic structures in the image.
\citet{DBLP:conf/emnlp/SilbererP18} introduced a VSRL model that used noisy data from image captions.
\citet{DBLP:conf/cvpr/CoorayCL20} modeled VSRL as a query-based visual reasoning.
\citet{DBLP:conf/bmvc/ChoYLK21} adopted the visual transformer to create image representations.
More recently, \citet{hromei-etal-2025-grounded} extended visual SRL to the situated human-robot interaction setting, proposing a Grounded SRL framework that combines frame semantics with perceptual grounding.
Despite these advances, existing datasets cover a limited verb vocabulary and rely on fixed noun inventories for role fillers, which restricts generalization to open-domain visual scenes where role fillers are not drawn from a closed set.

\begin{table*}[t]
\fontsize{6}{7}\selectfont
\setlength{\tabcolsep}{0.5mm}

\begin{center}
\begin{NiceTabular*}{\textwidth}{@{\extracolsep{\fill}}lccccc}[
    code-before = \rowcolor{blue!15}{1} \rowcolor{gray!15}{4,6,8}
]
\Xhline{0.08em}
\multicolumn{6}{c}{\textit{\textbf{Visual SRL}}} \\
\bf Method & \bf Value & \bf Val-all & \bf Verb & \bf Grnd & \bf Grnd-all \\
\hline
CRF \citep{DBLP:conf/cvpr/YatskarZF16} & 24.6 & 14.2 & 32.3 & - & -\\
CRF+dataAug \citep{DBLP:conf/cvpr/YatskarOZF17} & 26.5 & 15.5 & 34.1 & - & -\\
VGG+RNN \citep{DBLP:conf/iccv/MallyaL17} & 27.5 & 16.4 & 35.9 & - & - \\
FC-Graph \citep{DBLP:conf/iccv/LiTLJUF17} & 27.5 & 19.3 & 36.7 & - & -\\
CAQ \citep{DBLP:conf/cvpr/CoorayCL20} & 30.2 & 18.5 & 38.2 & - & -\\ 
Kernel-Graph \citep{DBLP:conf/iccv/SuhailS19} & 35.4 & 19.4 & 43.3 & - & - \\
\end{NiceTabular*}

\begin{NiceTabular*}{\textwidth}{@{\extracolsep{\fill}}lccccc}[
    code-before = \rowcolor{blue!15}{1} \rowcolor{gray!15}{4,6,8,10}
]
\specialrule{.2em}{.05em}{0.05em} 
\multicolumn{6}{c}{\textit{\textbf{Visual SRL (grounded)}}} \\
\bf Method & \bf Value & \bf Val-all & \bf Verb & \bf Grnd & \bf Grnd-all \\
\hline
ISL \citep{DBLP:conf/eccv/PrattYWFK20} & 30.1 & 18.6 & 39.4 & 22.7 & 7.7\\
JSL \citep{DBLP:conf/eccv/PrattYWFK20} & 31.4 & 18.9 & 39.9 & 24.9 & 9.7\\
GSRTR \citep{DBLP:conf/bmvc/ChoYLK21} & 32.5 & 19.6 & 41.1 & 26.0 & 10.4\\
SituFormer \citep{DBLP:conf/aaai/Wei00YC22} & 35.5 & 21.9 & 44.2 & 29.2 & 13.4\\
CoFormer \citep{DBLP:conf/cvpr/ChoYK22} & 36.0 & 22.2 & 44.7 & 29.1 & 12.2\\
CLIP Event \citep{DBLP:conf/cvpr/LiXWZ0Z0JC22} & 33.1 & 20.1 & 45.6 & 26.1 & 10.6\\
GSRFormer \citep{DBLP:conf/mm/ChengDLM022} & 37.5 & 23.3 & 46.5 & 31.5 & 14.2\\
CRAPES$_2$ \citep{DBLP:conf/starsem/BhattacharyyaPH23} & 66.1 & 30.6 & 41.9 & 36.7 & 6.47\\
\end{NiceTabular*}

\begin{NiceTabular*}{\textwidth}{@{\extracolsep{\fill}}lccccccccc}[
    code-before = \rowcolor{blue!15}{1} \rowcolor{gray!15}{4,6,8,10,12}
]
\specialrule{.2em}{.05em}{0.05em} 
\multicolumn{10}{c}{\textit{\textbf{Video SRL}}} \\
\bf Method & \bf V-Acc@5 & \bf V-Rec@5 & \bf CIDEr & \bf R-L & \bf C-Vb & \bf C-Arg & \bf Lea & \bf Lea-S & \bf ER-Acc \\
\hline
VidSitu-GPT2 \citep{DBLP:conf/cvpr/SadhuGYNK21} & - & - &  34.67 &  40.08 &  42.97 &  34.45 &  48.08 &  28.10 & - \\
VidSitu-I3D \citep{DBLP:conf/cvpr/SadhuGYNK21} & 66.83 & 4.88 &  47.06 &  42.41 &  51.67 &  42.76 &  48.92 &  33.58 & - \\
VidSitu-SlowFast \citep{DBLP:conf/cvpr/SadhuGYNK21} & 69.20 & 6.11 &  45.52 &  42.66 &  55.47 &  42.82 &  \underline{50.48} &  31.99 & \underline{34.13} \\
VidSitu-e2e \citep{DBLP:conf/aaai/YangLZ0JC23} & 75.90 & 23.38 & 30.33 & 29.98 & 39.56 & 23.97 & 35.92 & - & - \\
OME \citep{DBLP:conf/aaai/YangLZ0JC23} & 83.88 & 28.44 & 47.82 & 40.91 & 54.51 & 44.32 & - & - & - \\
OME(disp) \citep{DBLP:conf/aaai/YangLZ0JC23} & \underline{84.00} & 28.61 & 48.46 & \underline{41.89} & 56.04 & \underline{44.60} & - & - & - \\
OME(disp)+OIE \citep{DBLP:conf/aaai/YangLZ0JC23} & 83.94 & \underline{28.72} & 47.16 & 40.86 & 53.96 & 42.78 & - & - & - \\
VideoWhisperer \citep{DBLP:conf/nips/KhanJT22} & 75.59 & 25.25 & \underline{52.30} & 35.84 & \underline{61.77} & 38.18 & 38.00 & - & - \\
HostSG \citep{DBLP:conf/mm/Zhao00LZWZC23} & 86.33 & 29.38 & 55.09 & 43.13 & 64.24 & 47.68 & 55.70 & 35.01 & 35.97 \\
Slow-D+TxE+TxD \citep{DBLP:conf/cvpr/XiaoKTM22} & 74.4 & 18.4 & 60.34 & 43.77 & 69.12 & 53.87 &  46.77 & - & 34.71 \\
\end{NiceTabular*}

\begin{NiceTabular*}{\textwidth}{@{\extracolsep{\fill}}lcccccccc}[
    code-before = \rowcolor{blue!15}{1} \rowcolor{gray!15}{5,7}
]
\specialrule{.2em}{.05em}{0.05em} 
\multicolumn{9}{c}{\textit{\textbf{Speech SRL}}} \\
\multirow{2}{*}{\bf Method} & \multicolumn{4}{c}{\bf CPB1.0} & \multicolumn{4}{c}{\bf AS-SRL}\\
\cmidrule{2-5}\cmidrule{6-9}
 & \bf CER & \bf P & \bf R & \bf F$_1$ & \bf CER & \bf P & \bf R & \bf F$_1$ \\
 \hline
Whisper \citep{DBLP:conf/acl/ChenLZZ24} & 3.16 & 81.25 & 72.53 & 76.64 & 4.45 & 74.21 & 71.44 & 72.80 \\
Whisper(+GS AUG) \citep{DBLP:conf/acl/ChenLZZ24} & 3.16 & 80.95 & 72.93 & 76.73 & 4.45 & 77.49 & 70.70 & 73.94\\
Whisper-End2End (gold SRL) \citep{DBLP:conf/acl/ChenLZZ24} & 3.16 & 81.21 & 73.15 & 76.92 & 4.48 & 75.35 & 70.86 & 73.04\\
Whisper-End2End \citep{DBLP:conf/acl/ChenLZZ24} & 3.16 & 80.27 & 74.19 & 77.11 & 4.47 & 75.84 & 72.25 & 74.00\\
\bottomrule
\end{NiceTabular*}

\end{center}
\caption{
Visual SRL results on SWiG, Video SRL results on VidSitu, and Speech SRL results on CPB 1.0 and AS-SRL. Val-all: Value-all, Grnd: Grounded-value, Grnd-all: Grounded-value-all, V-Acc@5: Verb Accuracy@Top 5, V-Rec@5: Verb Accuracy@Recall 5, R-L: Rouge-L, C-Vb: CIDEr Verb, C-Arg: CIDEr Argument, ER-Acc: Event Relation Accuracy. A dash (-) indicates that the corresponding metric was not reported in the original paper.
}
\label{tab:result-visual}
\end{table*}

\subsubsection{Video-related SRL}
Similar to VSRL, \citet{DBLP:conf/cvpr/SadhuGYNK21} extended the task to the scenario of video modality, namely VidSRL, and presented the benchmark VidSitu.
Different from the static scenes in image modeling, video understanding is concerned with understanding both the spatial semantics and the temporal changes.
This temporal dimension introduces a challenge unique to video SRL: a semantic role filler identified in one frame must be tracked consistently across subsequent frames, even under motion blur, occlusion, or scene transitions, none of which arise in textual or image-based SRL.
The modality gap in video SRL is therefore two-dimensional, requiring alignment between linguistic role categories and both the spatial appearance and the temporal trajectory of visual entities, and current models based on frame-level feature aggregation or object tracking do not fully resolve the identity consistency problem when a role filler undergoes significant appearance change across a clip.
The annotation cost of video SRL is substantially higher than that of image SRL, as each role filler must be localized both spatially and temporally across multiple frames, and the resulting datasets remain small relative to the complexity of the task.
The evaluation of VidSRL therefore requires metrics that assess not only role label accuracy but also temporal coherence of argument tracking, making direct comparison with textual SRL F$_1$ scores inappropriate.
\citet{DBLP:conf/nips/KhanJT22} grounded the visual objects and entities across verb-role pairs for VidSRL.
\citet{DBLP:conf/aaai/YangLZ0JC23} proposed tracking the object-level visual arguments so as to model the changes of states.
\citet{DBLP:conf/mm/Zhao00LZWZC23} used a graph-based framework to model event-level semantic structures in both spatio and temporal dimensions.

\subsubsection{Speech-related SRL}
Traditional approaches typically employed a pipeline architecture, where ASR preceded text-based SRL, but this approach suffered from error propagation and loss of valuable acoustic feature.
A unique challenge in speech SRL, absent from both textual and visual SRL, is that the input signal carries no explicit word boundary markers, and recognition errors in upstream ASR directly corrupt the predicate-argument structure predicted downstream.
The modality gap in speech SRL manifests as a misalignment between acoustic feature sequences and the token-level role spans defined in text-based annotation: the acoustic encoder operates over continuous frames whereas role labels are defined over discrete word units, and this representational mismatch must be resolved before role classification can proceed.
End-to-end training mitigates but does not fully eliminate this error propagation, since the acoustic encoder must simultaneously learn to segment speech and assign semantic roles, two objectives that may conflict under low-resource or noisy conditions.
\citet{DBLP:conf/acl/ChenLZZ24} introduced the first end-to-end learning framework for Chinese speech-based SRL using a Straight-Through Gumbel-Softmax module to bridge ASR and SRL components.
This innovative approach enables joint optimization and direct utilization of acoustic features, solving key challenges of ASR-annotation alignment and acoustic feature integration.
The only publicly available speech SRL dataset covers Chinese and contains approximately 9,000 utterances, which is insufficient to support robust evaluation of cross-domain or cross-speaker generalization, and constructing larger and more diverse speech SRL corpora remains an open challenge.

\section{SRL Applications}
\label{Applications}

SRL has demonstrated its fundamental value across diverse application domains, extending beyond traditional NLP to emerging fields in artificial intelligence.
This section explores three major application areas where SRL has made significant impacts:
NLP tasks where SRL enhances semantic understanding for various language processing applications, robotics where it enables natural language instruction interpretation, and embodied AI where SRL bridges language comprehension with physical world interaction.
These applications showcase the versatility of SRL in advancing human-machine interaction across different domains.

\subsection{Downstream NLP Tasks}
SRL has established itself as an important component in various NLP applications, providing crucial semantic information that enhances the performance of downstream tasks.
In this subsection, we explore the significant applications of SRL in different domains, highlighting its contributions and practical implications.

In information extraction, SRL provides structured representations that enhance the identification of events and their participants.
This structural information improves the accuracy of relationship extraction between entities and helps in understanding event chains.
Early work demonstrated that SRL-derived argument structures can serve as a backbone for open information extraction, enabling systems to identify who did what to whom across large corpora \citep{christensen-etal-2010-semantic,DBLP:conf/kcap/ChristensenMSE11}.
More recent studies have further shown that sentence simplification guided by SRL annotations improves the coverage and precision of extracted relations \citep{DBLP:conf/ranlp/EvansO19}.

In machine translation, SRL improves translation quality by providing explicit predicate-argument structures, which help maintain semantic consistency across languages with different syntactic patterns.
The semantic roles identified help ensure that the relationships between events and their participants are preserved during translation, a benefit that has been validated across both phrase-based and neural translation frameworks \citep{DBLP:conf/acl/ShiLRFLZSW16,DBLP:conf/naacl/MarcheggianiBT18}.

For question answering systems, SRL facilitates better answer extraction by matching the semantic structures between questions and potential answers.
By analyzing the alignment of predicate-argument structures, systems can more accurately identify relevant answers, particularly for complex questions involving multiple entities or events.
This line of research has progressed from early pipeline approaches to tightly integrated neural architectures that jointly model question semantics and answer spans \citep{DBLP:conf/emnlp/ShenL07,DBLP:conf/emnlp/BerantCFL13,DBLP:conf/emnlp/HeLZ15,DBLP:conf/acl/YihRMCS16}.

Text summarization benefits from SRL through improved content selection and organization.
The predicate-argument structures help identify key semantic relationships in the source text, ensuring that essential semantic information is preserved in the generated summaries while maintaining coherence \citep{DBLP:journals/asc/KhanSK15,DBLP:journals/ipm/MohamedO19}.

\subsection{Language Modeling}

SRL serves as an attractive component for enhancing language modeling capabilities, especially in improving semantic understanding and generation of natural language.
Integrating SRL into language modeling has shown consistent improvements in capturing dependencies and relationships between predicates and their arguments.

A representative line of work exploits predicate-argument structure to enrich pretrained representations.
\citet{DBLP:conf/aaai/0001WZLZZZ20} showed that incorporating SRL into BERT-based architectures enables language models to capture semantic relations more precisely and produce more accurate semantic interpretations.
A related direction applies SRL to dialogue understanding.
\citet{DBLP:conf/emnlp/XuTSWZSY20} demonstrated that SRL annotations can significantly improve a dialog rewriting task without adding additional model parameters, by introducing cross-transitive predicate-argument annotations that improve dialog coherence and information integrity.
Beyond supervised settings, SRL has also been combined with search-based optimization strategies.
\citet{DBLP:journals/jksucis/Onan23a} integrated SRL with the Ant Colony Optimisation algorithm, allowing models to more accurately capture semantic relationships and role information in sentences and generate more natural and meaningful text.
\citet{DBLP:conf/coling/ZouG0LCLAS24} further revealed that SRL can guide the extraction of key local semantic components while filtering out noisy elements such as punctuation and discourse fillers, resulting in a more robust feature representation.

\subsection{Robotics}
\label{section:robot}
SRL has emerged as a powerful tool in advancing the field of robotics, particularly in enhancing natural language understanding for human-robot interaction.
This section focuses on the fundamental aspects of robot command interpretation and execution based on semantic role analysis.

A central application of SRL in robotics is the interpretation of natural language instructions for task execution.
\citet{DBLP:conf/ecai/BastianelliCCBN14} demonstrated how SRL can map linguistic elements to specific robot actions and environmental objects by combining linguistic information with contextual knowledge about the environment.
Building on this foundation, \citet{DBLP:journals/ieeejas/LuC17} advanced robot instruction understanding by leveraging argument typed dependency features and integrating open knowledge resources, moving away from traditional hand-coded knowledge approaches toward a more flexible and scalable solution for human-robot interaction.
\citet{DBLP:conf/cvpr/0016J0021} further contributed to this direction by addressing event compatibility and sample suitability through verb-specific semantic roles, offering insights for robotic scene understanding that extend beyond the primary focus on image captioning.
\citet{hromei-etal-2025-grounded} advanced this line of research by proposing a Grounded SRL framework for situated human-robot interaction, in which frame-semantic representations are directly linked to perceptual evidence in the robot's visual field.
This work represents a concrete step toward closing the gap between linguistic semantic parsing and real-world robotic command execution.

\subsection{Advanced Embodied Intelligence}
While Section~\ref{section:robot} covered basic robotic applications, this section explores how SRL enables more sophisticated embodied intelligence capabilities, particularly in scenarios requiring complex environmental perception, context understanding, and adaptive behavior.

A key advancement in embodied AI is the integration of environmental perception with semantic understanding.
\citet{DBLP:conf/acl-jssp/BastianelliCCB13} pioneered this direction by developing a real-time SRL approach that generates semantic tree representations while considering the physical environment.
Resolving language ambiguities in real-world contexts represents another frontier in embodied AI. \citet{DBLP:conf/naacl/YangGLXZC16} addressed this through a visual-linguistic framework applied to cooking scenarios, where both explicit and implicit semantic roles must be understood in relation to the physical environment.
A significant advance in scalable embodied intelligence came from \citet{DBLP:journals/ai/VanzoCBBN20}, who developed a language-independent framework for robotic command interpretation that achieves context-aware disambiguation of instructions based on the current state of the environment.
Building upon these advances, \citet{DBLP:journals/kbs/ZhangTZD21} introduced a validation mechanism using SRL tags to ensure semantic consistency in complex action sequences, enabling the verification of not just individual actions, but entire sequences of behaviors against their original semantic intentions.

\section{Future Directions}
\label{Feature Research}

\subsection{Knowledge-Enhanced SRL}
Knowledge enhancement presents several promising research directions for advancing SRL systems.
Current SRL models primarily rely on surface-level textual features, leaving three categories of knowledge underrepresented: commonsense knowledge about typical event participants and causal relations, domain-specific ontological constraints on role fillers, and cross-lingual lexical knowledge for low-resource transfer.
The integration of external resources such as ConceptNet, WordNet, and domain ontologies could enable SRL systems to resolve argument boundaries where surface syntax provides insufficient cues, and to recover implicit arguments that current models consistently fail on.
However, generic ontologies often provide labels that are too abstract for domain-specific applications, particularly in technical fields such as engineering, where precise terminology and fine-grained relational structures are essential for accurate role assignment \citep{siddharth2022natural}.
Domain-specific ontologies, by contrast, are better suited to capturing the specialized predicate-argument patterns that arise in such contexts, and their construction and integration into SRL pipelines represents an important direction for future research.
Knowledge distillation offers a complementary path, transferring role-labeling knowledge from large-scale models to lightweight deployable systems.
Evaluation of knowledge integration should go beyond standard F$_1$ to include held-out tests on implicit argument recovery and out-of-domain role classification, isolating gains attributable to external knowledge rather than model capacity.
Furthermore, domain-specific knowledge graphs offer targeted opportunities in biomedical and legal domains, where fine-grained semantic relationships and specialized terminology pose persistent challenges for general-purpose SRL models.

\subsection{Scenario-optimized SRL}
\label{Scenario-optimized SRL}

A critical direction for advancing SRL is domain-specific optimization and practical application.
Current SRL models have shown strong performance on general-domain benchmarks, yet substantial gaps remain when these models are applied to specialized domains such as healthcare documents, legal contracts, and financial reports, where domain-specific semantic structures and terminologies pose unique challenges.
In low-resource specialized domains, graph-based methods with explicit syntactic supervision provide a useful inductive bias that partially compensates for limited in-domain annotations.
PLM-based methods fine-tuned on small domain-specific datasets remain prone to overfitting, and retrieval-augmented approaches that inject domain knowledge at inference time offer a complementary path that avoids full retraining.
When domain shift is anticipated at deployment time, models should be evaluated on held-out out-of-domain test sets as a standard requirement rather than relying solely on in-domain benchmark scores.
Furthermore, with the increasing popularity of interactive AI systems, optimizing SRL models for real-time processing in human-computer dialogue scenarios has become an important research direction, especially for reducing latency while maintaining accuracy in dynamic conversational contexts.
In dialogue and streaming settings where response latency is critical, compact discriminative architectures with deterministic decoding are generally preferable to autoregressive generative models.
The adaptation of SRL systems to emerging application domains such as augmented reality interfaces and automated reasoning systems is also an important frontier to be investigated, given the evolving nature of human-computer communication and the increasing complexity of semantic understanding tasks in these new contexts.
A further consideration in scenario-specific deployment is the trustworthiness of SRL outputs in high-stakes settings.
When SRL models are integrated into applications such as clinical decision support or legal reasoning, biases inherited from training corpora can translate into consequential errors that affect real users.
Scenario-optimized SRL development should therefore incorporate fairness auditing and bias assessment as standard components of the evaluation pipeline, alongside domain-specific performance metrics.

\subsection{Interpretable and Robust SRL}
An important future direction for SRL research is to improve the interpretability and robustness of SRL systems.
While current SRL systems achieve strong benchmark performance, their decision processes remain opaque, making error diagnosis and trust calibration difficult in deployment settings.
Concrete approaches include attention-based explanation methods that identify input tokens driving role assignments, probing classifiers that test whether syntactic properties are encoded in intermediate representations, and counterfactual analysis that measures prediction sensitivity to controlled input perturbations.
Interpretability gains should be evaluated through faithfulness metrics assessing whether highlighted evidence causally influences predictions, alongside human studies testing whether explanations support effective error diagnosis.
On the robustness side, evaluation should extend beyond standard test sets to cover domain shift, label noise, and adversarially constructed role-swapping examples.
Targeted training strategies such as syntactic paraphrase augmentation, adversarial role-confusable examples, and confidence calibration methods offer concrete paths toward more reliable SRL systems in real-world deployment.
Improving interpretability is also a prerequisite for responsible deployment in socially sensitive contexts, where stakeholders require transparent and auditable model behavior to identify and mitigate potential harms arising from biased predictions.

\subsection{Multimodal SRL}
The development of a unified multimodal framework for seamlessly integrating semantic role analysis across text, image, video, and speech modalities is an essential future direction for SRL research.
The field needs to move beyond traditional text-centric approaches to capture the rich semantic interactions that naturally occur in multimodal communication.
This evolution requires addressing several interrelated challenges.
First, the modality gap between perceptual representations and linguistic role categories calls for cross-modal alignment objectives that explicitly supervise the correspondence between visual or acoustic features and predicate-argument structures, with contrastive learning between modality-specific encoders and a shared role embedding space offering one promising direction.
Second, annotation scalability remains a bottleneck: active learning strategies and cross-modal transfer from richly annotated text SRL corpora could reduce the annotation burden in visual and speech settings, while semi-automatic pipelines combining object tracking with human verification are particularly relevant for video SRL.
Third, the field lacks a unified evaluation protocol that can assess role label accuracy, argument localization quality, and cross-modal consistency within a single framework, and developing such a protocol is a prerequisite for meaningful cross-modal progress measurement.
Establishing a unified representation scheme to efficiently model the semantic roles of different modalities, developing cross-modal knowledge transfer mechanisms to reduce the heavy reliance on annotated training data, and enhancing the robustness of multimodal SRL systems for real-world applications where noise and modal mismatches are prevalent remain the core research agenda.
In scenarios where meaning is conveyed simultaneously through multiple channels, such as in multimedia content analysis, human-computer interaction, and multimodal event understanding, these advances will help to achieve a more comprehensive semantic understanding.

\subsection{LLM-based SRL}
The relationship between LLMs and specialized SRL systems is best understood as complementary rather than competitive.
LLMs encode vast linguistic knowledge that benefits SRL through richer contextual representations and improved generalization, while specialized systems retain distinct value by producing verifiable predicate-argument structures that downstream tasks can consume directly.
Current evidence indicates that LLMs alone are not sufficient replacements for specialized SRL systems, as they struggle with precise argument boundary detection, long-distance dependency resolution, and consistent role assignment in structurally complex sentences.
In high-stakes domains such as clinical text processing and legal document analysis, where output reliability and auditability are required, specialized SRL systems offer controllable and interpretable predictions that LLMs currently cannot match.
Future research should therefore explore hybrid approaches in which LLMs supply rich contextual representations while specialized architectures enforce structural constraints, and investigate retrieval-augmented prompting strategies that bring LLM performance closer to supervised baselines in zero-shot and few-shot settings.

\subsection{Efficient and Deployable SRL}
\label{Efficient and Deployable SRL}
As SRL systems grow in complexity, computational efficiency and practical deployability have become increasingly important considerations.
Large Transformer-based models and LLMs achieve strong performance on standard benchmarks, but their high memory requirements and slow inference speeds create barriers to real-world adoption, particularly in latency-sensitive or resource-constrained settings.
Several directions deserve attention in this regard.
Model compression techniques such as pruning, quantization, and knowledge distillation offer promising paths toward reducing model size without substantial loss in accuracy, enabling SRL systems to run efficiently on commodity hardware or edge devices.
Efficient attention mechanisms, including linear attention variants and sparse attention patterns, can reduce the quadratic complexity of standard Transformers and improve throughput on long-document inputs.
For streaming or interactive applications, such as real-time dialogue systems and robotic command interpretation, incremental or online inference strategies are needed to process inputs as they arrive rather than waiting for complete sentence boundaries.
Beyond inference efficiency, the reproducibility and stability of model outputs matter for deployment in production pipelines.
Prompt-based LLM approaches, while flexible, introduce variability that can be difficult to control in downstream systems that rely on consistent structured outputs.
Specialized SRL architectures with deterministic decoding offer a more predictable alternative in such scenarios.
Future work should therefore pursue a balanced approach that advances both accuracy and efficiency, developing SRL systems that are not only competitive on benchmarks but also practical to deploy across diverse real-world environments.
Data efficiency represents a parallel challenge across all four paradigms: higher-capacity models achieve stronger performance but demand more annotated data, while lower-capacity models are more annotation-efficient but sacrifice accuracy on complex phenomena.
Active learning, semi-supervised training, and cross-task transfer offer promising directions for reducing annotation cost, and remain especially critical for extending SRL to low-resource languages and specialized domains.
Beyond model-side efficiency, reducing the annotation burden through pre-annotation tools, question-answer driven crowdsourcing, and shared annotation infrastructure represents an equally important path toward scalable SRL development.

\subsection{Discourse SRL}
Discourse-level SRL represents another crucial direction for future research, particularly given the capabilities of modern language models.
While research in this area has been relatively quiet since the seminal works, the emergence of LLMs with extensive context windows creates new opportunities for advancement.
These models' ability to process and understand long-range dependencies makes them particularly well-suited for addressing traditional challenges in discourse SRL, such as resolving implicit arguments and connecting semantic roles across sentence boundaries.
The sophisticated understanding of document-level coherence exhibited by modern LLMs could help bridge local predicate-argument structures with broader discourse contexts, potentially leading to more comprehensive semantic analysis systems.
This direction could be especially valuable for applications requiring deep understanding of long documents, such as document-level information extraction and reading comprehension.

\section{Conclusion}
\label{Conclusion}

This paper presents a comprehensive survey of semantic role labeling (SRL) research over the past two decades, capturing its theoretical foundations, methodological advancements, and practical applications. 
We categorize SRL methodologies into four key perspectives: model architectures, syntax feature modeling, application scenarios, and multi-modal extensions. 
We also provide an overview of SRL benchmarks, evaluation metrics, and paradigm modeling approaches, highlighting the evolution of SRL across text, visual, and speech modalities. 
Furthermore, we explore the practical applications of SRL in various domains, emphasizing its significance in real-world tasks. 
Finally, we discuss the future directions of SRL research, particularly its integration with large language models (LLMs) and its potential impact on the NLP and multimodal landscape.
Although LLMs have transformed the broader NLP landscape, SRL retains distinct and lasting value as a structured semantic layer that provides verifiable predicate-argument representations, supports downstream task reliability, and serves as a diagnostic tool for evaluating the semantic capabilities of foundation models.
The most productive path forward lies in combining the representational power of LLMs with the structural precision of specialized SRL systems, enabling more robust and interpretable semantic understanding across diverse real-world applications.


\bibliography{custom}




\end{document}